\documentclass{article}
\usepackage[margin=1.0in]{geometry}
\usepackage{hyperref}
\usepackage{graphicx} %
\usepackage{float}    %
\usepackage[dvipsnames]{xcolor}
\usepackage{subcaption} %
\usepackage{siunitx}

\usepackage[sort&compress]{natbib}

\usepackage{listings}
\usepackage[most]{tcolorbox}
\usepackage{authblk} %

\usepackage[normalem]{ulem} %

\title{Questioning Representational Optimism in Deep Learning: \\The Fractured Entangled Representation Hypothesis}

\makeatletter
\renewcommand\AB@affilsepx{, }
\makeatother

\author[1]{\mbox{Akarsh Kumar}}
\author[2,3]{\mbox{Jeff Clune}}
\author[4]{\mbox{Joel Lehman}}
\author[5]{\mbox{Kenneth O. Stanley}}
\affil[1]{MIT}
\affil[2]{University of British Columbia}
\affil[3]{Vector Institute}
\affil[4]{University of Oxford}
\affil[5]{Lila Sciences}

\newcommand{\codelink}{https://github.com/akarshkumar0101/fer}

\date{}

\begin{document}

\maketitle

\vspace{-10mm}
\begin{figure}[!h]
    \centering
    \includegraphics[width=0.8\textwidth]{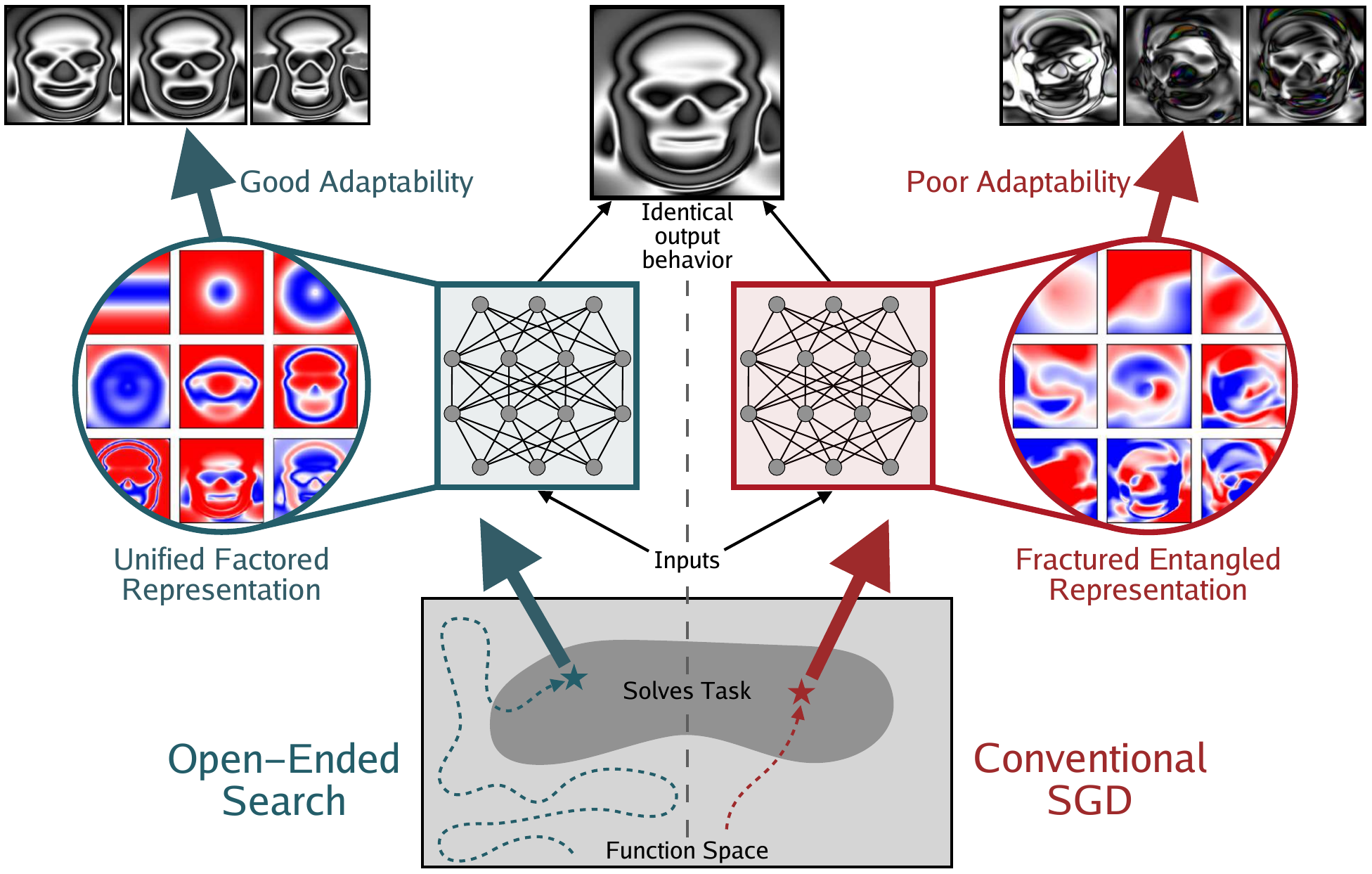} %
    \caption{
    \textbf{Overview.}
    An open-ended search process (left) from the Picbreeder experiment yielded a neural network that outputs the image of a skull.
    Conventional SGD (right) is able to learn to produce precisely the same image, but by following a completely different path through function space during learning.
    However, peering inside these networks with identical output behavior reveals a radically different style of representation, raising questions about the adaptability of networks (including large models) trained through conventional SGD, which can impact generalization, creativity, and (continual) learning.
    }
    \label{fig:teaser}
\end{figure}

\vspace{-3mm}

\begin{abstract}
Much of the excitement in modern AI is driven by the observation that scaling up existing systems leads to better performance.
But does better performance necessarily imply better internal representations?
While the representational optimist assumes it must, this position paper challenges that view.
We compare neural networks evolved through an open-ended search process to networks trained via conventional stochastic gradient descent (SGD) on the simple task of generating a single image.
This minimal setup offers a unique advantage: each hidden neuron's full functional behavior can be easily visualized as an image, thus revealing how the network's output behavior is internally constructed neuron by neuron.
The result is striking: while both networks produce the same output behavior, their internal representations differ dramatically.
The SGD-trained networks exhibit a form of disorganization that we term \textit{fractured entangled representation} (FER).
Interestingly, the evolved networks largely lack FER, even approaching a \textit{unified factored representation} (UFR).
In large models, FER may be degrading core model capacities like generalization, creativity, and (continual) learning.
Therefore, understanding and mitigating FER could be critical to the future of representation learning.

\bigskip

\noindent \textbf{Code and additional data: }\href{\codelink}{\codelink} %
\end{abstract}

\section{Introduction}

A neural network’s knowledge, generalization, creativity, continual learning ability, and overall potential are ultimately determined by its internal representations.
These representations---how it models the world---are encoded through the structure of its neural circuitry (connection weights).
Even as researchers increasingly probe the internal representations of neural networks (NNs)~\citep{yosinski2015understanding, nguyen2016synthesizing, olah2017feature, elhage2022toy, olsson2022context, lindsey2024biology},
an implicit philosophy of \emph{representational optimism} has emerged---rarely stated outright but carrying profound implications.
The hope of the representational optimist is that, as data and compute are scaled, good representations naturally develop on their own.

While scaling clearly improves the \textit{performance} of the representations on downstream tasks~\citep{kaplan2020scaling, bubeck2023sparks}, less attention has been given to understanding the \textit{nature} of these representations as models grow.
Of course, even representational optimists still want to understand internal representations (e.g.\ through mechanistic interpretability as in \citealt{sharkey2025open}), but the motivation is more through a desire to 
understand how the networks work~\citep{olah2017feature, yosinski2015understanding}, how they can be controlled, 
and their
level of alignment~\citep{bereska2024mechanistic, lindsey2024biology} than through any concern that their representations may be fundamentally flawed.

At the heart of representational optimism is the belief that representation improves with scaling in deep learning \citep{kaplan2020scaling, hoffmann2022training, huh2024platonic}.
For example, the performance leaps from GPT-2~\citep{radford2019language} to GPT-3~\citep{brown2020language} and GPT-4~\citep{achiam2023gpt} suggest that the underlying representation should also be improving.
As \citet{brown2020language} state in the introduction to the technical report introducing GPT-3, where they place focus on the implications for representation (boldface added for emphasis):
\begin{quote}
Recent years have featured a trend towards pre-trained language \textbf{representations} in NLP systems, applied in increasingly
flexible and task-agnostic ways for downstream transfer.
First, single-layer \textbf{representations} were learned using word vectors %
and fed to task-specific architectures, then RNNs with \textbf{multiple layers of representations} and contextual state \textbf{were used to form stronger representations} 
(though still applied to task-specific architectures), and more recently pre-trained recurrent or transformer language models...
\end{quote}
The implication here that representation is improving makes intuitive sense; after all, how could it be possible that performance improves without the underlying representation also getting better~\citep{bengio2013representation}?
Because of that, the thinking goes, as models are scaled further, they will continue improving, and little worry need arise, at least with respect to representation.
Moreover, the mechanistic interpretability literature often studies a single neuron \citep{yosinski2015understanding,nguyen2016synthesizing,carter2019activation} or a small group of neurons, reinforcing representational optimism by inadvertently diverting attention away from potential problems at a more holistic rather than local level~\citep{templeton2024scaling}.
Even the representation learning literature mostly focuses on only the final layer features~\citep{chen2020simple, grill2020bootstrap, he2020momentum, radford2021learning, caron2021emerging}.

Yet what if something important is actually missing in the learned representation?
The main contribution of this position paper is to raise the possibility that in practice representation might not simply work itself out well by scaling after all (Figure~\ref{fig:teaser}).

To address the question of whether representations may be flawed \emph{even when benchmark performance is good}, we shift the focus from studying individual neurons to studying the broader, \emph{holistic representational strategy} across the entire network.
By doing so, we are able to make a distinction between an adequate representation that solves a task and an ideal representation of that solution.
The core of this concern is crystallized through a novel hypothesis that is based on a set of little-known observations of neural network representation that arose initially in the fields of neuroevolution and artificial life \citep{lehman2011abandoning, lehman2010efficiently, secretan2011picbreeder, clune2011performance, woolley2011deleterious, huizinga2018emergence, lehman2025evolution}:

\paragraph{The Fractured Entangled Representation (FER) hypothesis.}
We hypothesize that when deep learning models are trained to achieve a specific objective,
typically through backpropagation and stochastic gradient descent, the resultant representation tends to be fractured, which means that information underlying the same unitary concepts (e.g.\ how to add numbers) is split into disconnected pieces.
Importantly, \emph{these pieces then become redundant} as a result of their fracture: they are separately invoked in different contexts to model the same underlying concept,
when ideally, a single instance of the concept would have sufficed.
In other words, where there would ideally be the reuse of one deep understanding of a concept, instead there are different mechanisms for achieving the same function.

At the same time, these fractured (and hence redundant) functions tend to become entangled with other fractured functions, which means that behaviors that should be independent and modular end up influencing each other in idiosyncratic and inappropriate ways.
For example, a set of neurons within an image generator that change hair color might also cause the foliage in the background to change as well, and separating these two effects could be impossible.

Ideally, FER would be absent after training and internal representations would be the opposite: unified (a single unbroken function for each key capability) and factored (keeping independent capabilities separated so that they do not interfere with each other).
In practice, a well-factored representation should resemble a modular decomposition of the desired behavior.
While such a perfect \textbf{unified factored representation} (UFR) may not be attainable in practice, it serves as an aspirational ideal contemplated in this paper.

Intuitively, FER is like spaghetti code---messy, redundant, filled with ``if'' statements for every possible case, and littered with copy-pasted logic that performs the same task in slightly different ways across different files.
In contrast, UFR is like clean, modular software where each function is defined once and reused appropriately, making the code easy to understand or extend.
This structural clarity matters: just as good software design supports robustness, adaptability, and innovation, a unified representation can enable generalization, creativity, and (continual) learning.

In fact, the potential for FER to significantly disrupt generalization, creativity, and learning is precisely why it is critical to bring awareness to the issue.
After all, these are the core capabilities tasked with propelling the field into the future.
If they are fundamentally hindered, then without mitigation many lofty expectations will remain unmet.
To make its potential damage clear, here are the potentially pernicious effects of FER in these three key areas:

\begin{enumerate}
    \item FER potentially impacts \emph{generalization} wherever coverage is sparse in the training data.
Where there are insufficient training examples, a neural network or LLM has to interpolate or extrapolate.
If general principles from outside the area of sparsity can be applied within that area, then the interpolation can still succeed.
However, if those principles are fractured and therefore only selectively applied to narrow and arbitrary subdomains (and entangled and therefore yielding unintended side effects), then interpolation will not be based on those more fundamental regularities, diminishing its power. Because it is more likely to be seen in areas of sparsity, the impact of FER on generalization is particularly problematic in the borderlands of human knowledge, where little is yet written or known.
A failure of LLMs to extend or apply relevant regularities to these borderlands would be expected to surface as a clumsiness in grappling with novel material.
If that is the case, it is a particularly unfortunate deficit, because the very place where AI can potentially make the most exciting contributions is at the borderlands of knowledge~\citep{amodei2024machines}.
\item \emph{Creativity} is an entire discipline~\citep{boden2004creative} over which the small commentary here cannot do justice.
Nevertheless, it is still important to highlight that the ability to imagine a new artifact of a particular type requires understanding the regularities of that type.
The iPhone was once a new idea, but it extended many of the regularities of the concepts of a computer and a phone (and blended them very well). A fractured and entangled representation of either phones or computers may not have enabled extending those concepts so far from their incarnations at that time.
Even if the creative act is intentionally to break a regularity, other regularities should still be preserved to the extent possible in alignment with that broken regularity: in a butterfly where one wing is smaller, it still might make sense to warp and compress the same pattern as in the larger wing, preserving \emph{some} of the symmetry and thereby creating a plausible version of something that has never existed. 
For these reasons, the toll of FER on creativity is likely steep.
Studies of weight sweeps in networks with FER in this paper confirm this intuition, showing a dramatic advantage in the ability to vary underlying concepts in networks closer to UFR versus FER.
\item Finally, FER is relevant to \emph{learning} because learning involves moving through the weight space.
Learning is easier, and more likely to settle on deeper truths, if nearby/adjacent points in weight space respect fundamental regularities.
In contrast, if nearby points overwhelmingly break regularities, learning is stifled from building and elaborating on deep discoveries.
For example, if different modules perform arithmetic in the service of calculus versus physics, then learning a new, more efficient method to do arithmetic in the context of calculus would not apply to physics. The tax imposed by such fracture would likely compound even further in a continual learning scenario, which is one of the next frontiers for the field.
\end{enumerate}

Interestingly, studies from open-ended learning (as opposed to objective-driven learning) have revealed the ability to create neural networks with surprisingly little FER,~\citep{huizinga2018emergence, woolley2011deleterious}, even approaching UFR.
These observations are little known today, which is one reason they are highlighted explicitly in this paper.
In particular, we present, for the first time, side-by-side visualizations of the internal representation learned by networks trained via open-ended evolution versus conventional objective-driven SGD, on the task of outputting a single image.
Because the output is a single two-dimensional image, intermediate representations are easy to visualize and understand intuitively, revealing that FER within the internal representation can differ dramatically depending on the training procedure, despite identical output behaviors.

Because evidence of FER will inevitably raise the question of the right ``culprit'' to blame for its presence, we want to avoid the implication that all systems based on SGD must necessarily exhibit FER. While SGD-driven optimization in the experiments in this paper does yield FER, it is possible that novel variants or uses of SGD, such as in the service of novelty instead of a conventional specific objective, could exhibit much less of it. Therefore, we choose the term \emph{conventional SGD} to refer the application of SGD on a fixed architecture with a single target objective in the experiments in this paper.  It is also important to bear in mind that FER is likely to emerge in other settings as well (e.g.\ in second order methods and evolutionary algorithms), especially or most likely when optimizing to solve a single fixed objective, so the issue is almost certainly not exclusive to SGD \citep{lehman2010efficiently,woolley2011deleterious,stanley2015greatness,nguyen2015innovation,huizinga2018emergence}.

The observation of FER in conventional systems and the lack of it in open-ended systems at least merits a broader discussion in the hope of deeper insight into the genesis of robust representations.
In the long run, such discussion could lead to dramatically new approaches to large-scale training that produce better internal representations.
In this position paper, we do not propose a singular ``solution'' or a new algorithm.
Rather, our aim is solely to initiate this important discussion.
We anticipate a range of opinions on the significance of the observation of FER in conventional SGD-trained networks.
Some may question whether it matters at all while others will seek out new ways of observing and mitigating it.
All these contrasting perspectives are healthy and their ultimate resolution will come from further study.

\section{Background}
In contrast to benchmark tasks, where the outputs of deep networks are compared to an ideal reference point, the internal representations of such networks are often analyzed without reference to any ideal.
After all, in most domains, what viable alternative exists?

In this paper, we leverage the Picbreeder experiment~\citep{secretan2008picbreeder, secretan2011picbreeder}, an unconventional neural network search process, that offers a rare opportunity to examine how drastically different underlying neural representations can actually be.
As it turns out, Picbreeder consistently discovered neural representations that are radically different from those found in modern deep networks trained via SGD, albeit in a toy domain.
\textbf{Our aim is \emph{not} to argue that Picbreeder’s search process is superior to conventional training methods.
}The question of the best approach to learning good neural representations remains open.
\textbf{Rather, the value of our argument is to provide a novel point of comparison that hints at a new perspective on \emph{what could be possible}.
}
By examining the internal neural representations of Picbreeder networks and contrasting them with networks found from current conventional approaches, we open a new lens for rethinking what constitutes a good representation and how it can be achieved.

Accordingly, this section explains several of the underlying approaches that drove Picbreeder. While these techniques may be unfamiliar to some readers 
familiar with deep learning and large models, their differences will turn out important for understanding how the surprising representations in Picbreeder networks came about, and ultimately for helping to speculate about what changes to more conventional approaches might yield similar representations. 
In particular, we cover a special kind of neural network called a
compositional pattern-producing network (CPPN), 
the SGD-free neuroevolution of augmenting topologies (NEAT) algorithm that updates CPPNs, and how these components come together in Picbreeder.

\subsection{Compositional Pattern Producing Networks (CPPNs)}

\begin{figure}[t]
    \centering
    \includegraphics[width=0.7\textwidth]{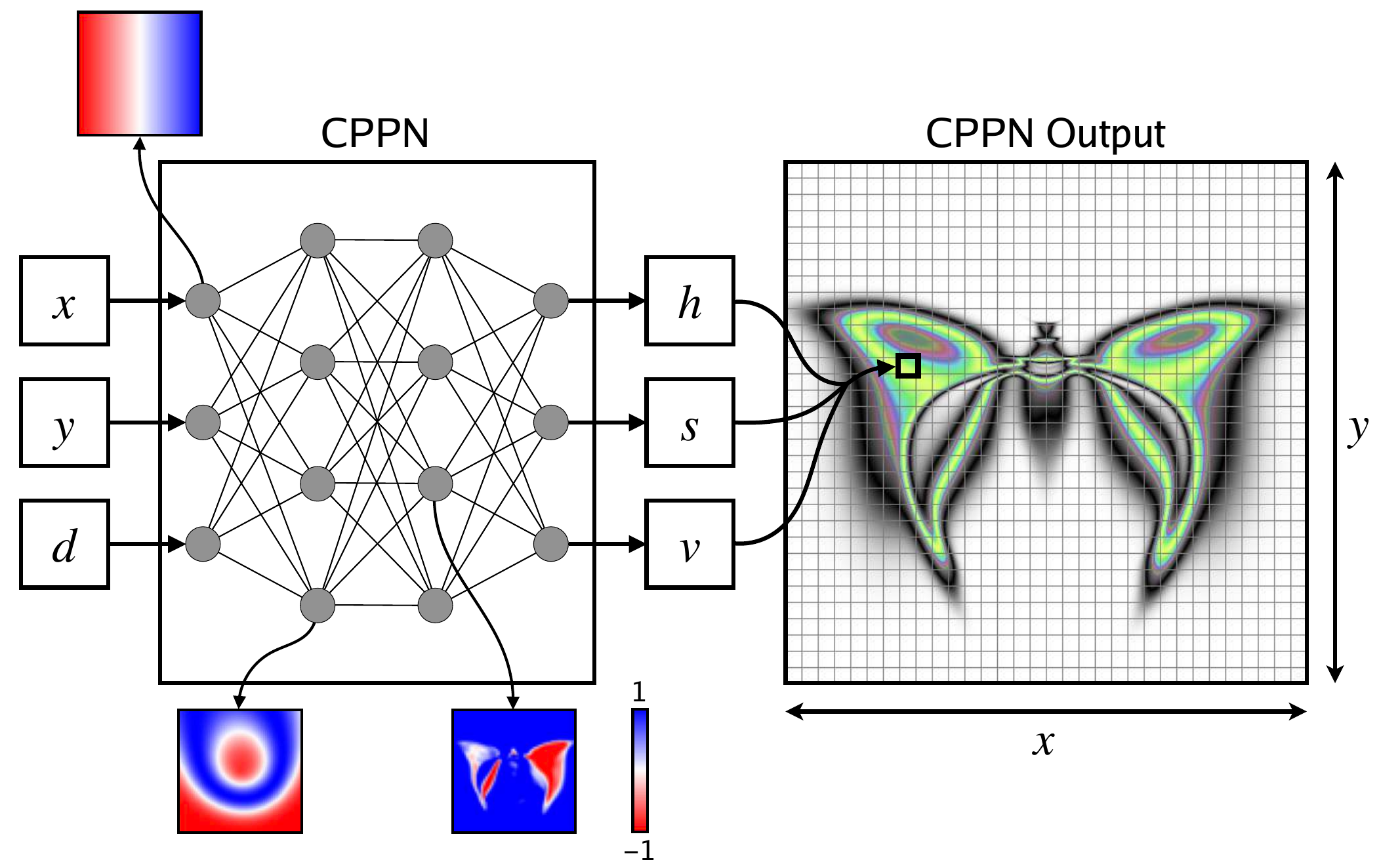}
    \caption{
    \textbf{Example compositional pattern producing network (CPPN)}. A CPPN is a neural network that takes as input the $x$, $y$, and $d=\sqrt{x^2+y^2}$ coordinates of a pixel and outputs the pixel's $h$, $s$, and $v$ color values.
    By sweeping over all possible $x$ and $y$ pixel locations, an image can be generated, visualizing the CPPN's output behavior for all inputs.
    Importantly, the activation of each hidden neuron over all inputs can be visualized as an image (shown in the red–blue insets, where red and blue indicates low and high activation), thus enabling inspection of how the CPPN internally represents and builds up its output behavior.
    In this paper, CPPNs serve as an analogy for large AI models like LLMs.
    }
    \label{fig:cppn}
\end{figure}

Originally inspired by the biological process through which organisms develop \citep{stanley2003taxonomy}, 
\emph{compositional pattern producing networks} (CPPNs) are a type of neural network that
is designed to encode structures with regularities and symmetries, such as images~\citep{stanley2007compositional}.
As shown in Figure~\ref{fig:cppn}, a CPPN takes as input the $x$, $y$, and $d$ (distance from center) coordinates of a pixel in a 2-D space and outputs its hue ($h$), saturation ($s$), and value ($v$)
color values.
By iterating over all 2-D pixel locations, the CPPN can generate a single image.

Due to this encoding, CPPNs can reuse their internal network structures at different pixel coordinates to create symmetric patterns (or any other type of regularity), much like how biological development reuses genetic information to form symmetric bodily structures in organisms (like arms).
This property allows CPPNs to generate visually coherent patterns with many regularities.

In effect, the image produced by a CPPN serves as a complete visualization of the network's output behavior for all possible inputs.
In contrast, fully visualizing the output behavior of an LLM is intractable due to the high-dimensionality of its inputs and outputs.
This ability to see the entire output behavior of CPPNs makes them useful for quickly understanding what exactly they are computing, and thus how they represent knowledge internally.
Even more helpful, because the inputs are just a grid sampling of a 2-D input space, the entire behavior of \emph{every neuron in the network} can be visualized in the same way (as an image), which means we can see precisely how the final output behavior is built up neuron by neuron underneath the hood.

The CPPNs in Picbreeder use unconventional activation functions---including identity, sine, cosine, tanh, sigmoid, and Gaussian---and mix them across neurons within a single CPPN.
This heterogeneous set of functions makes it easy to express different kinds of regularities; for example, symmetry across the y-axis can be captured by a single neuron computing $\text{Gaussian}(x)$, or a repeating pattern is captured by $\text{sin}(x)$.
More details on CPPNs are provided in Appendix~\ref{sec:details_cppn}.

Unlike networks trained through backpropagation, CPPNs are typically \emph{evolved} through the neuroevolution of augmenting topologies (NEAT) algorithm \citep{stanley2002evolving}. 
NEAT evolves not only the weights but also the topology (i.e.\ architecture) of the network, allowing complex networks to be grown over generations from smaller ones. When evolving CPPNs, NEAT also picks the activation function of each neuron. 

NEAT's CPPNs are essentially graphs of interconnected neurons.
By later building on the regularities learned by earlier CPPNs, NEAT can iteratively explore new architectures that increase expressiveness while preserving the foundational patterns of earlier CPPNs.
For example, if a new neuron is added downstream of a neuron with a symmetric activation that outputs a symmetric pattern, then the new neuron can also end up outputting a symmetric pattern because the output of its predecessor is itself symmetric.
More details on NEAT are provided in Appendix~\ref{sec:details_neat}.
While multifaceted motivations underpin the unique NEAT approach, its importance in this paper is mainly that it yields radically different kinds of internal representations when evolving CPPNs through the open-ended search in Picbreeder, providing us a point of comparison to more conventional methods that would otherwise not exist. (A more open-ended approach to SGD is also conceivable, but no such method is yet established.)

Note that one CPPN only generates one image, and thus is very different from modern generative image models~\citep{rombach2022high}.
The goal of CPPNs is not large scale, photorealistic image generation, but rather to introduce a new form of representation that can be used to explore complex structures with regularities and symmetries \citep{stanley2007compositional}.
To be clear, because the entire output space of the CPPN can be viewed as a single image, some readers may be confused into viewing the output space of the CPPN as a single output instance (e.g.\ like a single image output from a text-to-image model).
That would be a misunderstanding.
The output image of a CPPN is not an instance, but rather all the instances (all of the $h, s, v$ combinations) for the entire space of possible inputs (all of the $x, y, d$ combinations).
In that way, the image itself is effectively a ``micro-metaphor'' for everything an LLM knows, which is an approximation of all of human knowledge.
In effect, the entire space of outputs of the CPPN (for all possible inputs) is a rough analogy for the entire space of output responses of LLMs to all possible inputs.

Clearly the LLM space is much vaster, but the questions raised through CPPN visualizations in this paper are general and have implications at a much greater scale.

To facilitate comparisons to modern networks, we convert the NEAT CPPN---an arbitrary graph of neurons---into a computationally identical dense MLP CPPN through a process called \emph{layerization}.
In this new MLP, NEAT connections that evolved to skip several layers are explicitly represented as multiple identity connections to transport the information unchanged through the intermediate layers.
Each layer's neurons retain activation functions that replicate the original CPPN's computations. Importantly, layerization does not alter the computation itself but restructures it in a new architecture.
This step makes the architecture more amenable to the SGD-based search that will serve as a comparison, as discussed later.
More details on layerization are provided in Appendix~\ref{sec:picbreeder_porting}.
(Note that layerization is necessary to allow SGD to reproduce Picbreeder image outputs; as Appendix~\ref{sec:picbreeder_porting} also shows, SGD fails to reproduce the target image when run on the raw architectures that emerge from NEAT in Picbreeder.)

\subsection{Picbreeder}

Picbreeder was a website where humans could breed CPPN images the way one might breed dogs or horses \citep{secretan2008picbreeder, secretan2011picbreeder}.
Underneath the hood, Picbreeder used a variant of NEAT to evolve the CPPNs based on users’ selections.
In practice, the user would see 15 images and select one or more of them as parents for the next generation.

\begin{figure}[t]
    \centering
    \begin{subfigure}{\linewidth}
        \centering
        \includegraphics[width=1.0\linewidth]{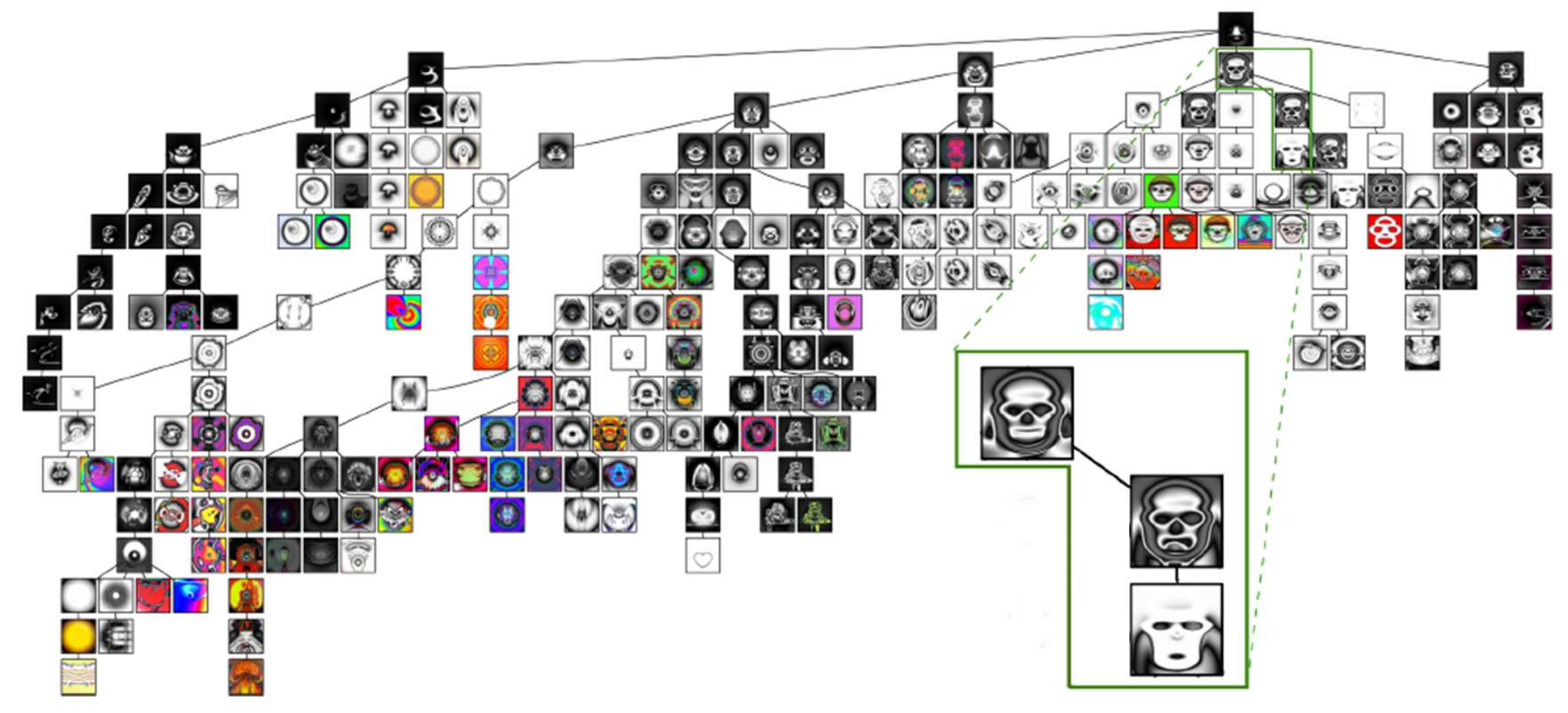}
        \caption{A subset of Picbreeder's phylogenetic tree}
        \label{fig:picbreeder_tree}
    \end{subfigure}
    \begin{subfigure}{\linewidth}
        \centering
        \includegraphics[width=1.0\linewidth]{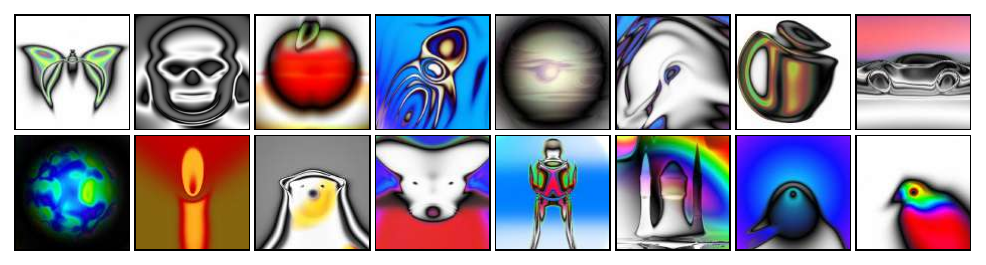}
        \caption{Notable CPPNs discovered by Picbreeder}
        \label{fig:picbreeder_imgs}
    \end{subfigure}
    \caption{
    \textbf{Picbreeder}. Picbreeder \citep{secretan2008picbreeder, secretan2011picbreeder} was a system that allowed humans to breed CPPNs based on their desires, using the NEAT evolutionary algorithm \citep{stanley2002evolving}.
    (a) This Picbreeder subphylogeny shows the open-ended nature of
    Picbreeder.
    It allowed users to keep generating interesting images from each other indefinitely.
    (b) Notable examples of CPPNs discovered by Picbreeder
    are in effect needles in a haystack in the space of all possible CPPNs.
    }
    \label{fig:picbreeder}
\end{figure}

Even though the vast majority of images in CPPN-space are uninterpretable blobs, 
Picbreeder users discovered thousands of interesting images (Figure~\ref{fig:picbreeder} gives a sampling).
This success can be attributed to the nature of the human-guided Picbreeder search process with CPPNs, which facilitates efficient exploration~\citep{stanley2015greatness, stanley2017open, secretan2008picbreeder}.
Picbreeder does not set a specific objective but instead encourages users to follow paths they find interesting.

This lack of a singular unifying objective enabled Picbreeder to remain open-ended, diverging and generating novel images indefinitely
(Figure~\ref{fig:picbreeder_tree}).
The search also embraced serendipity, as users were often not fixated on evolving a specific target image.
For example, if mutating an ``egg'' image led to a ``teapot,'' users could easily pivot to pursue this new direction.
This kind of serendipitous search echoes a similar open-ended dynamic in natural evolution and scientific discovery~\citep{stanley2015greatness}.
Later experiments showed that encouraging such ``goal switching'' and otherwise capturing principles of open-endedness in algorithms improves performance and the discovery of high-quality, diverse solutions (\cite{lehman2011abandoning, mouret2015illuminating, wang2019paired, wang2020enhanced, zhang2023omni, cully2015robots, ecoffet2021first}.

Furthermore, it was later found that Picbreeder facilitates the evolution of evolvability~\citep{huizinga2018emergence}, a core principle of natural evolution~\citep{dawkins1996blind, dawkins2019evolution, kirschner1998evolvability, wagner1996perspective}.
By structuring the CPPNs' internal representation with the right axes of variation, the CPPNs were able to better adapt in ways that align with human preferences while remaining resilient to deleterious mutations~\citep{huizinga2018emergence, lehman2018more}. Systematic studies by \cite{huizinga2018emergence}, including via an open-source software tool called CPPN-Explorer (CPPN-X), showed that it was actually \emph{commonplace} in CPPN genomes to find representations that were factored and respected key regularities (e.g.\ where you could separately control different dimensions of variation that humans naturally would wish to control, like the size of an apple or, separately, the size or angle of its stem).

Picbreeder thus generated CPPNs with structured representations that captured key visual regularities of the objects they depicted.
In essence, the Picbreeder CPPNs feature \textit{unified factored representations}, which are presented next.

\section{Unified Factored Representations (UFR)}

Representations are often analyzed in isolation because it is rare to have functionally equivalent artifacts that nevertheless have radically different representations.
Interestingly, given an image from Picbreeder, it is straightforward to perform supervised learning to predict ($h$, $s$, $v$) from ($x$, $y$, $z$).
As a result, we can compare Picbreeder-evolved CPPNs (which prior research such as \citealt{huizinga2018emergence} has already shown to have compelling representations) directly to those produced by deep learning.
Enabling such a comparison in the image domain gives us a toy model that still has depth, where we can visually inspect the whole representational structure. 

\begin{figure}[t]
    \centering
    \begin{subfigure}{0.25\linewidth}
        \centering
        \includegraphics[width=\linewidth]{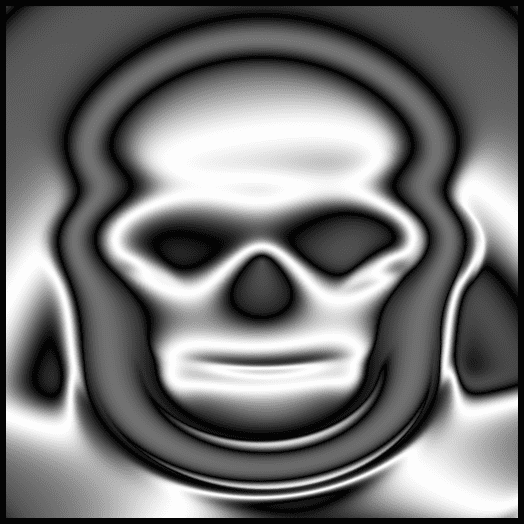}
        \caption{Picbreeder CPPN}
        \label{fig:img_576_pb}
    \end{subfigure}
    \hspace{1cm}
    \begin{subfigure}{0.25\linewidth}
        \centering
        \includegraphics[width=\linewidth]{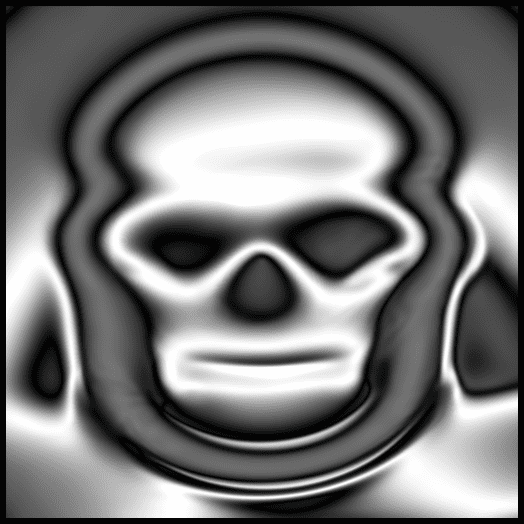}
        \caption{Conventional SGD CPPN}
        \label{fig:img_576_sgd_pb}
    \end{subfigure}
    \caption{
    \textbf{The Picbreeder skull CPPN and its conventional SGD equivalent.}
    The image on the left (a) is the original discovery of the Picbreeder skull, represented by an evolved CPPN. On the right (b), the skull produced by SGD when trained to output the Picbreeder skull is effectively identical, showing that SGD can easily learn the same behavior output as training targets.
    However, despite their outward similarity, their internal representations are radically different (as discussed next in the main text).
    Appendix \ref{sec:relucppn} shows that SGD results also hold for ReLU-only architectures.}
    \label{fig:img_576}
\end{figure}

For these reasons, this section focuses on the \emph{Picbreeder Skull CPPN}, shown in Figure~\ref{fig:img_576_pb}, as a canonical example of a \emph{Unified Factored Representation} (UFR).
Appendix~\ref{sec:other_picbreeder_examples} similarly analyzes two other examples from Picbreeder in depth.

To provide the conventional SGD-based comparison, we trained a CPPN with SGD to replicate the same skull that was discovered in Picbeeder (Figure~\ref{fig:img_576_sgd_pb}).  This reproduction looks identical because SGD succeeds at reproducing the output.
In more detail, beginning with the same neural architecture as the layerized version of the Picbreeder CPPN, we initialize a new random set of weights and train them with SGD to match the Picbreeder CPPN’s output.
We provide each $(x, y, d)$ instance and its corresponding $(h, s, v)$ label as a regression target.
As an (imperfect) analogy, the Picbreeder CPPN is like a preexisting ``human'' brain, whereas this new conventional SGD CPPN is analogous to an LLM trained to mimic the brain's output behavior for a vast set of inputs (analogous to current LLM internet training).
Complete details of the SGD training process are in Appendix~\ref{sec:sgd_details}.

Notice that the skull is symmetric, even if not perfectly so.
This observation raises the question: does the CPPN that generates this skull ``know'' that it is symmetric?
In other words, does the CPPN explicitly represent this symmetry internally (in its neural circuits), i.e.\ in a unified way such that the two sides of the skull are not fractured into separate chains of computation?

One possible response is that \textit{it does not actually matter} whether the network ``knows'' about the symmetry, as long as the output \textit{appears symmetric}.
After all, if the skull looks correct, why should it matter how the network processes or understands its symmetry?

This line of reasoning is relevant because one could ask the same question about LLMs.
That is, is it fair to say that, as long as an LLM answers your questions well, then there is no reason to worry about the form of its underlying representation?
Just like the skull, it might be interesting to know its inner workings, but it may be that the only thing that matters is that the system functions well.

The question of whether the underlying representation truly matters is a recurring theme in this paper that will be explored in depth.
One key initial insight is that, as it turns out, the CPPN behind the Picbreeder skull in Figure~\ref{fig:img_576_pb} \emph{does know} that it is symmetric.
That is, there is a lot of evidence to support the idea that this CPPN fundamentally understands and represents the skull's symmetry.

More specifically, for the neural network to know about the symmetry, both the left and right sides of the skull, despite being distinct, should be represented mainly by the same information within the weights of the CPPN.
The pattern you see on the left side is nearly identical to the pattern on the right side, which means that the information within the neural network that encodes that pattern (which is within its connection weights) should only be expressed once.  

If this information is only expressed once, then it can be \textit{reused} to generate both sides.
This concept of reuse recalls similar reuse that occurs within DNA, where the information encoding one side of a symmetric body is primarily represented once and then reused during embryonic development to form the full structure~\citep{wolpert2015principles}.
This similarity between biological and neural representations highlights that the principle we observe is not exclusive to neural networks: regardless of the encoding mechanism, whether DNA or neurons, good representation follows similar principles.

Of course, reuse does not necessarily imply perfect symmetry~\citep{stanley2007compositional}.
Complex patterns of all types, whether in biology, poetry, or visual art, often lack perfect symmetries and regularities.
If you look closely at the skull in Figure~\ref{fig:img_576_pb}, you will notice that the left and right sides are not perfectly symmetric.
The important point is that such imperfect symmetry, yet symmetry nonetheless, is normal in intricate structures.
Consider that the human body is largely symmetric, though the heart is on only one side.
Yet the fact that the heart is on one side does not mean that the DNA that encodes the left side of the body is disjoint from that which encodes the right side.
Rather, it simply means that the encoding method is nuanced.
Even so, there is still significant reuse from one imperfectly symmetric side to the other~\citep{wolpert2015principles}.
The fingers of the hand are another good example of regularity with variation.

\begin{figure}[p]
    \centering
    \begin{subfigure}{0.92\textwidth}
        \centering
        \includegraphics[width=\textwidth]{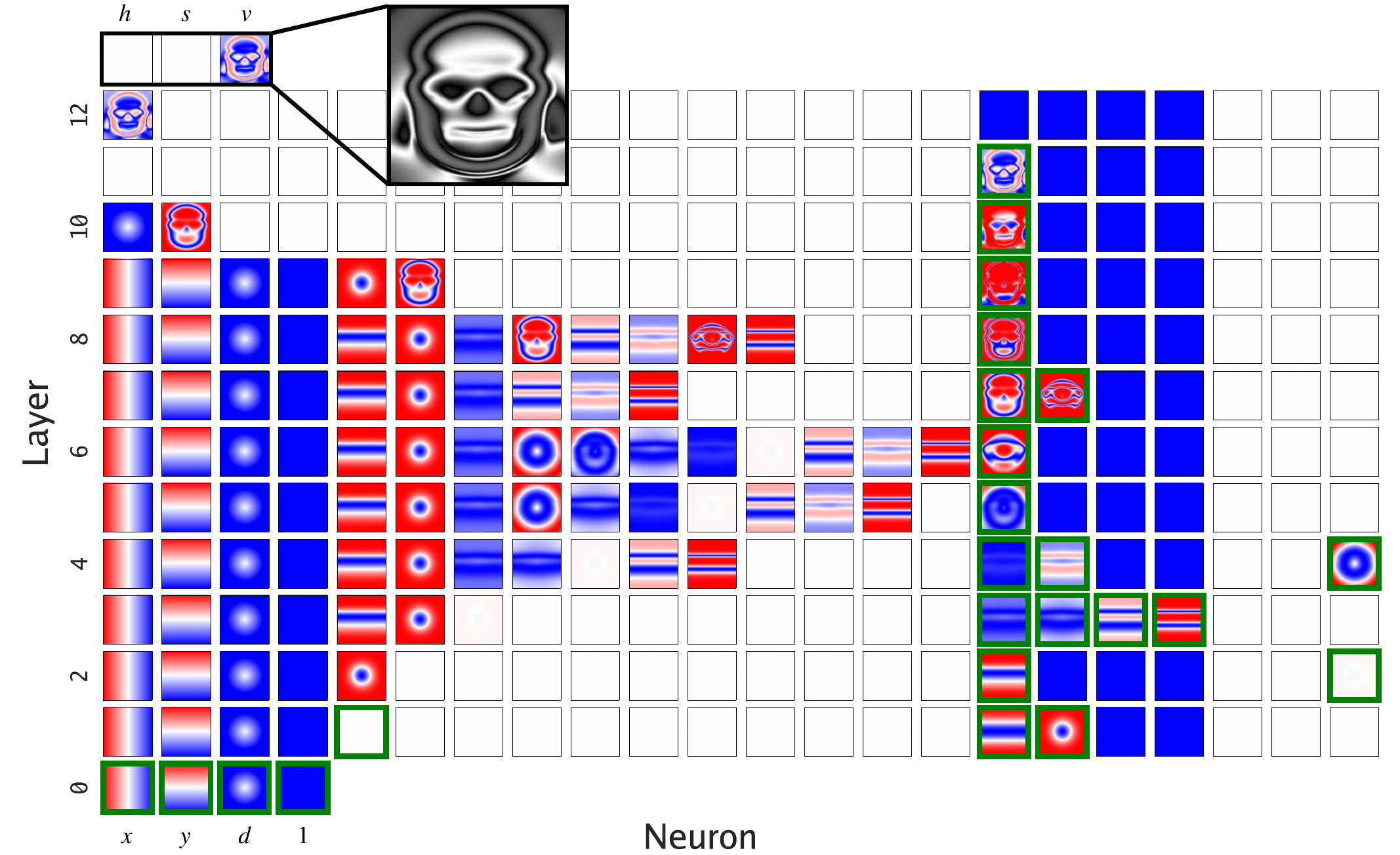} %
        \caption{Picbreeder CPPN}
        \label{fig:fmaps_576_pb}
    \end{subfigure}
    \vspace{0.0cm}
    \begin{subfigure}{0.92\textwidth}
        \centering
        \includegraphics[width=\textwidth]{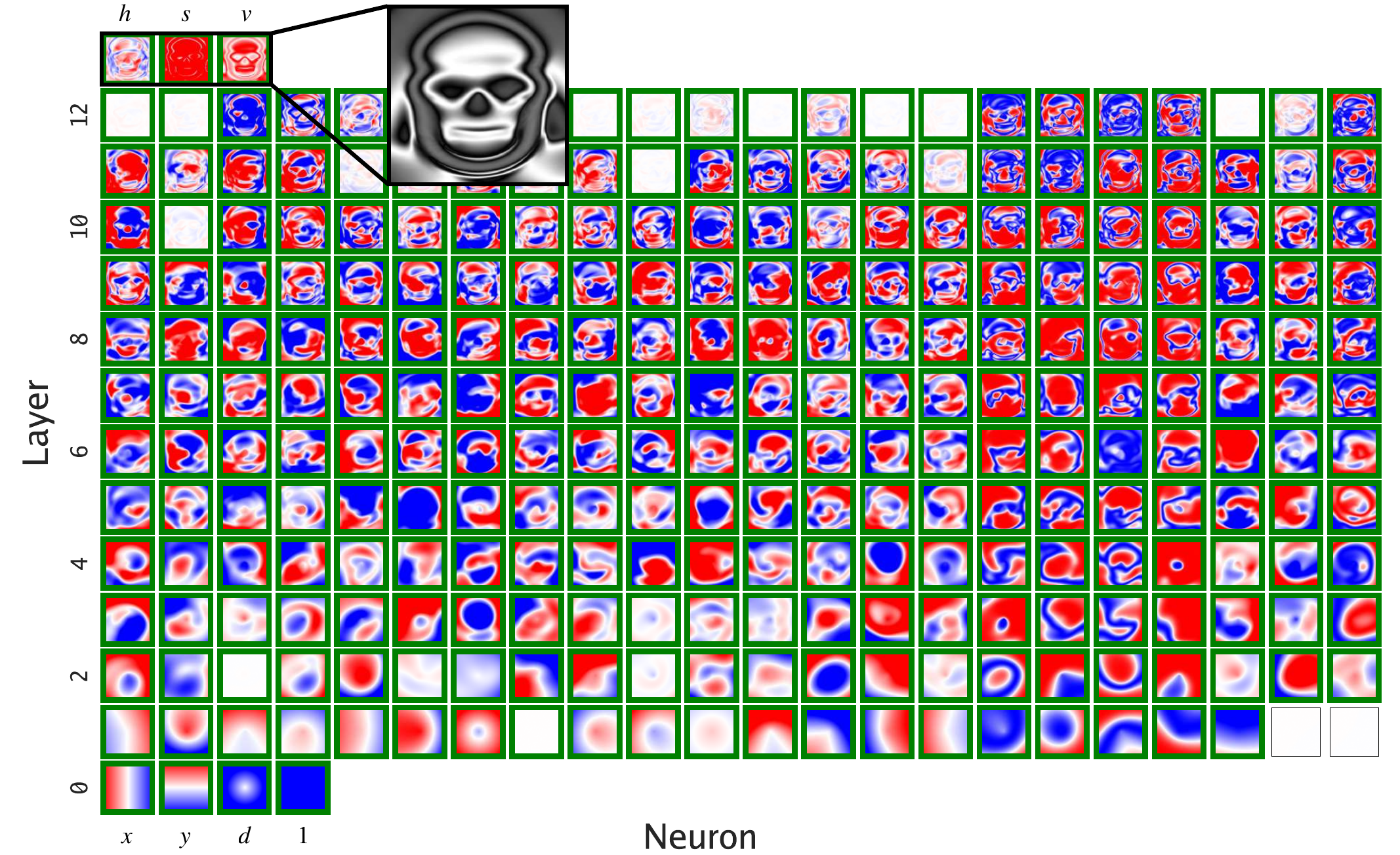} %
        \caption{Conventional SGD CPPN}
        \label{fig:fmaps_576_sgd_pb}
    \end{subfigure}
    \caption{
    \textbf{Internal representations of the Picbreeder and conventional SGD skull CPPNs.}
    This figure illustrates how each CPPN processes its input through its neural circuitry to generate its output.
    Each image visualizes the latent representation of all inputs at a specific layer and neuron (red and blue represent low and high activation, respectively).
    Notice the stark contrast between the Picbreeder CPPN (a), which represents the skull through organized circuitry, and the conventional SGD CPPN (b), which represents \textit{the same skull} through disorganized patchwork.
    A green border indicates the latent representation is novel (not seen in previous layers).
    The layerized version of the Picbreeder CPPN includes only 24 novel representations, highlighting the efficiency of its encoding.
    }
    \label{fig:fmaps_576}
\end{figure}

The most explicit evidence that the CPPN encodes symmetry is found through a direct analysis of the behavior of each individual neuron in the CPPN and its contribution to the output behavior, as shown in Figure~\ref{fig:fmaps_576_pb}.
(Figure~\ref{fig:fmaps_576_sgd_pb} shows the same intermediate representations in the conventional SGD version of the skull for contrast; these will be discussed more later.)

This visualization reveals \textit{the way} in which the neurons in the neural network progressively build up their output behavior, layer by layer and neuron by neuron.
The representation in Figure~\ref{fig:fmaps_576_pb} is highly organized, with symmetry across the y-axis emerging early and persisting through the layers.
Key features of the skull appear exactly once and influence the entire skull as a whole, rather than affecting individual parts separately, thus forming a UFR.
Moreover, the skull representation suggests an implicit factorized decomposition of its semantic structure.
For example, distinct neurons correspond to the overall shape of the skull, as well as to specific features such as the eyes, nose, and mouth.
This kind of decomposition and reuse of information implies a form a representation more akin to understanding (e.g.\ like understanding how to multiply two numbers) as opposed to memorization (e.g.\ memorizing the answer to many multiplication problems). (In contrast, Figure~\ref{fig:fmaps_576_sgd_pb}, discussed later, shows the significantly more disorganized representation learned by conventional SGD.)

While the evidence for UFR is striking, the deeper question remains: does it actually matter?
Why should we care whether the sides of the skull involve reuse under the hood or whether semantically intuitive components of the skull are factored in its representation if the skull looks correct anyway?
The key insight, which again connects back to LLMs, is that underlying representation \emph{is not about capturing a particular pattern or output behavior}.
Rather, it is about whether the system can \emph{build on the regularities of that representation to learn and generate
new behaviors}.
These implications ultimately play out in generalization, creativity, and continual learning.

\begin{figure}[t]
    \centering
    \begin{subfigure}{0.49\linewidth}
        \centering
        \includegraphics[width=\linewidth]{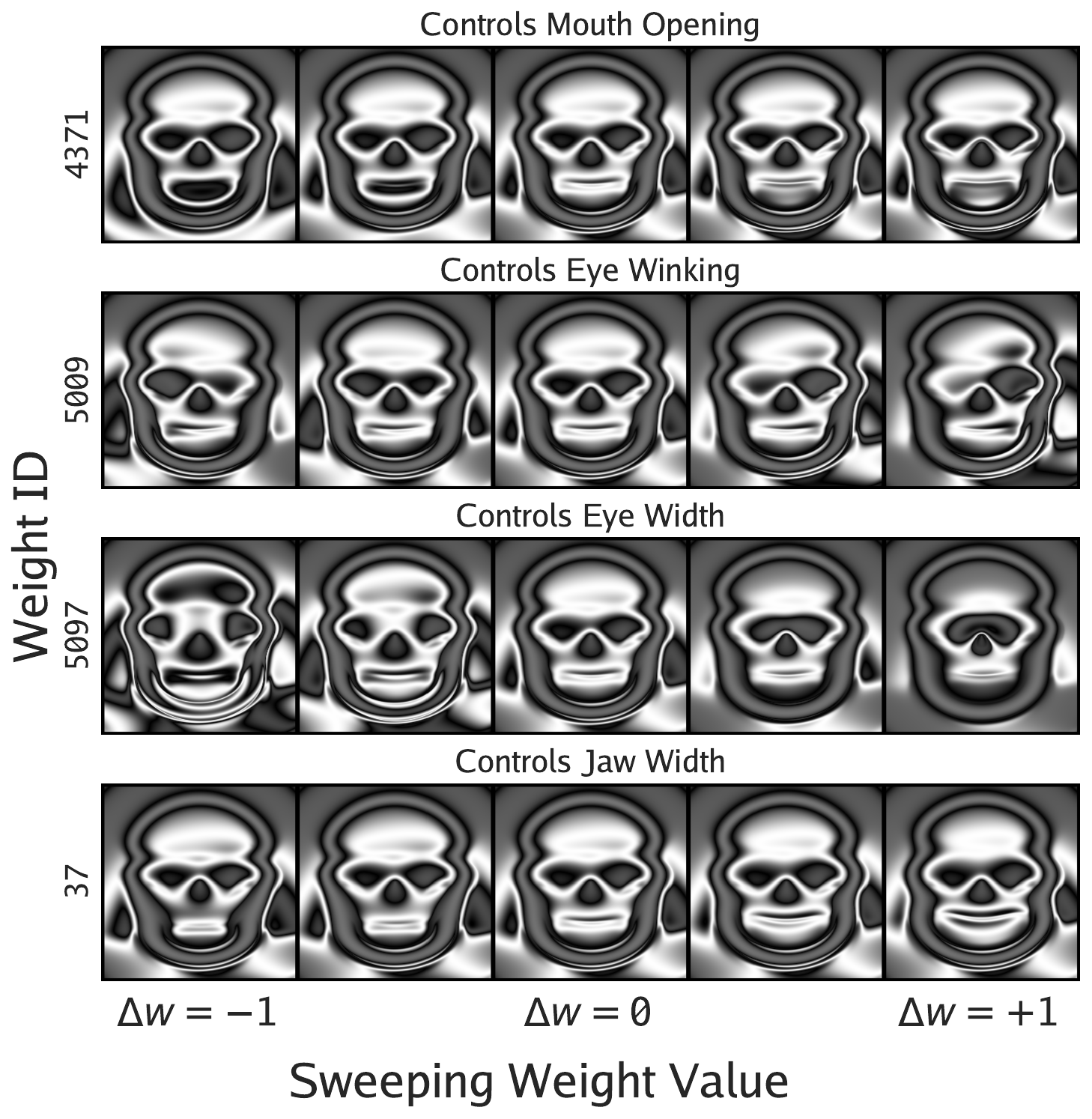}
        \caption{Picbreeder CPPN}
        \label{fig:weight_sweeps_576_pb}
    \end{subfigure}
    \hspace{0cm}
    \begin{subfigure}{0.49\linewidth}
        \centering
        \includegraphics[width=\linewidth]{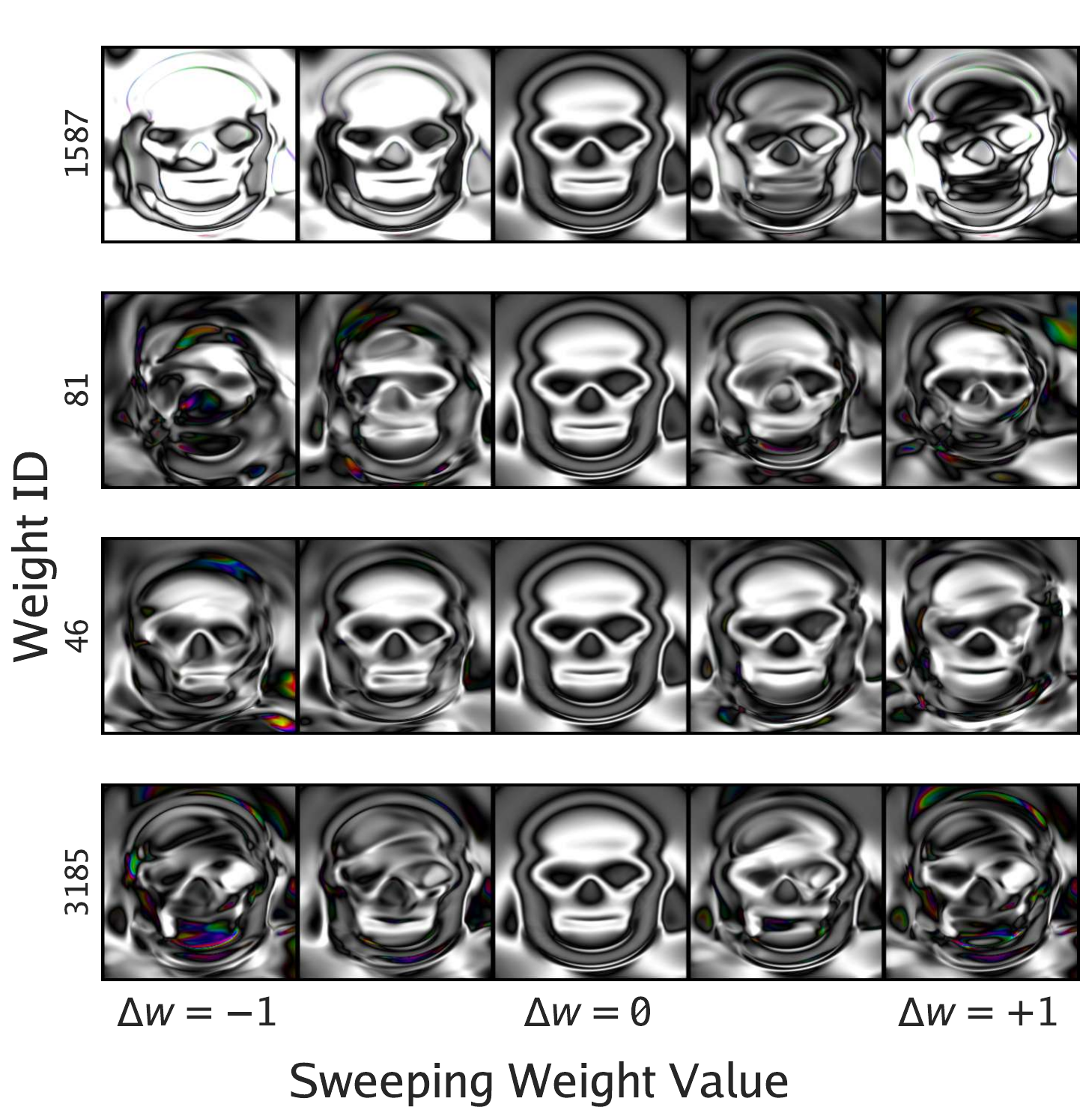}
        \caption{Conventional SGD CPPN}
        \label{fig:weight_sweeps_576_sgd_pb}
    \end{subfigure}
    \caption{
    \textbf{Select weight sweeps of the Picbreeder and conventional SGD skull CPPNs.}
    These visualizations show how sweeping individual weights of the Picbreeder and SGD CPPNs affects their output, that is, it shows how the overall behavior of the network is impacted by a single weight.
    In the Picbreeder CPPN (a), remarkably, these specific weights have meaningful semantic control, aligning with human-interpretable changes to the skull.
    In contrast, sweeping the weights in the SGD CPPN (b) leads to meaningless distortions of the skull.
   All weights in the SGD CPPN lead to similarly meaningless perturbations.
    This property shows the critical importance of the difference in the underlying representations that are seen in Figure \ref{fig:fmaps_576} of these two CPPNs. (\citealt{huizinga2018emergence} first showed similar weight sweeps of Picbreeder CPPNs in a study of canalization in Picbreeder.)
    }
    \label{fig:weight_sweeps_576}
\end{figure}

For example, if a model fundamentally represents that skulls are generally symmetric, then it can in principle imagine new skulls or faces with new expressions (perhaps based on perturbations to the original skull).
In this way, it would implicitly understand that modifications to one side are reflected in the other.
Accordingly, there is evidence of such reflection:
Sweeping through the range of \emph{individual weights} in the CPPN yields orderly and meaningful changes that preserve the overall ``skull-ness'' of the pattern, as shown in Figure~\ref{fig:weight_sweeps_576_pb}.\footnote{The ``winking'' sweep in row 2 of Figure~\ref{fig:weight_sweeps_576_pb} disrupts overall symmetry, but does so in an orderly and coherent manner that preserves most of it.} These sweeps also highlight remarkable modular decomposition (disentangling key semantic factors), such as independent control of the vertical width of the skull's mouth.
(For the sweeps shown in this paper, we sought to select the weight sweeps for both methods that present them in their best light, while ensuring that the chosen sweeps are broadly representative of the behavior of the majority of the sweeps.  The database of sweeps at the  \href{\codelink}{code link} provides full transparency by disclosing all possible sweeps. Sweeps for other Picbreeder CPPNs are in Appendix \ref{sec:other_picbreeder_examples}.)

However, if the model does not fundamentally represent 
that the skull is symmetric (even though it can draw a single nice looking skull) the opposite will happen: almost any random perturbation would break the symmetry or disorganize the image in a chaotic and entangled manner (as occurs with the skull produced by conventional SGD, Figure~\ref{fig:weight_sweeps_576_sgd_pb}). 
In effect, this kind of representational pathology is a failure of representing symmetry and other regularities that will cripple imagination, creativity, and learning: if you do not understand that faces are symmetric, even if you memorize a single face and draw it to perfection, you will have little ability to conceive of any other face or take meaningful creative leaps.
An understanding of the underlying symmetry is crucial.

Based on these arguments, at least in the case of the skull, it may matter
how the skull is represented even if the output image is perfect at every pixel location.
Accordingly, the next section argues that such dramatic differences in representation have broader implications for the adaptability of conventional SGD's products more generally.

\section{Fractured Entangled Representations (FER)}

When we feed data into a machine learning model, the hope is that it will ultimately discover and encode the regularities that exist in the data.
If there are enough instances of a problem type then the chance of learning its regularities increases.
However, in cases where sufficient data are unavailable, such as in cutting-edge domains, or where there are so many combinations of elements that only a small subset can be represented in the data, the model may struggle to identify and represent underlying regularities.
In the world of the skull, all of the data points are subject to the globally symmetric regularity of the pattern.  So the question is, what is the chance that a learning model will ``notice'' an underlying regularity like symmetry?
The evidence from conventional SGD-based networks such as in Figure \ref{fig:fmaps_576_sgd_pb} hints that the chances are low.

As Figure~\ref{fig:img_576_sgd_pb} showed, after many SGD training iterations, this SGD-based CPPN is able to match the Picbreeder skull exactly.
However, when visualizing the internal representation of this conventional SGD CPPN in Figure~\ref{fig:fmaps_576_sgd_pb}, it is apparent this CPPN never undstandstands the underlying symmetry of the skull.
For example, the SGD CPPN seems never to encode the two sides of the image as symmetric until the very final (output) layer, failing to take advantage of this regularity throughout all 12 underlying layers.  
The result is a highly entangled patchwork,
with far more novel patterns than necessary, hinting at FER.

The underlying pathology of FER is also easily observed simply by sweeping the conventional SGD CPPN's weights, as shown in Figure~\ref{fig:weight_sweeps_576_sgd_pb}.
In the case of the Picbreeder CPPN we saw that the underlying, overall symmetry is usually preserved under perturbation (and when violated, as in winking, that occurs in a regular, smooth, natural way).
Yet in the SGD CPPN, though it appears identical, symmetry is almost always broken incoherently---the opposite behavior.  

One potential reason to question this conclusion is the possibility that the feature space is simply being viewed in the wrong vector basis, and that a simple rotation might reveal more unified, holistic features.
However, this scenario appears unlikely given the high degree of disorder in the conventional SGD skull representation---particularly the presence of arbitrary high-frequency patterns that bear little resemblance to the skull itself.
Additional empirical evidence addressing this concern is showcased in Appendix~\ref{sec:pca_feature_space}, where PCA helps to visualize an alternative feature basis and weight sweeps agnostic to the choice of basis continue to show evidence of FER (while as expected, the Picbreeder skull maintains meaningful variations even along random directions in weight space).

This problem is deeper than just inefficiency: it limits the ability of the model to build anything new (like a new skull or new face) that would require an understanding of faces.
Even though it draws a perfect skull, it does not understand the underlying regularities or any modular decomposition of what it is drawing at all.
Recall that more examples are given in Appendix~\ref{sec:other_picbreeder_examples}, which also show the same general FER phenomenon.
Appendix~\ref{sec:relucppn} shows that FER persists even when doing SGD with standard ReLU networks.

In effect the learned skull is an \emph{imposter}---its external appearance implies its underlying representation should be authentic, but it is not the real thing underneath the hood.

\section{Imposter Intelligence}

We often think of individual images generated by large models as a single instance of generation.
For that reason, it is important to recall that, though they appear superficially similar, the CPPN images in this paper are \textit{not} analogous to a single large model generation step.
Instead, these CPPN images are analogous to an entire \emph{space} of inputs and outputs: each $(x,y,d)$ pixel coordinate is an input and the corresponding pixel $(h,s,v)$ is the output.
In this way, CPPN images are actually a metaphor for the entire behavior and intelligence of a neural network (analogous to a comprehensive intelligence and personality profile/snapshot of a neural network, such as an LLM).

In that context, the imposter skull is a simple metaphor for a vastly more complex ``imposter intelligence.''
It shows the danger of judging a book by its cover, except here, the ``cover'' is the sum total of all its behaviors.
No matter what input we give the imposter skull CPPN, it outputs an unassailable match to our hopes and expectations, even though underneath the hood it looks nothing like it should.  

Now is therefore a good time to revisit the question of whether it matters how information is represented internally as long as a model works well.
In effect, the deeper question is whether an LLM (or any foundation model) can be analogous to the skull.
Can it appear healthy on the surface (where ``the surface'' is now an almost inconceivably vast span of abilities) but beneath the hood be suffering from pervasive FER?  And does that even matter?

First, note that it is at least theoretically possible to have FER in a large model: it simply means that in cases where the same information could have been reused effectively, instead it was not reused and the underlying principle is represented twice or more. Furthermore, capabilities that should be separate in principle may be entangled in subtle ways.
For example, if the computational mechanism involved in counting bricks was fundamentally disjoint from the mechanism for counting apples, 
and the nature of the objects being counted interfered with counting itself,
the concept of counting would exhibit FER.

The skull's simplicity makes it a useful conceptual model for understanding this idea: in the skull, we can literally see the symmetry in the internal representations of the pattern and how outputs respond to weight perturbations, %
whereas through conventional SGD it usually ends up redundantly fractured and entangled in the sense that there are two uncorrelated yet entangled pathways to representing both sides of the skull (which is evident through the weight perturbations in Figure~\ref{fig:weight_sweeps_576_sgd_pb}).

To transfer these observations and intuitions to an LLM, we will draw an analogy between the symmetry of the skull and the vast amount of \textit{conceptual} regularities that occur throughout the breadth of human faculties.

\subsection{Evidence of Imposter Intelligence}

Finding evidence of FER in LLMs and other networks requires focusing on unusual distinctions unlike those found in conventional benchmarks.
Nevertheless, there are many examples of such phenomena, which will be showcased in this section.

That said, some of the hints of FER can become subtle or even disappear as models scale up~\citep{bubeck2023sparks}.
Importantly, that does not necessarily mean that the internal FER itself has disappeared: as the skull shows, a model can achieve total and perfect coverage of a domain even if it is fractured underneath.
However, it does mean that behavioral evidence of FER (i.e.\ without the need to look underneath the hood) is more difficult to find in more recent models.
For this reason, we will start by looking back at GPT-3 \citep{brown2020language} for the first example.

It turns out that GPT-3's ability to perform arithmetic depended inappropriately upon what kinds of objects are being summed:

\colorlet{color_user}{PineGreen}
\colorlet{color_chatgpt}{BrickRed}
\colorlet{color_metadata}{NavyBlue}

\begin{tcolorbox}[enhanced jigsaw,breakable,pad at break*=1mm, colback=black!5!white,colframe=black!75!black,title=GPT-3 on Counting Items]
\label{box:counting}

\textcolor{color_user}{\textbf{User:}}

I have 3 pencils, 2 pens, and 4 erasers.  How many things do I have?

\textcolor{color_chatgpt}{\textbf{GPT-3:}}

You have 9 things. \textcolor{color_metadata}{\textbf{[correct in 3 out of 3 trials]}}

\textcolor{color_user}{\textbf{User:}}

I have 3 chickens, 2 ducks, and 4 geese.  How many things do I have? 

\textcolor{color_chatgpt}{\textbf{GPT-3:}}

You have 10 animals total. \textcolor{color_metadata}{\textbf{[incorrect in 3 out of 3 trials]}}
\end{tcolorbox}

Of course, the idea of summing should be independent of the type of object, and thus the same neural algorithm should be invoked regardless of which objects appear in the context.
The fact that GPT-3 fails this test demonstrates that it uses different neural algorithms in different contexts (like how different circuits are used for the left and right sides of the symmetric skull), providing clear evidence of FER.

Interestingly, later models like GPT-4 no longer exhibit this specific behavior. 
When evidence of FER in a particular context disappears with scale or further training, the nature of the rectification is not straightforward to determine and will likely require further study. 
For example, there could simply be better coverage, but still with significant FER under the hood (like with the conventional SGD skull).  Alternatively, the model might somehow unify into UFR, maybe through a process like grokking~\citep{power2022grokking,liu2022towards}, but precisely when or how that might happen is not currently known. 
Even if FER reduces where coverage is high, if FER is an inevitable side effect of the current training paradigm, then there will likely still be hints of FER somewhere, especially at the frontiers of knowledge or anywhere where data coverage is more sparse.

Indeed, other works have found more evidence of FER (though without referring to it with the term ``FER'').\footnote{\citet{mitchell2025llms} consolidated a useful recent list of such works, from which some of the references here are drawn.}
For example, with small datasets, models often find spurious features
that solve the task, but rely on heuristics that are not true to the problem, which is called ``shortcut learning''~\citep{geirhos2020shortcut}.
Concerningly, there is evidence of this phenomenon still happening even with much larger data distributions.

One study trained a transformer model on a massive synthetic dataset of valid moves from the Othello board game~\citep{li2023emergent}.
As a result, from only predicting move sequences, the model implicitly learned an emergent world model within its features by encoding the state of the board~\citep{li2023emergent, nanda2023actually}.
However, further investigation found that the model did not learn the rules of the game as a human would (with unified factored rules that apply to all situations), but rather achieved this feat by learning a big bag of fractured entangled local heuristics~\citep{mitchell2025llms, jylin2024othellogpt}.

The same result has been found in the domain of modular arithmetic.
By analyzing LLMs at a mechanistic level, it is apparent that even when LLMs master arithmetic, they often do so using a collection of heuristics (such as only applying a certain rule within a specific range) rather than the correct unified arithmetic algorithm underlying the task~\citep{nikankin2024arithmetic}.
For example, the Claude model was adding 36 and 59 in part based on the heuristic that ``around 40 + around 50 is around 92''~\citep{lindsey2024biology}.

This bag of heuristic solutions may explain why the world models learned by current networks often exhibit incoherence and fragility~\citep{vafa2024evaluating}.
For example, one study found that while LLMs reason well in standard scenarios, they struggle significantly in counterfactual ones in which the LLM has to reason about what would happen under less common conditions (such as arithmetic in base 9 instead of base 10), which suggests that the right kind of reasoning only emerges in some contexts but not others~\citep{wu2024reasoning, han2025general}.
This result suggests that LLMs may rely on FER heuristics that are overfit to common situations rather than learning fundamental UFRs that generalize to counterfactual contexts.
Some models are so brittle (and overfit) that even simply changing the numbers in a popular gradeschool math benchmark~\citep{cobbe2021training} results in a performance drop~\citep{mirzadeh2024gsm}.

Returning to the Othello model, one of the typical heuristics it learned exhibits its tendency towards over-specificity for one specific position when instead it should be learning general principles that can apply to many potential positions on the board:
``If the move A4 was just played AND B4 is occupied AND C4 is occupied, then update B4+C4+D4 to `theirs'{\hspace{0.02in}}''~\citep{jylin2024othellogpt}.
If a model does indeed learn heuristics like this one, but for all conceptual knowledge on the internet, how can it build on such a shaky foundation to create new ideas about the world, or to learn new topics from few experiences?

More recent models can still exhibit signs of FER in relatively simple scenarios.  For example, GPT-4o (April 2025) can easily replace numbers at specific positions in a sequence of numbers.  However, it cannot perform the identical task of replacing words at specific positions in a sequence of words, hinting at FER by failing to apply the same algorithm in different contexts:

\begin{tcolorbox}[enhanced jigsaw,breakable,pad at break*=1mm, colback=black!5!white,colframe=black!75!black,title=GPT-4o on Number Sequences: \textbf{correct in 3 out of 3 responses}\\(only the numerical part of each response is shown)]
\textcolor{color_user}{\textbf{User:}}\\

Replace the 1st, 3rd, 5th and and 7th number in this sentence with different numbers (do not change any other numbers): 1 5 8 3 12 14 2\\

\textcolor{color_chatgpt}{\textbf{GPT-4o Response 1}}

9 5 7 3 11 14 6 \textcolor{color_metadata}{\textbf{[correct]}}\\

\textcolor{color_chatgpt}{\textbf{GPT-4o Response 2}}

7 5 11 3 9 14 6 \textcolor{color_metadata}{\textbf{[correct]}}\\

\textcolor{color_chatgpt}{\textbf{GPT-4o Response 3}}

7 5 9 3 11 14 6 \textcolor{color_metadata}{\textbf{[correct]}}

\end{tcolorbox}

Now, the task is changed to sequences made of words instead of numbers.
Notice that the prompt also uses almost identical language:

\begin{tcolorbox}[enhanced jigsaw,breakable,pad at break*=1mm, colback=black!5!white,colframe=black!75!black,title=GPT-4o on Word Sequences: \textbf{incorrect in 3 out of 3 responses}\\(only the word sequence part of each response is shown)]
\textcolor{color_user}{\textbf{User:}}\\

Replace the 1st, 3rd, 5th and and 7th word in this sentence with different words (do not change any other words): I am coming for tomorrow that way.\\

\textcolor{color_chatgpt}{\textbf{GPT-4o Response 1}}

You were coming for dinner that night way.
\textcolor{color_metadata}{\textbf{[second word wrong, too many words]}}\\

\textcolor{color_chatgpt}{\textbf{GPT-4o Response 2}}

You are leaving for today that evening.
\textcolor{color_metadata}{\textbf{[second word wrong]}}\\

\textcolor{color_chatgpt}{\textbf{GPT-4o Response 3}}

We are leaving for today that night.
\textcolor{color_metadata}{\textbf{[second word wrong]}}

\end{tcolorbox}

Notice that inappropriate substitution of ``were'' or ``are'' for ``am'' also hints at entanglement between the grammatical ability of the network and its ability to follow instructions that are agnostic to grammar.
Modern LLMs have a strong preference for answers that are statistically more likely in the training data~\citep{mccoy2023embers}, hurting performance when this assumption might be violated.

\begin{figure}[t]
    \centering
    \begin{subfigure}{\linewidth}
        \centering
        \includegraphics[width=0.25\linewidth]{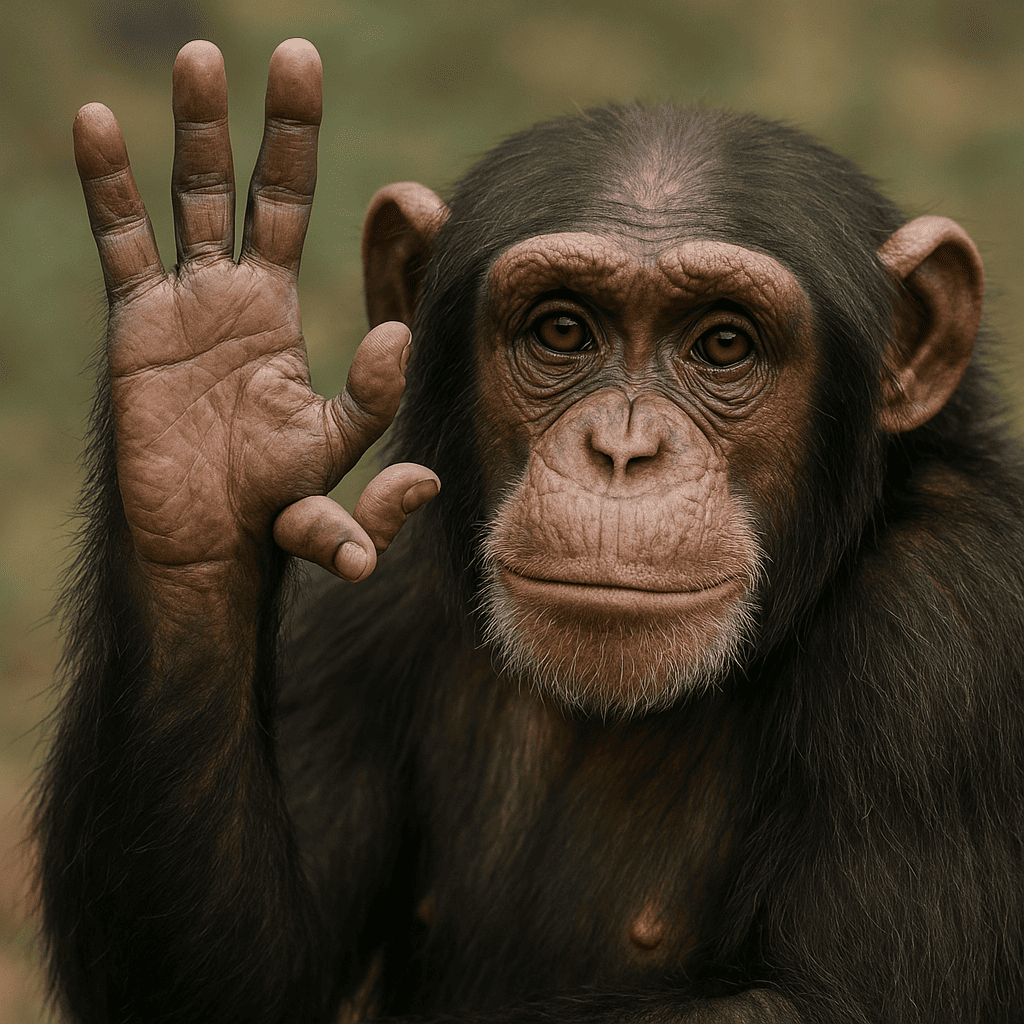}
        \includegraphics[width=0.25\linewidth]{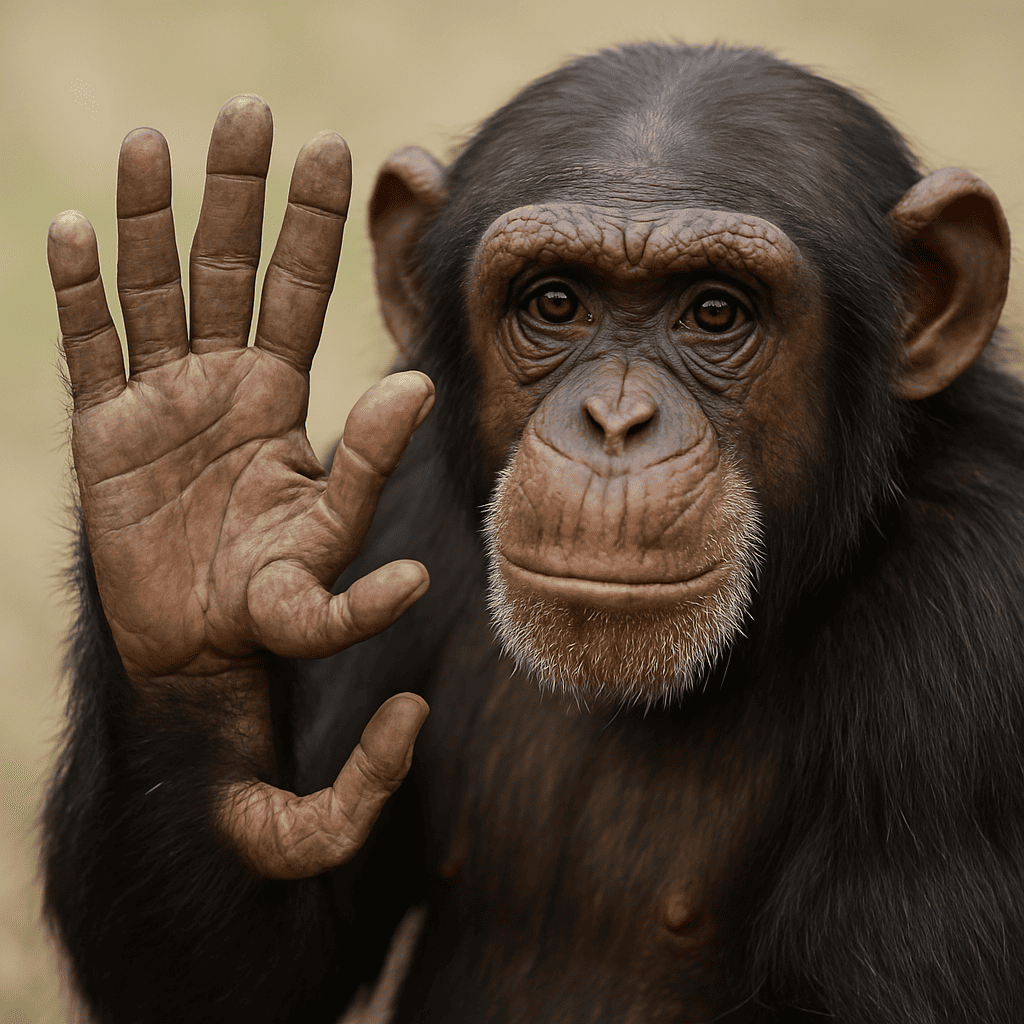}
        \includegraphics[width=0.25\linewidth]{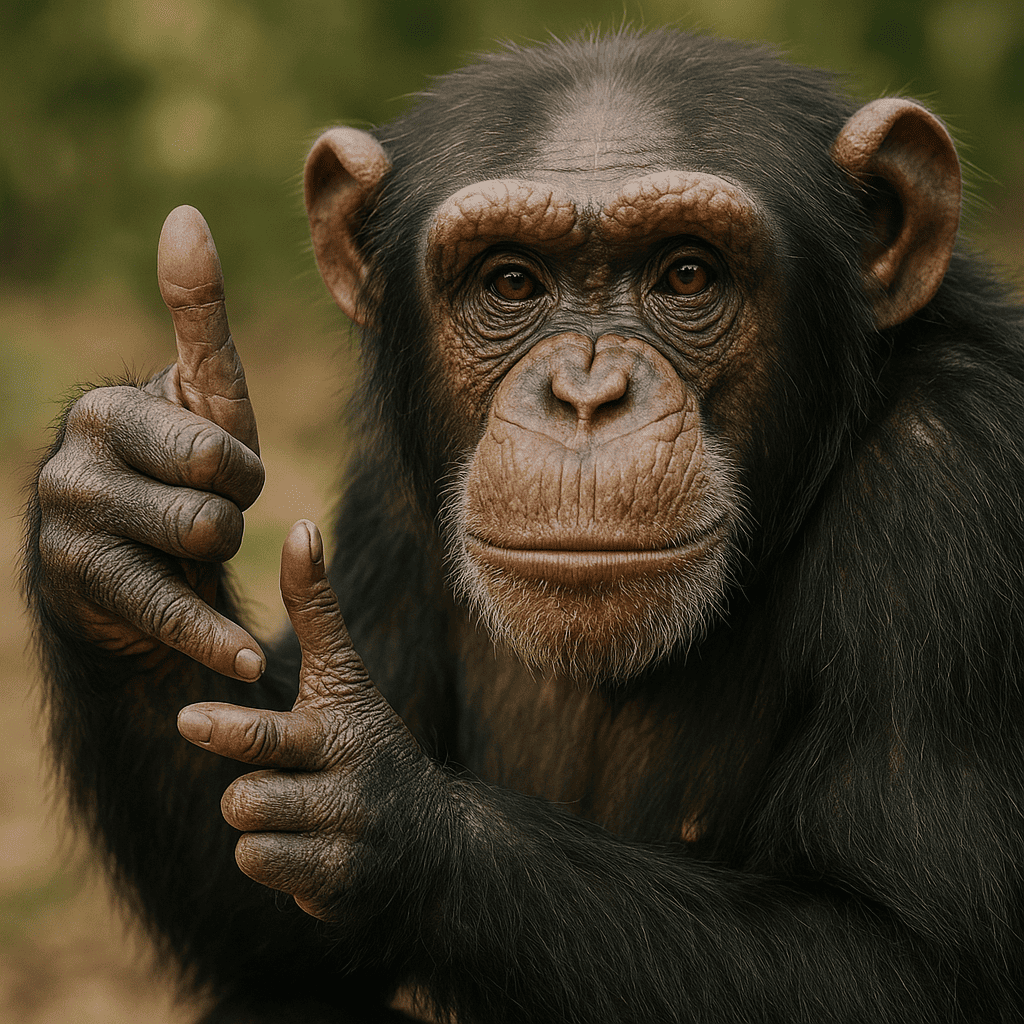}
        \caption{``generate an image of an ape holding up a hand with an extra thumb stuck onto it''}
        \label{fig:thumbs_ape}
    \end{subfigure}
    \begin{subfigure}{\linewidth}
        \centering
        \vspace{0.1in}
        \includegraphics[width=0.25\linewidth]{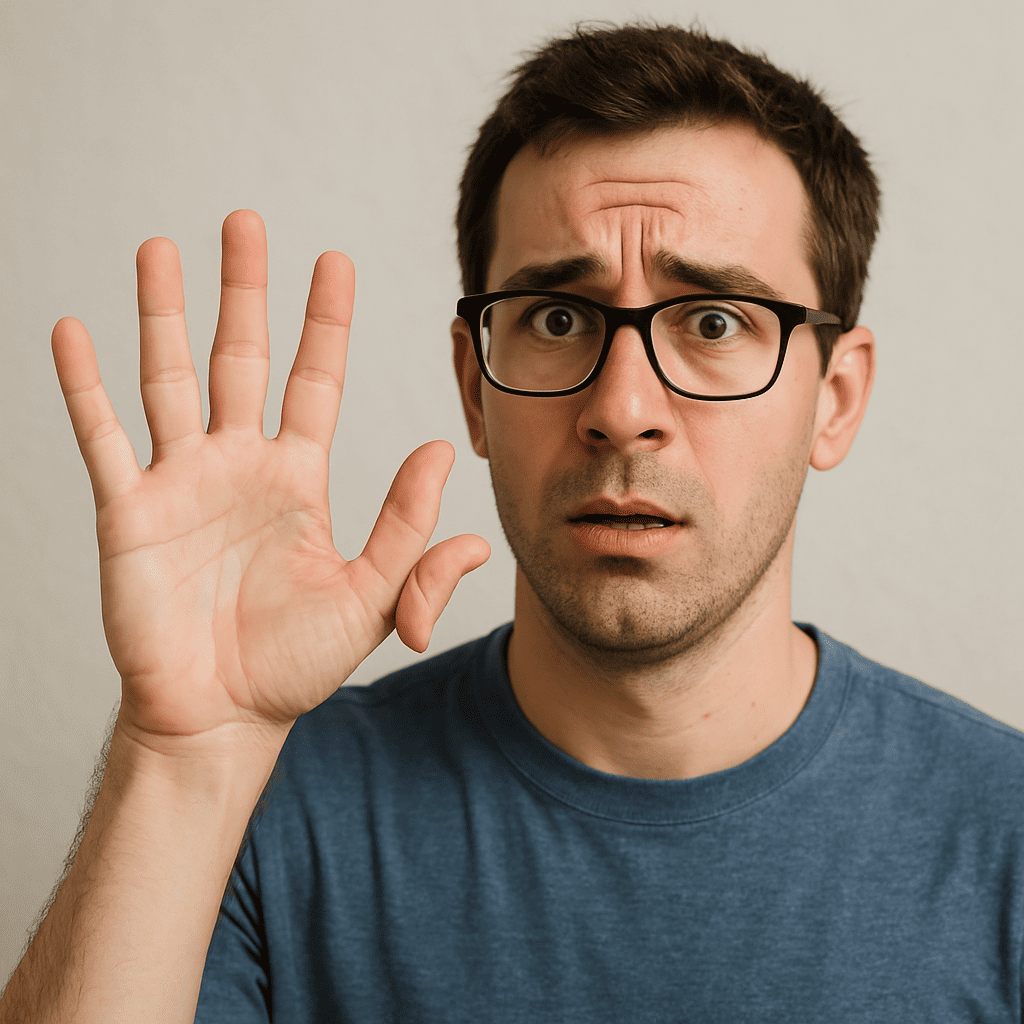}
        \includegraphics[width=0.25\linewidth]{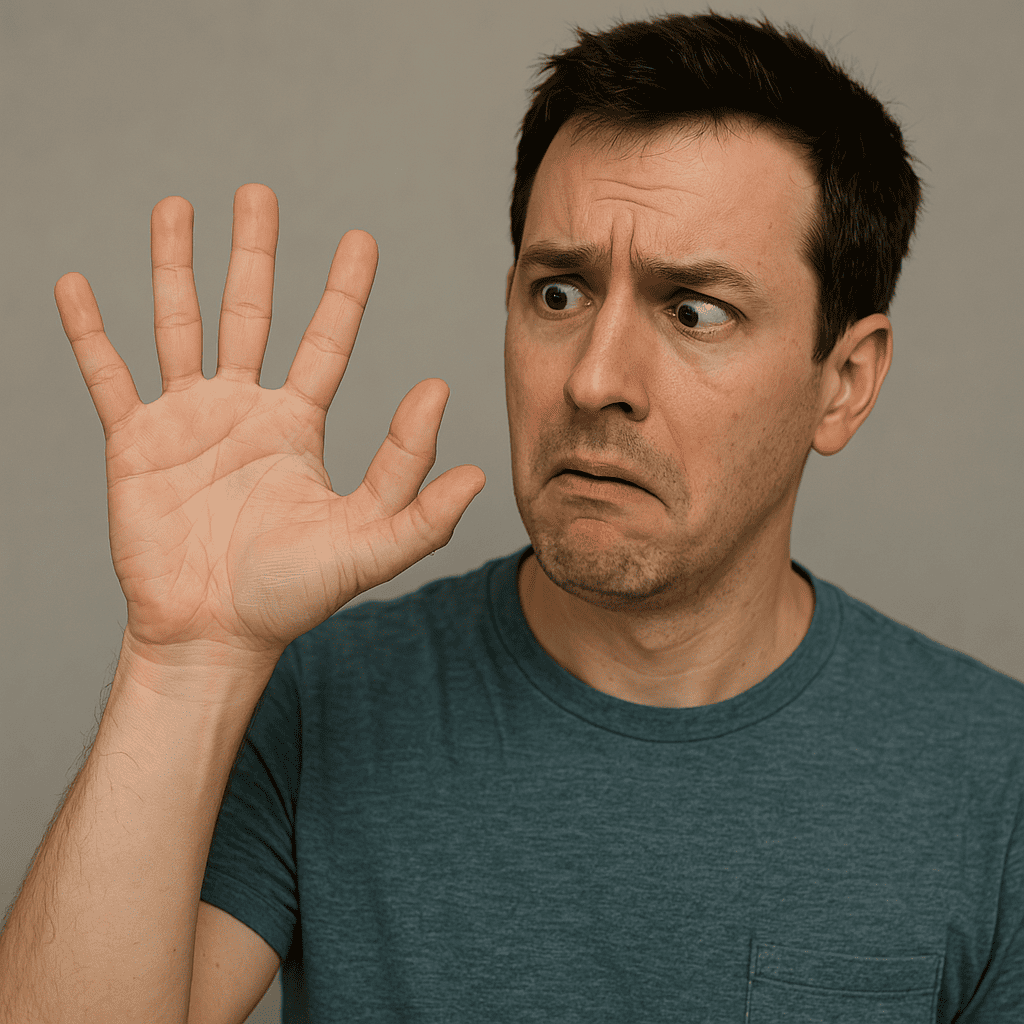}
        \includegraphics[width=0.25\linewidth]{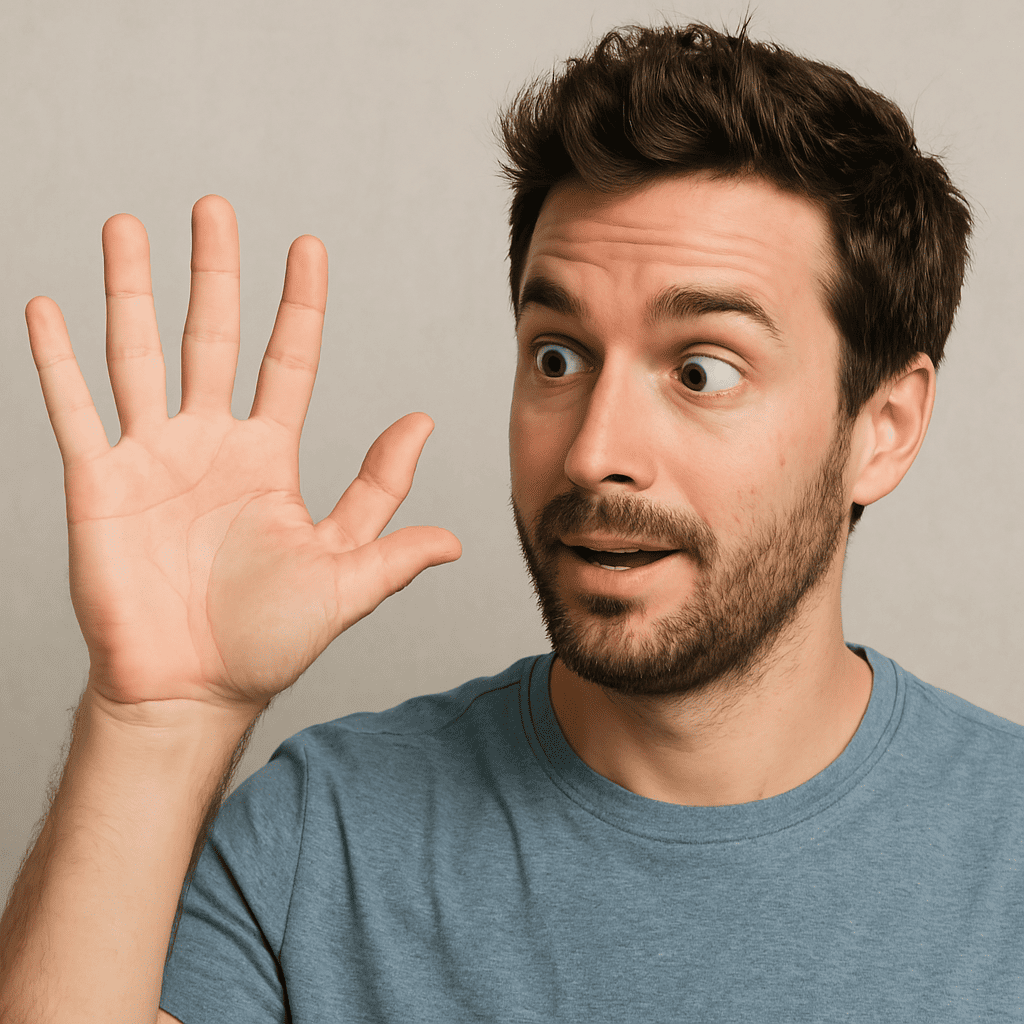}
        \caption{``generate an image of a man holding up a hand with an extra thumb stuck onto it.''}
        \label{fig:thumbs_man}
    \end{subfigure}
    \caption{
    \textbf{Evidence of FER in image generation through ChatGPT-4o (new release from March 2025).}
The inability to depict an ape hand with a single extra thumb (a) seems to imply that the image generator lacks the knowledge to depict such a hand.  However, its ability to depict a human hand with an extra thumb (b) shows that the conceptual apparatus is actually present, but the network is fractured in such a way that it does not represent adding a thumb internally as a general regularity.}
    \label{fig:thumbs}
\end{figure}

Another effect of FER in current models is evident in text-to-image models like the newly enhanced one released in March 2025 by OpenAI as part of GPT-4o: they often can only satisfy specific kinds of requests in some contexts but not others. 
For example, Figure \ref{fig:thumbs_ape} shows three examples of what happens if GPT-4o is prompted with, ``generate an image of an ape holding up a hand with an extra thumb stuck onto it.'' 
(Apes have four fingers and a thumb, like humans.)
The network always fails, differently in each case.
For example, in the leftmost image, the ape has only three fingers and has two extra thumbs (instead of one extra), in the middle image it also has two extra thumbs, and in the rightmost image it is holding up two hands, both missing a finger and neither with an extra thumb.  
Based on this behavior, the network seemingly cannot understand the concept of ``with an extra thumb.''
Yet how can we be confident that the problem is FER rather than a general capability limitation?

The evidence of FER is that the network succeeds every time when asked, ``generate an image of a man holding up a hand with an extra thumb stuck onto it.'' (The only change is from the word ``ape'' to ``man.'') 
The hint of FER is that the network clearly actually \emph{does} know what it means to add a single extra thumb to a hand, as shown in Figure \ref{fig:thumbs_man}.
The problem is that it does not seem to be able to apply that understanding to the case when the hand is not human, suggesting two fractured and redundant procedures for generating an image of a hand with a specific finger configuration. The kinds of mistakes the model makes also suggest some degree of entanglement between the method for rendering a normal ape's hand and the method for organizing a specific finger configuration.

The evidence of FER may explain to some extent the challenge faced by the field of mechanistic interpretability in fully understanding the computations performed by large models~\citep{bereska2024mechanistic, lindsey2024biology}.
This difficulty is often attributed to superposition~\citep{elhage2022toy} and polysemanticity~\citep{olah2017feature, olah2020zoom}.
Superposition refers to the phenomenon where networks are able to represent (exponentially) more features than there are neurons, by entangling multiple features within single neurons.
This entangling leads to polysemanticity, where individual neurons respond to multiple, seemingly unrelated concepts.
These properties are likely consequences of the fractured and entangled circuitry that underlies the network’s representation.

The ideal of disentangled representations and the weight sweeps from Figure~\ref{fig:weight_sweeps_576} may remind some readers of the literature on disentangled latent space control in variational autoencoders~\citep{higgins2017beta, burgess2018understanding} and generative adversarial networks (GANs)~\citep{radford2015unsupervised, karras2019style, harkonen2020ganspace}.
These works show that the latent space for these kinds of generative models is surprisingly interpretable and often features directions of variation that correspond to meaningful semantic changes in the generated images.
However, even these models still fall victim to the biases of the dataset~\citep{jahanian2019steerability} and thus fail to fully capture the underlying regularities of natural images.
Moreover, note that the term ``latent space'' in these works
refers to sweeping over inputs or activations, whereas in this paper we sweep over the \emph{weights} of networks and illuminate the holistical representational strategy of the whole network.
The evidence in this paper from many different angles---the SGD CPPN skull (and other images in Appendix~\ref{sec:other_picbreeder_examples}), the odd difference in the ability to count objects based on their type, the extra thumb request working for humans, but not chimps, and the fact that mechanistic interpretability work on math and Othello exhibit inelegant collections of hacks that lead to outwardly impressive behavior---all point to the fact that it remains possible that the internal machinery
of large models is flawed with FER and poor heuristics.
The potential connection of FER and UFR to work on disentangling latent input spaces highlights the need for deeper investigation into the holistic representational strategies of networks beyond just their input-output mappings.

Overall, the conclusion is that LLMs do exhibit behaviors suggestive of FER.
Considering the results with Picbreeder and conventional SGD CPPNs, this result is not surprising---in the experiments in this paper, left to its own devices, SGD exhibits a tendency to produce FER in the absence of any special intervention. Of course, any specific example of inconsistent output behavior could be improved in a future model update, but the the point here is not that any specific error can never be fixed.  Rather, the point is that there is reason to believe that FER continues to exist in models' underlying representations even as they scale.

Foundation models face many known challenges today, including sample inefficiency~\citep{kaplan2020scaling}, lack of reliability~\citep{zhou2024larger}, hallucination~\citep{huang2025survey}, poor generalization out of distribution~\citep{wu2024reasoning}, idiosyncratic failures on simple tasks~\citep{mirzadeh2024gsm}, poor continual learning~\citep{berglund2024reversalcursellmstrained}, etc.
An interesting and open question is, to what extent are each of these challenges caused by FER? To what extent does FER pose a fundamental challenge to the foundations of the entire modern AI enterprise?

The natural next question then is, are there interventions that might mitigate this problem?

\section{Key Factors in FER and UFR}

This section contemplates the causes of FER and speculates on potential mitigations that can help achieve UFR.

\subsection{The Role of Order and the Ability to Reorganize Representation}
\begin{figure}[t]
    \centering
    \begin{subfigure}{1.0\textwidth}
        \centering
        \includegraphics[width=\textwidth]{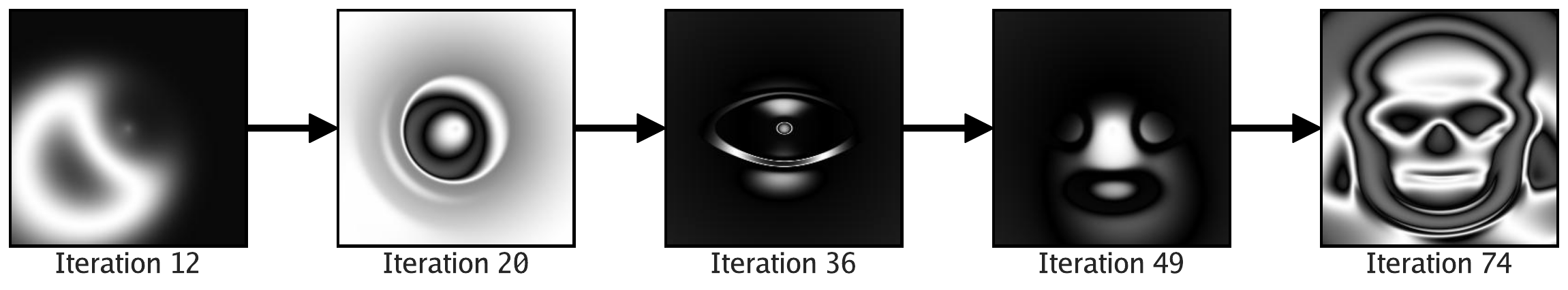}
        \caption{Picbreeder path to the skull}
        \label{fig:path_to_576_pb}
    \end{subfigure}
    \vspace{0.0cm}
    \begin{subfigure}{1.0\textwidth}
        \centering
        \includegraphics[width=\textwidth]{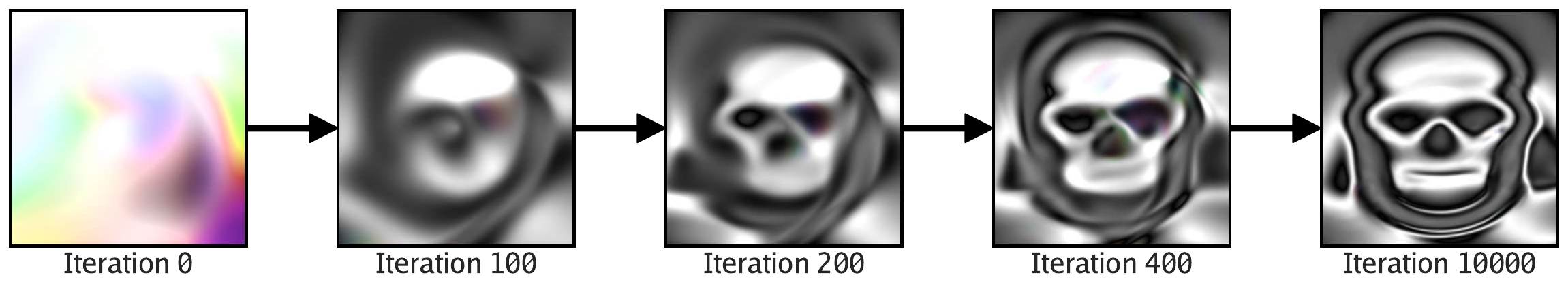}
        \caption{Conventional SGD path to the skull}
        \label{fig:path_to_576_sgd_pb}
    \end{subfigure}
    \caption{
    \textbf{Differing paths taken by Picbreeder and conventional SGD to reach the skull image.}
    Picbreeder follows a non-intuitive, serendipitous curriculum \citep{stanley2015greatness}, whereas SGD follows a more direct, greedy path.
    These distinct curricula influence how the neural network circuits are formed and how the skull is internally represented.
    The vast disparity in number of iterations to reach the final image is also interesting.
    }
    \label{fig:path_to_576}
\end{figure}

In the ancestry of the Picbreeder Skull CPPN, symmetry emerged long before the skull did, and even a face-like structure (with a mouth and eyes) was discovered before it looked anything like a skull, shown in Figure~\ref{fig:path_to_576_pb}.
That is why the skull that later emerged was able to avoid FER: it inherited these previously learned regularities.
The same phenomenon of regularities being discovered and elaborated upon later is seen in other Picbreeder CPPNs like the butterfly (Appendix~\ref{sec:other_picbreeder_examples}).
In contrast, as shown in Figure~\ref{fig:path_to_576_sgd_pb}, the conventional SGD Skull CPPN took a direct path to the skull without building regularities along the way.
The tendency of humans to choose fundamental regularities early is probably related also to an intuitive appreciation for \emph{evolvability} \citep{dawkins2019evolution,kirschner1998evolvability,wagner1996perspective}, or the potential of an artifact to produce even more interesting artifacts in the future.
Humans also are drawn to things they find beautiful, and research has shown that human notions of beauty often involve properties like regularity, including symmetry, and other forms of compressibility~\citep{enquist1994symmetry, schmidhuber1997low}.

Notice how such a virtuous ordering
(symmetry then face) leads to a much easier path to future creativity than needing to reorganize representation and unify fractured concepts to later achieve UFR.
If the network does not learn symmetry first, as in the conventional SGD skull, it ends up learning the two similar sides of the skull separately. For it to fix the consequent FER, it would need to reconcile that oversight later on, which is much less straightforward and probably expensive (likely requiring a lot of data or compute), if possible at all. 
At some point, the amount of proliferating FER is potentially insurmountable.

\begin{figure}[t]
    \centering
    \includegraphics[width=1.0\textwidth]{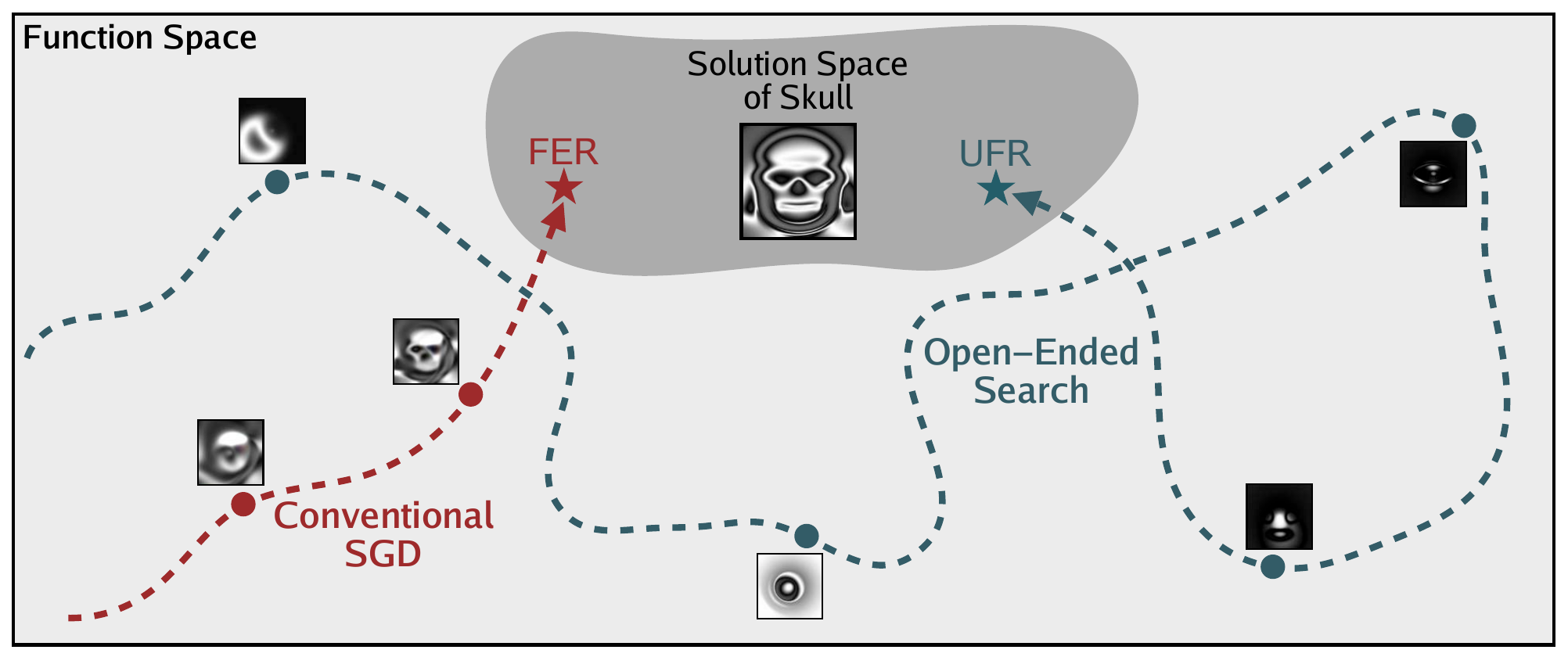}
    \caption{
    \textbf{Different trajectories to FER and UFR for outwardly identical solutions.}
    Within the function space, the solution
    space is the subset of the space that fully satisfies an objective (e.g.\ perfectly reconstructing the skull or modeling all text on the internet).
    Within this solution space, both FER and UFR solutions exist.
    Conventional SGD often follows a direct path toward this solution space, typically leading to FER solutions.
    However, alternative search algorithms---including open-ended and serendipitous ones \citep{lehman2011abandoning,stanley2015greatness,secretan2011picbreeder,mouret2015illuminating,wang2019paired,nguyen2015innovation}---may discover UFR solutions.
    }
    \label{fig:hypothesis_space}
\end{figure}

In other words, the order in which a concept is learned is fundamental to the representation eventually achieved, and hence (because FER can potentially impact creativity) to the ability to imagine something compelling that is new %
(Figure~\ref{fig:hypothesis_space}).
Indeed, as Figure~\ref{fig:weight_sweeps_576} shows, the future concepts a network can learn are radically different depending upon how its output pattern was learned, even when the learned pattern itself is identical.
In this light, the unnatural order in which concepts are learned in LLMs via SGD might hold important implications for their eventual internal representations.
In particular, during training, LLMs learn many concepts in parallel that humans would absorb strictly sequentially.

For example, during pre-training, the data for arithmetic and calculus are absorbed simultaneously, even though a proper understanding of calculus depends on arithmetic.
This predicament can lead to potential FER for arithmetic because learning calculus requires at least some minimally functional arithmetic, but because those skills are not yet mature enough to build upon, the model is incentivized to develop separate calculus-specific arithmetic heuristics.
Just as the two sides of the skull or the two wings of the butterfly (Figures \ref{fig:fmaps_4376_pb} and \ref{fig:fmaps_4376_sgd_pb}) can be represented through separate heuristics, FER can emerge on arithmetic, and probably not just in the context of calculus either: there are innumerable domains and skills that require arithmetic thinking.
That may be why GPT-3 can count office supplies but not animals.

Of course, through the intentional design of educational curricula, humans are not presented with the problem of learning arithmetic and calculus at the same time.
It is an interesting thought experiment to imagine what would happen if we were.
Humans also encounter technological, scientific, and artistic innovations in a particular order, and that chronology may impact the way we represent and eventually build on these higher-level ideas as well. 

These arguments may lead some readers to explore engineering more natural curricula, such as determining the most effective ordering of internet data to minimize FER; indeed, researchers are exploring the impact of curricula in general~\citep{soviany2022curriculum}.
While these attempts to make the ordering more logical may improve representation to some extent, they are unlikely to be a silver bullet.

The problem is that trying to figure out the optimal order for everything by hand is virtually impossible.
Do toasters need to be taught before ovens?
There are billions of arbitrary decisions to be made.
We humans too face some degree of arbitrary ordering
in our life experience, which is a reason beyond raw intelligence that some may be more predisposed to great discovery than others.
Of course, school is ordered intentionally to build more advanced topics upon simpler ones.

Given that order cannot easily be engineered, or at least not entirely, another useful factor in human representation learning might be the ability to avoid absorbing information that we are not ready for.
Most children would not be able to start absorbing calculus at the same time they are learning to count.
Rather, most children would simply tune it out and find it boring or incomprehensible.
This natural tendency to know what you are and are not ready to learn is another missing capability in current LLMs.
In fact, when people talk about LLMs ``not reasoning''~\citep{kambhampati2024can, dziri2023faith}, the place where reasoning is most lacking may not actually be at inference time, but rather during learning itself.

However, order cannot be everything---humans seem to be capable of intentionally reorganizing information through reanalysis or recompression,
without the need for additional input data, all in an attempt to smooth out FER.
It is like having two different maps of the same place that overlap and suddenly realizing they are actually the same place.
While clearly it is possible to change the internal representation of LLMs through further training, this kind of \emph{active and intentional} representational revision has no clear analog in LLMs today.

While in effect we seem to have two key mechanisms for attaining UFR in human (and animal?) brains, through knowing what to ignore and through continual revision of internal representation, reality is likely more messy.
For example, both processes are likely happening simultaneously.
Consolidation can happen even as new kinds of data become admissible.
Furthermore, it is conceivable that humans also transiently hold more than one representation simultaneously, perhaps experimenting with different conceptualizations of an idea until one ends up winning.
While that scenario sounds reminiscent of FER, an important distinction is that this hypothetical human diversity of representations is intentional, benefits from self-awareness, and temporary.
That is, the process of trying multiple frameworks and pruning as more is learned is arguably more organized than in an LLM, though much is still unknown.

\subsection{NEAT and Increasing Complexity}

Other factors may also contribute to the UFR seen in Picbreeder CPPNs. For example, the NEAT algorithm \citep{stanley2002evolving} inside Picbreeder is designed to start with small, simple networks and then occasionally add more structure (and thereby increase the size and potential complexity of the network) gradually over generations. This escalating trajectory from simple to more complex probably explains at least in part the tendency of Picbreeder users to stumble across fundamental organizing principles (on which further elaborations are later built) early in their explorations, and perhaps also the efficiency with which such discoveries are routinely made. 

That said, the ultimate importance of NEAT (or whether an unconventional kind of SGD can be led through a similar trajectory) to the overall lessons drawn here is not currently well understood. Furthermore, it is important to note that the open-ended style of search may end up more important than the distinction between NEAT and SGD: in \citet{woolley2011deleterious}, NEAT networks trained to output Picbreeder images as explicit objectives without human users guiding the search exhibit significantly more bloated representations (i.e.\ have more neurons and connections) than the original identical discoveries in Picbreeder, which hints at FER. That result suggests that the open-ended (non-objective) nature of the Picbreeder search process is potentially a more fundamental factor than the NEAT algorithm.

While we do not yet fully understand the principles involved, the Picbreeder skull CPPN (and others in Appendix~\ref{sec:other_picbreeder_examples}) at least demonstrates that neural networks \textit{can} in principle exhibit UFR, though it arises there from a radically different training paradigm than that used for modern LLMs.

\subsection{More Data and Holistic Unification}
Probably the biggest hope would be that more data would solve the problem on its own, as it has helped to improve frontier model capabilities in general~\citep{kaplan2020scaling, huh2024platonic}.

While conventional SGD on a single skull image tends towards FER, what if the network were instead trained to output two different skulls, conditioned on an input ID?
What about four skulls, or eight?
What about trillions of skulls?
(These kinds of experiments are interesting directions for future work.)
At some point, it is conceivable
that the network might experience \textit{holistic unification} that flips from FER to UFR (either gradually or quickly), 
perhaps underlying some cases of ``grokking''~\citep{power2022grokking,liu2022towards},
and start understanding that skulls—its only possible outputs—are inherently symmetric. Alternatively, more examples might just lead to more FER with better loss, which would not be the holistic unification we want.

This idea of holistic unification resulting from multiple examples feels satisfying in part because it is reminiscent of what we expect to see in deep networks and even LLMs.
Large models see numerous examples of a phenomenon during training.  For example, recall the simple math problem from above that stumped GPT-3, which always gets it wrong: 

\begin{verbatim}
I have 3 chickens, 2 ducks, and 4 geese.  How many things do I have?
\end{verbatim}

GPT-4 seems to have experienced holistic unification: it gets it right every time, with every object category.
The general procedure for adding up items (whatever types of items they are) seems to have become unified rather than fractured, presumably because GPT-4 has seen many more examples of simple arithmetic.
At some point it ``realized'' that all these different types of adding are actually examples of the same thing.

The role of grokking~\citep{power2022grokking, liu2022omnigrok, liu2022towards, varma2023explaining} in such transitions and its potential to undo FER is an interesting open question.
\citet{nanda2023progress} raise the possibility that grokking may include a period of ``cleanup'' wherein obsolete memorized circuits are washed away from the final generalized learned algorithm.
The prospect of a late-stage phenomenon like grokking cleaning up the mess of FER is attractive, but it does not address the question of how often such a unification actually happens in practice.
\citet{nanda2023progress} note that grokking depends on the precise parameterization of weight decay and dataset size, suggesting that grokking could be hard to produce without just the right settings. It is likely even more complex when many cognitive capabilities are optimizing within the same model. 
Perhaps even more importantly, even if grokking were a reliable and consistent process in all of learning, it is intriguing to think that there may be a different kind of learning (as seen with Picbreeder CPPNs) that would not have necessitated all that effort and clean up in the first place. 

Furthermore, the prospect of holistic unification through comprehensive data (even if generally true) does not sufficiently address the underlying worry.
The problem is that any data on the frontier of knowledge---scientific hypotheses, novel algorithms, new kinds of art, etc.---will in effect appear in relatively little data.
Because of that, the frontiers themselves, the very places we want the most cogent and incisive insight and imagination, are likely the most rife with FER.

\subsection{Architectural and Algorithmic Changes}

The examples in this paper showcase conventional SGD on simple feedforward networks.
There is a possibility that architectural changes to the network, or algorithmic changes (including carefully chosen regularization) to the usual process of SGD (or some combination of both) could mitigate FER.
In fact, modern large models such as LLMs already differ in important ways from simple MLPs, so there is even the possibility that one of these more recent architectural elaborations is already helping with FER.

For example, Mixture of Experts (MoE)~\citep{jacobs1991adaptive, shazeer2017outrageously}, which is often part of modern LLM training~\citep{liu2024deepseek, jiang2024mixtral}, explicitly enforces a degree of modular decomposition.  In principle, the consequent disentanglement of multiple functionalities from each other might yield a better representation.  However, on the flipside, it is also possible that by forcing specific ``expert'' sub-components to activate for different tasks, the same underlying principle might need to be relearned for multiple expert instances, which would be a recipe for redundancy and FER.  It is also possible that the expert subnetworks are so large anyway that MoE does not actually have a significant effect in either direction.

Without further rigorous empirical study (which perhaps this paper will motivate), these kinds of ``maybe it does help and maybe it doesn't'' arguments are likely to recur for MoE, sparsity constraints, and many other architectural and algorithm augmentations, both current or proposed.
Different network architectures like convolutional networks~\citep{lecun1989backpropagation}, residual networks~\citep{he2016deep}, and transformers~\citep{vaswani2017attention} all exhibit different kinds of weight sharing and sparsity~\citep{elsken2019neural}, and thus have different prospects in mitigating or exacerbating FER.

Another opportunity may be a kind of \emph{representational massaging}, such as through pruning excess neurons and then repairing any loss through retraining, which could lead to holistic unification. Or perhaps SGD itself can be modified, e.g.\ to direct updates to more semantically appropriate areas, thereby encouraging more unification and less entanglement. %

It is certainly possible and even likely that \emph{something} helps, but the deeper principles that can guide our thinking towards a more systematic approach to mitigating FER will likely only emerge with further study.

\subsection{Open-Ended Search }

As noted above, the chronology of ancestral patterns that precede the skull is important.
In that chronology, left-right symmetry was discovered early (between iterations 20 and 36 in Figure~\ref{fig:path_to_576_pb}).
The reason that the skull has symmetry under the hood is because it was built upon a previous chance discovery of symmetry.
Though the initial discovery was chance, once it was discovered, it served as the organizing basis for many subsequent progeny.
That initial discovery of symmetry in Picbreeder is analogous to the emergence of bilateral symmetry in flatworms in natural evolution~\citep{carroll2005endless}, which, in both cases, becomes a fundamental organizing principle for vast numbers of future discoveries.

\begin{figure}[t]
    \centering
    \includegraphics[width=1.0\textwidth]{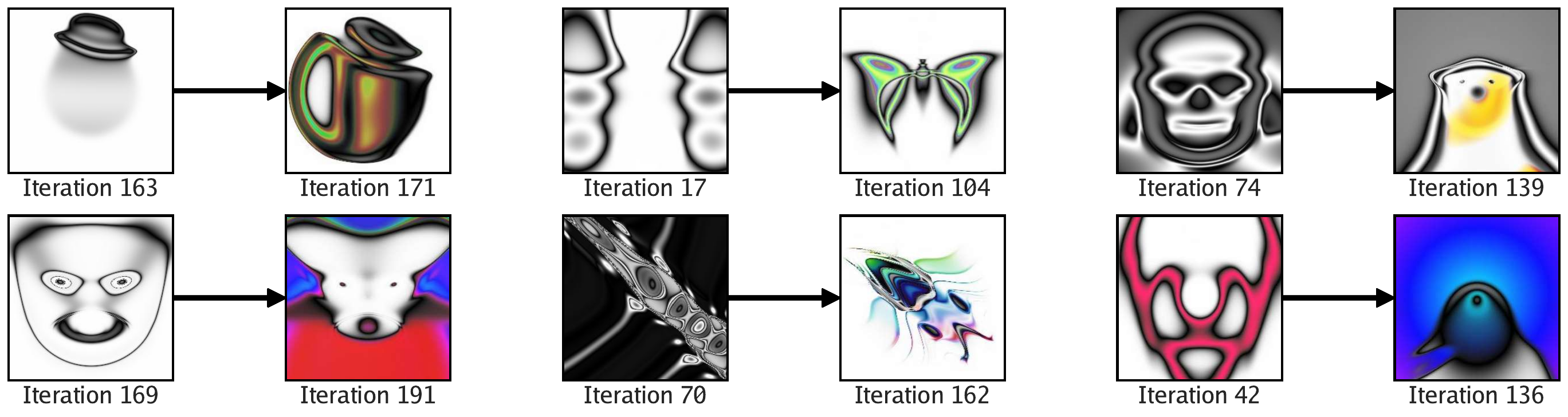}
    \caption{
    \textbf{Picbreeder enables serendipitous discoveries.}
    The majority of images are found in Picbreeder through serendipitous paths, where the stepping stones did not resemble the final discovery.
    }
    \label{fig:pb_serendipity}
\end{figure}

In the literature, the success of discovery processes like Picbreeder, natural evolution, or scientific discovery are attributed to \emph{open-endedness}, which refers to a divergent search process that collects interesting discoveries along the way without a final goal~\citep{stanley2015greatness, stanley2017open,wang2019paired,wang2020enhanced,nguyen2015innovation,hughes2024open}.
While now an established research area within AI \citep{team2023human,hughes2024open},
the field's origins extend back to research into open-ended evolution in artificial life, where the aim is to understand how natural evolution facilitates its own form of open-ended discovery~\citep{bedau11998classification,channon2006unbounded,standish2003open,taylor2019evolutionary,oeeworkshop2016,maley1999four,miconi2008road}.
Because open-ended search maintains a diverse archive of interesting discoveries, serendipitous stepping stones to even more intriguing findings become exponentially more likely~\citep{stanley2017open}.
The ``curriculum'' of discoveries that eventually leads to final outcomes often only makes sense in hindsight, as shown in Figure~\ref{fig:pb_serendipity}.
This principle is true in technological progress as well, where the stepping stones to major advances often were not originally invented with that final destination in mind \citep{stanley2015greatness}.

Open-ended search also fosters the emergence of \textit{evolvable} artifacts—those that are not only interesting in their own right, but are easily adapted by the search process to generate other interesting artifacts~\citep{kirschner1998evolvability, wagner1996perspective, huizinga2018emergence}.
In well-structured open-ended evolutionary settings, these evolvable artifacts rapidly dominate the population, outcompeting less adaptable alternatives.
In the context of neural networks, what emerges should be modularity, hierarchy, and UFR, as these properties enable the greatest adaptability.

Because open-ended processes often produce surprisingly elegant representations under the hood (as seen in the examples in this paper), it suggests that a more open-ended approach to large model training might yield less FER as well.
For example, MAP-Elites provides a good abstraction to capture the serendipitous curricula similar to ones found in open-ended search~\citep{mouret2015illuminating}, and the
minimal criterion coevolution (MCC) \citep{brant:gecco17,brant:gecco20},
POET ~\citep{wang2019paired, wang2020enhanced} and OMNI \mbox{\citep{zhang2023omni, faldor2024omni}} algorithms may provide inspiration for open-endedly generating problems and solutions at the same time.
However, there is not yet a definitive candidate for such a process paired with SGD for large models.
There is a lot of room for future creativity for the field here.

\section{Discussion}

Representation underlies all generative processes.
A good representation and a bad representation can \emph{both} be accurate (e.g.\ both can draw a perfect skull or butterfly).
The existence of the Picbreeder variants of these images shows that at least \emph{in principle} much better representations are theoretically achievable.
The key question is when it starts to matter.
This question is particularly important for today's LLMs, which 
could be as similarly riddled with FER under the hood as the SGD-derived skull CPPN.

While in this paper we focused on regularities like y-axis symmetry, hand image generation, arithmetic, and text substitution, there are innumerable additional regularities that are possible in more complex search spaces.
For example,
imagine the number of regularities that the human brain represents in concept space through its neural circuits: invariance to lighting in a scene, recognizing a dangerous animal regardless of the background, scientific laws, social norms, etc.
Essentially, these regularities, composed on top of each other, determine how the human brain interprets and makes sense of the world in a robust way.

The reason it matters is that generalization, creativity, and learning all
fundamentally depend on neural representation actually capturing such regularities. If it does not, all these critical capabilities start breaking down when extended, modified, and combined.  Generalization degrades if fundamental regularities are missed;
creative ideation is thwarted by uninhibited violations of deep regularities; and learning becomes outlandishly expensive when the adjacent possible is littered with the constant unraveling of regularities seemingly (but not actually) solidly ensconced.

A common sentiment in AI is that ``intelligence is just compression.'' In this view, intelligence is framed through an information theory lens, drawing on concepts like Occam's Razor and Minimum Description Length~\citep{solomonoff1964formal, rissanen1978modeling, hutter2000theory}.
However, importantly, good UFR is related but fundamentally distinct from compression alone:
the maximal compression of a skull in a CPPN might not capture dynamic features like eye winking or jaw widening, even though these very movements are crucial in the context of creativity and continual learning. The \emph{factored} property of UFR captures this additional aspect of good representation.  While extensive training examples showing these features explicitly varying along relevant dimensions (like a mouth opening and closing) could help, that does not address cases where the data is not sufficiently extensive.  Yet the insight here is deeper than just pointing out that representation might not work out well without a lot of data: the surprising insight offered by the Picbreeder representations is that there is a path to such factorization \emph{without} the need for extensive representative examples. That result deserves further contemplation.

Of course, the simple demonstrations in this paper do not fully illuminate all these issues.
Much could be written on the intricacies of these implications beyond the preliminary discussion here, but the hope is that this paper will mark the start of a larger discussion, rather than covering every route to completing one.

\section{Conclusion}

We anticipate that opinions will vary on the seriousness of the implications of FER.
Representational optimists will argue that FER washes away with enough data, or that it does not even matter in the first place. Time will tell.
Yet we believe it cannot be denied that the difference between outward performance and inward representation is a harbinger of lessons yet to be learned.
Two students may both ace a math Olympiad exam, but one can go on to become a great mathematician while the other achieves nothing novel in their field~\citep{stanley2025conversation}.
The danger of outward performance---of benchmarks---is that they miss this critical distinction.
Somewhere beneath the surface, where representations form and churn, 
there is an explanation for why simply doing well on a test or benchmark does not foretell later creative feats of great distinction.
FER may be a critical yet heretofore unappreciated part of that story.
Moreover, the hope that FER will vanish with scale---representational optimism---may be misplaced and merits scrutiny and additional research.

\section*{Acknowledgments}
We thank Eric Frank for insightful discussion on potential implications of vector rotation (addressed in Appendix \ref{sec:pca_feature_space}), Phillip Isola for his feedback on the conceptual framing of the paper, and Jyo Pari for discussions on Mixture of Experts. We also thank Joost Huizinga for leading his extensive 2018 study of the elegant representations found in Picbreeder CPPNs~\citep{huizinga2018emergence}, including building the open-source software tool CPPN-Explorer (CPPN-X) that facilitates such research.  
This work was supported in part by an NSF GRFP Fellowship to A.K. J.C. was supported by the Canada
CIFAR AI Chairs program, a grant from Schmidt Futures, an NSERC Discovery Grant, and a generous donation from Rafael Cosman.

\bibliographystyle{apalike} %
\bibliography{references}

\newpage
\appendix

\section{Details of NEAT, CPPNs, and Picbreeder}
\label{sec:details}

\subsection{Details of NEAT}
\label{sec:details_neat}

NeuroEvolution of Augmenting Topologies (NEAT) is an evolutionary algorithm that evolves both the weights and topology of neural networks~\citep{stanley2002evolving}.
NEAT begins with a population of simple networks and progressively evolves more complex networks over many generations.

NEAT encodes a neural network (NN) as a list of node genes, which specify the existence of neurons, and a list of connection genes, which specify synaptic connection weights between neurons.
Together, this genetic encoding represents NNs as directed acyclic graphs.
The node list specifies each neuron's activation function, which, when NEAT is used to evolve CPPNs, is chosen from the following set: $\text{Identity}(x)=x, \sin(x), \cos(x), \tanh(x), \text{Sigmoid}(x)=\frac{2}{1+e^{-x}}-1$, and $\text{Gaussian}(x)=2e^{-x^2}-1$.
A key feature of NEAT is the use of innovation numbers (also called historical markings) to track new genes, preserving historical information about the evolutionary process and facilitating comparisons between different NNs (for example, during crossover).
During reproduction, NEAT uses mutation operations that can add new nodes, add connections, split a connection, or mutate weight values.
Additionally, NEAT can cross over two networks by aligning their genomes using innovation numbers and performing an informed crossover of the genes.

In the original NEAT algorithm, ``speciation'' separates genomes by similarity to encourage diversity in the population.
However, this feature is not present in Picbreeder because humans naturally optimize for many different objectives, providing an implicit selection pressure for diversity.

\subsection{Details of CPPNs}
\label{sec:details_cppn}
Each CPPN takes as input a pixel's $x$- and $y$- coordinates, both ranging from $[-1, 1]$.
In addition, the CPPN receives the input $d = 1.4 \sqrt{x^2+y^2}$, and $b=1$ (for historical compatibility with Picbreeder CPPNs).The CPPN processes these inputs to produce the raw $h, s, v$ color value for the pixel.
These values are then converted to $r, g, b$ by the equation
\begin{equation*}
    r, g, b =  \text{hsv2rgb}\left(h \mod 1,\quad \text{clip}(s, 0, 1),\quad \text{clip}(|v|, 0, 1)\right).
\end{equation*}
$\text{hsv2rgb}$ is a standard HSV to RGB conversion function with all inputs and outputs $\in [0, 1]$.
By sweeping over the entire grid of possible $x$ and $y$ values, this process can generate an image at any resolution.

\subsection{Details of Picbreeder}
\label{sec:details_picbreeder}
Picbreeder was an online website where humans could evolve CPPN images using NEAT~\citep{secretan2008picbreeder}.
In the beginning of a user session starting from scratch, a random set of initial small NEAT CPPNs would be generated.
For each generation, the human would be shown 15 CPPN images, and then could select the ones they like.
If multiple were selected, a crossover between the selected CPPNs would be performed.
These CPPNs were then mutated to arrive at the next generation of CPPNs.
At any point, if a user was satisfied with an evolved image, they could publish its CPPN to the Picbreeder website, adding it to a public gallery.
Others could browse this gallery and select images to continue evolving from where previous users left off, resulting in the open-ended evolution of CPPN images.

We provide the Picbreeder data for the three CPPNs shown in this paper (skull, butterfly, apple), and the code to interpret this data at the \href{\codelink}{code link}.

\section{Porting NEAT CPPNs to Dense MLP CPPNs}
\label{sec:picbreeder_porting}

\begin{figure}[t]
    \centering
    \begin{subfigure}{0.28\linewidth}
        \centering
        \includegraphics[width=\linewidth]{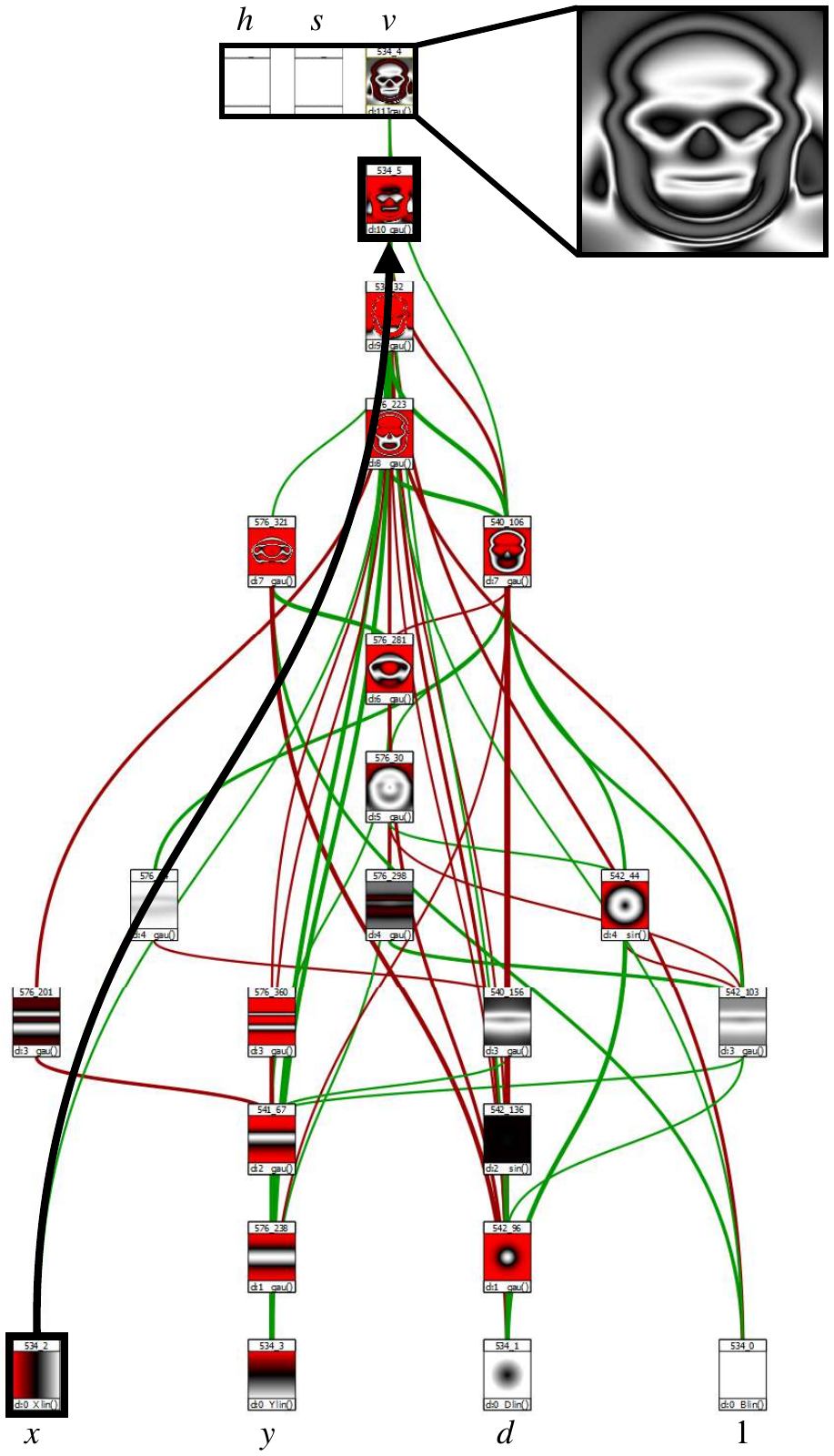}
        \caption{Original NEAT CPPN}
        \label{fig:neat_cppn}
    \end{subfigure}
    \hspace{0cm}
    \begin{subfigure}{0.66\linewidth}
        \centering
        \includegraphics[width=\linewidth]{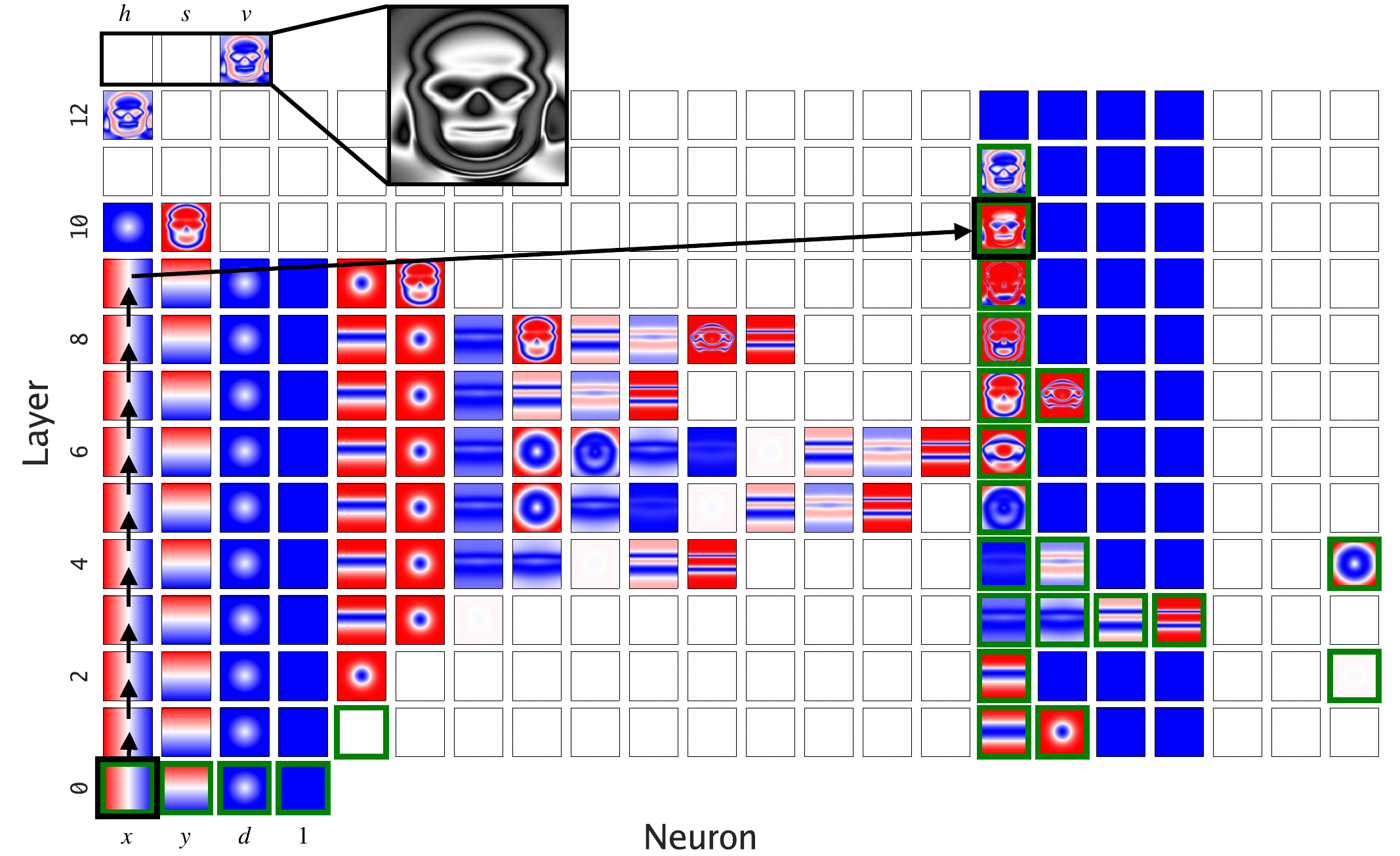}
        \caption{Layerized MLP CPPN}
        \label{fig:neat_cppn_layerized}
    \end{subfigure}
    \caption{
    \textbf{The layerization process}.
    The layerization process ports a NEAT CPPN like the one shown in (a) to a CPPN with a dense MLP architecture, as shown in (b).
    The underlying computation performed by these two networks is identical.
    The aim of this conversion is to create an architecture that is more compatible with optimization via SGD.
    In (a), neuron feature map colors represent values: red = -1, black = 0, white = 1. In (b), red = -1, white = 0, and blue = 1.
    Green borders in (b) indicate neurons with novel latent representations (not seen in previous layers), showing there are the same number of novel neurons in both networks.
    In both (a) and (b), we highlight two corresponding neurons (with black borders) and the connection between them (with black arrow), showing how this structure is layerized.
    }
    \label{fig:layerization}
\end{figure}

The original CPPNs in Picbreeder were produced by the NEAT algorithm.
Because NEAT grew the networks from small to large through a process of complexification, the resulting architectures were graph-like, with connection patterns that were highly unstructured compared to modern deep learning architectures.
Figure~\ref{fig:neat_cppn} provides an example of a NEAT-generated CPPN.
These custom topologies often feature connections between neurons in earlier layers and those in middle or later layers, resulting in a distinctive, irregular flow of information.
Moreover, each neuron could use a specific activation function selected from a predefined set.

SGD is commonly applied to multilayer perceptrons (MLPs), which are dense, sequential networks where each neuron in a given layer receives incoming connections from \textit{all and only} the neurons in the previous layer (we do not use residual connections).
This convention differs drastically from the exotic NEAT architectures.

\begin{figure}[t]
    \centering
    \includegraphics[width=0.7\textwidth]{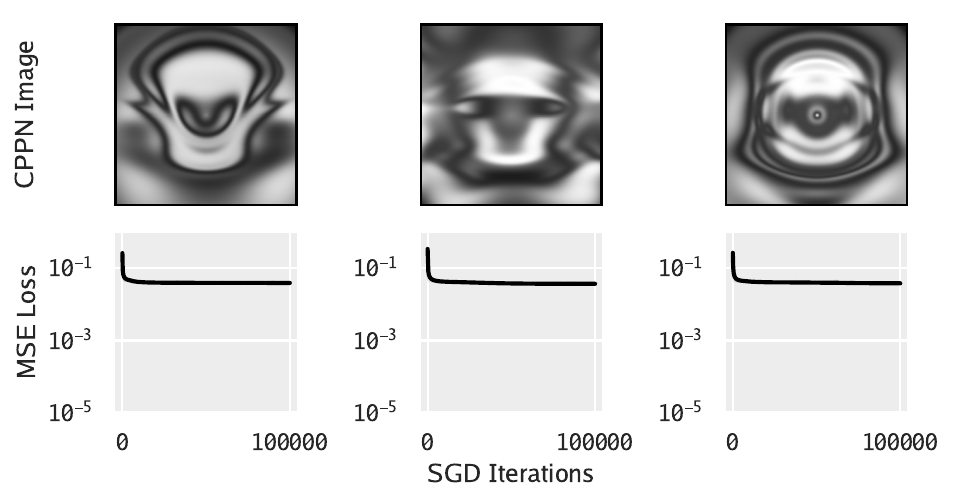}
    \caption{
    \textbf{SGD training curves on the original Picbreeder architecture.}
    To attempt SGD on the original Picbreeder architecture, we take the raw Picbreeder skull architecture (before layerization), randomize the weights, and then perform SGD on it to recreate the skull as a target.
    All three seeds of SGD fail on this architecture to effectively optimize and get stuck in local optima. (Compare these images and loss curves to those in Fig.~\ref{fig:training_curve}.)
    }
    \label{fig:training_curve_pb_arch}
\end{figure}

In fact, applying SGD to search for weights in the evolved NEAT architectures, which are sparser than dense MLPs, often fails by getting stuck in local minima, as shown in Figure~\ref{fig:training_curve_pb_arch}.
Thus, even given the architectural inductive bias, SGD is unable to exploit it properly.

To determine whether SGD can discover these kinds of NEAT CPPNs (with UFR), we wanted to ensure that SGD could operate within a comparable search space to NEAT---or at least within a space that included the final NEAT CPPN.
To achieve this scenario, we converted the Picbreeder NEAT CPPN into an MLP format, mapping each weight individually through a process we call \emph{layerization}, which is visualized in Figure~\ref{fig:neat_cppn_layerized}.

At a high level, layerization involves first determining the MLP architecture with the smallest width and depth, and neuron activation functions, that can fully capture the information flow graph of the NEAT CPPN.
Then, the weights of the NEAT CPPN are ported over to perfectly reconstruct the computation graph.

Concretely, the process begins by topologically sorting the NEAT CPPN neurons, assigning a layer-wise ordering to the neurons (with neurons that can be computed in parallel placed in the same layer).
Then, for each neuron, we determine the latest layer in which its value is needed during the forward pass, after which, that information can be safely discarded.
For example, if a neuron computed in layer 3 is used at layers 4, 6, and 9, its output only needs to persist until layer 9.
Using this information, we assign each layer the \textit{minimal} set of the NEAT neurons necessary to fully represent the forward pass state at that step, in a Markovian fashion---i.e. the current layer contains all the information needed, without referencing earlier layers.
Taking the union of these sets across all layers yields the smallest MLP architecture that can preserve the full computation of the original CPPN.

Next, the appropriate activation functions are assigned to each neuron in all layers, ensuring that the information flow is preserved.
For example, if the NEAT CPPN contains neurons with sinusoid, absolute value, and Gaussian activation functions, all the MLP layers are constructed to include these activation functions in a way that matches the original computations.
For skip connections in the NEAT CPPN---such as a connection from a neuron in layer 2 to one in layer 7---we introduce ``identity'' neurons.
These neurons pass the information unchanged through each intermediate layer (from layer 2 to layer 3, layer 3 to layer 4, and so on) until it reaches 6, where it can be used by the neuron in layer 7.
This accommodation explains the repetition of certain activation maps in the intermediate representations shown in Figures~\ref{fig:fmaps_576_pb},~\ref{fig:fmaps_4376_pb},~\ref{fig:layerization} and ~\ref{fig:fmaps_5736_pb}.
Importantly, the layerized CPPN performs the exact same computation as the original NEAT CPPN from which it was derived.

This new MLP architecture provides a function space where (1) conventional SGD can succeed in reconstructing the target image (Section \ref{sec:sgd_details})and (2) where there exists a known oracle UFR of the target image (the ported NEAT CPPN).

We note that the layerization process was originally introduced by \citet{budenbender2021revisiting} for efficient inference of NEAT networks and was independently reinvented in our work.

\section{SGD Training Details}
\label{sec:sgd_details}

\begin{figure}[t]
    \centering
    \includegraphics[width=0.7\textwidth]{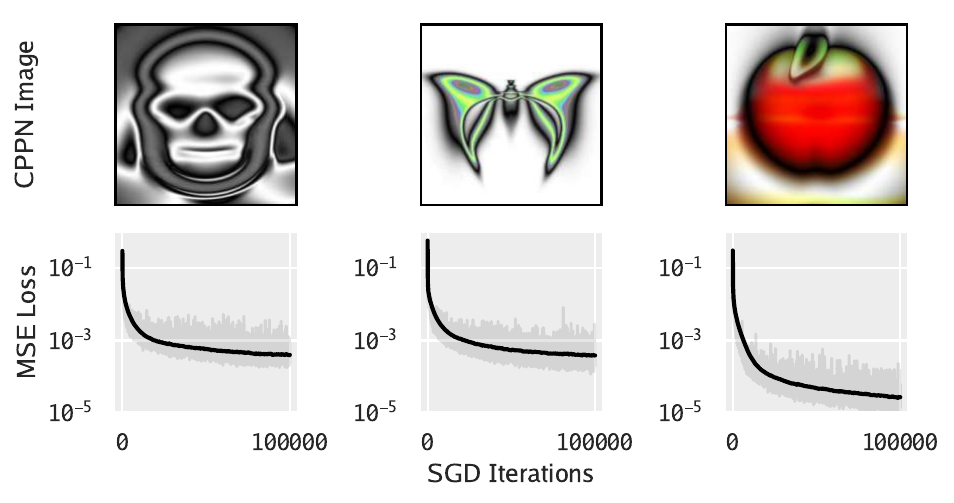}
    \caption{
    \textbf{SGD training curves on layerized architectures.}
    The training loss curves for the conventional SGD CPPNs (layerized architectures) show a smooth learning profile for all three image targets.
    The actual loss value seemingly shows very good ``performance'' in modeling its target, but reveals nothing about the nature of the internal representations.
    }
    \label{fig:training_curve}
\end{figure}

For the SGD trained CPPNs, we start by initializing the weight matrices using LeCun normal initialization~\citep{lecun2002efficient}.
Then, the CPPN is trained to match the target image, with a resolution of $256\times256$ using a mean squared error (MSE) loss.
The networks is trained for $100$,$000$ iterations with a full batch size (all the pixels in the entire image) and a learning rate of \num{3e-3}.
For the apple CPPN, we use a learning rate of \num{3e-4}.
Figure~\ref{fig:training_curve} shows that networks set up in this way exhibit no difficulty in learning to reproduce Picbreeder images.

\section{Other CPPN Examples}
\label{sec:other_picbreeder_examples}
This section provides additional analysis of other CPPNs from Picbreeder and their conventional SGD counterparts.

\begin{figure}[t]
    \centering
    \includegraphics[width=1.0\textwidth]{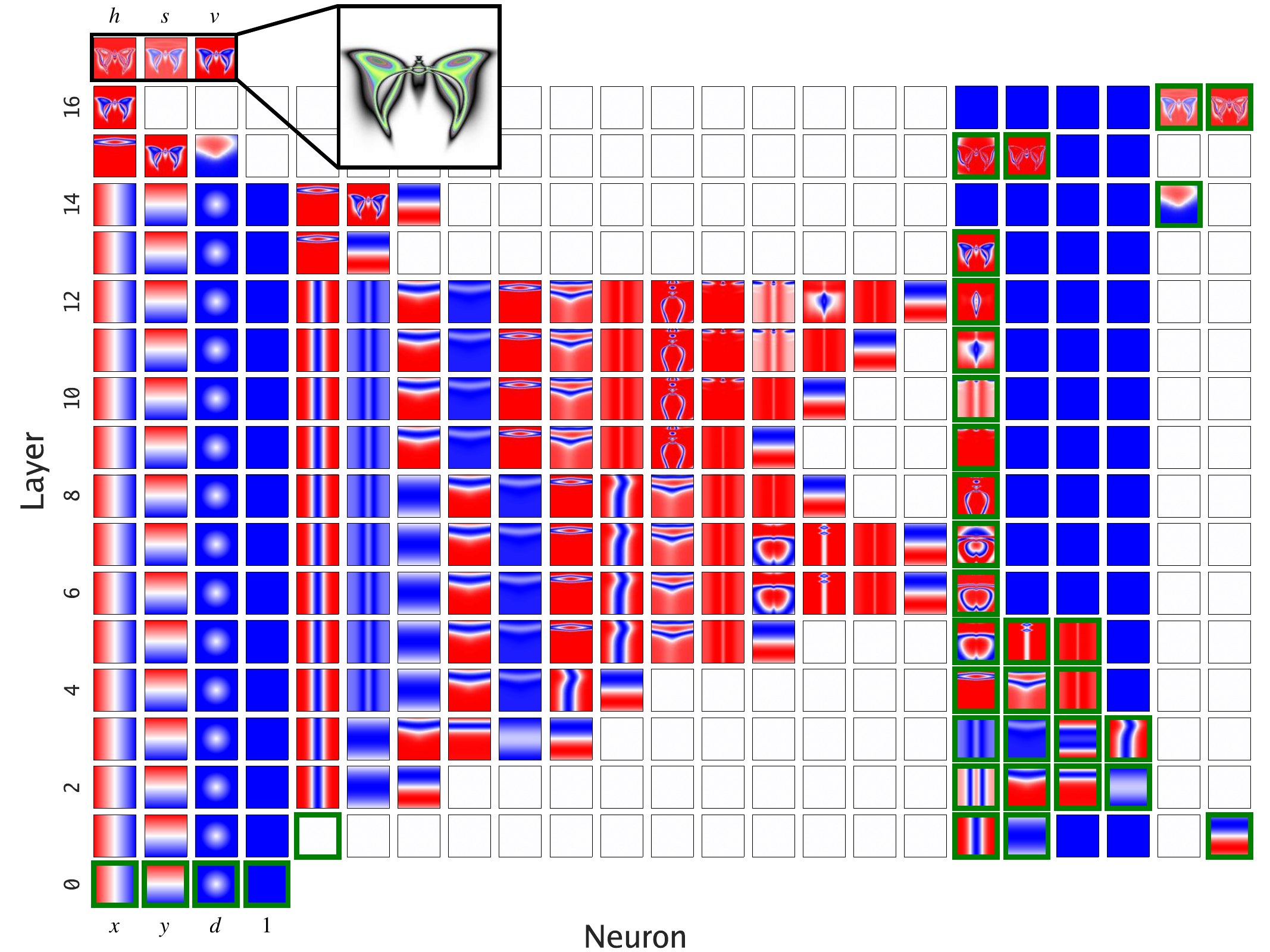}
    \caption{
    \textbf{Internal representation of the Picbreeder butterfly CPPN.}
    As with the skull, this Picbreeder representation is notably well-organized.
    Y-axis symmetry emerges early in the network and each aspect of the butterfly's output behavior is specified once, indicating a UFR, unlike the conventional SGD Butterfly CPPN shown in Figure~\ref{fig:fmaps_4376_sgd_pb}.
    }
    \label{fig:fmaps_4376_pb}
\end{figure}

\begin{figure}[t]
    \centering
    \includegraphics[width=1.0\textwidth]{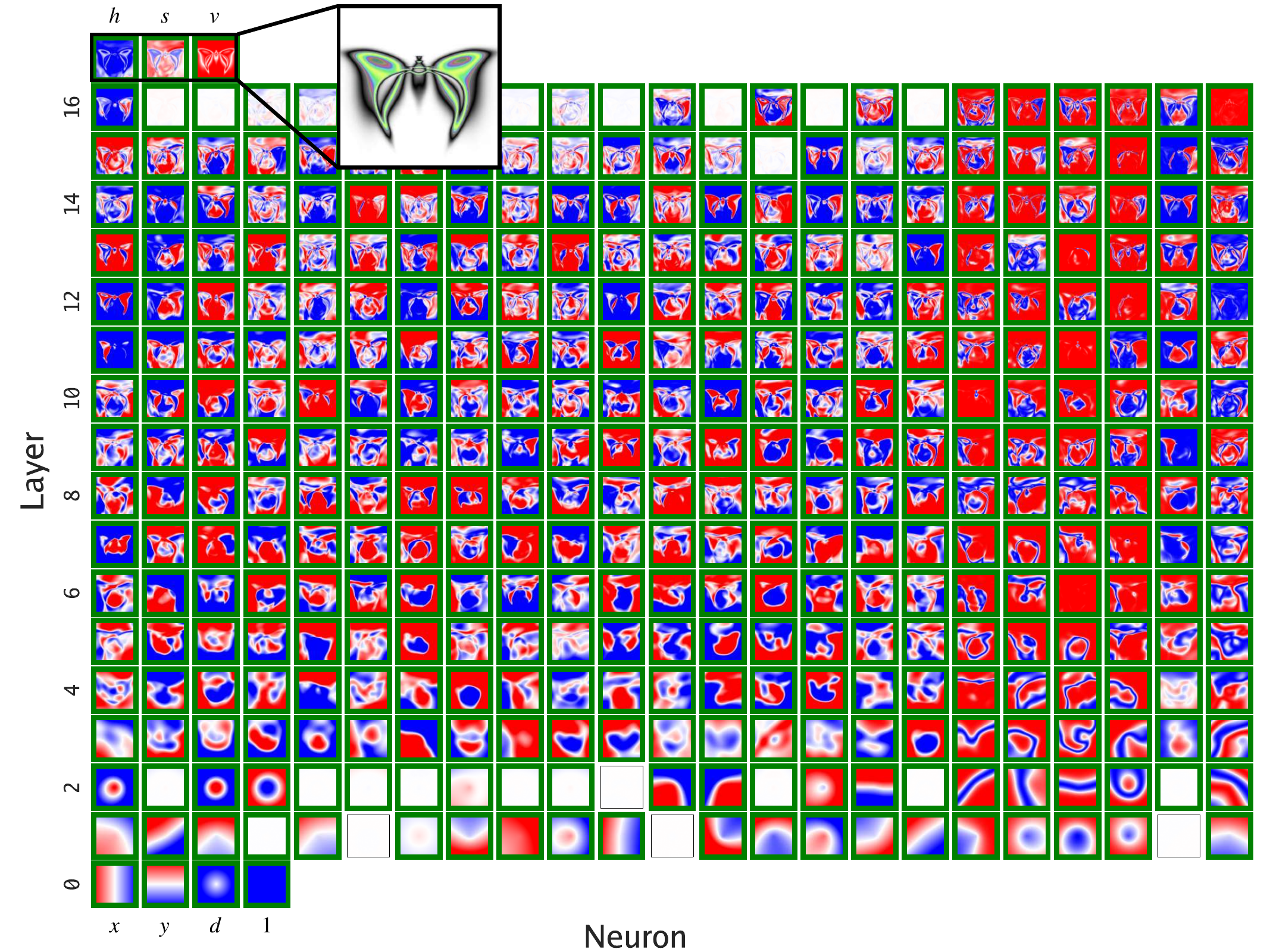}
    \caption{
    \textbf{Internal representation of the conventional SGD butterfly CPPN.}
    As with the skull, this SGD representation is nowhere near as organized as the Picbreeder representation from Figure~\ref{fig:fmaps_4376_pb}.
    The left and right wings are encoded separately, with many high-frequency patterns arbitrarily stitched together showing no holistic organization---an example of FER.
    }
    \label{fig:fmaps_4376_sgd_pb}
\end{figure}

\begin{figure}[t]
    \centering
    \begin{subfigure}{0.49\linewidth}
        \centering
        \includegraphics[width=\linewidth]{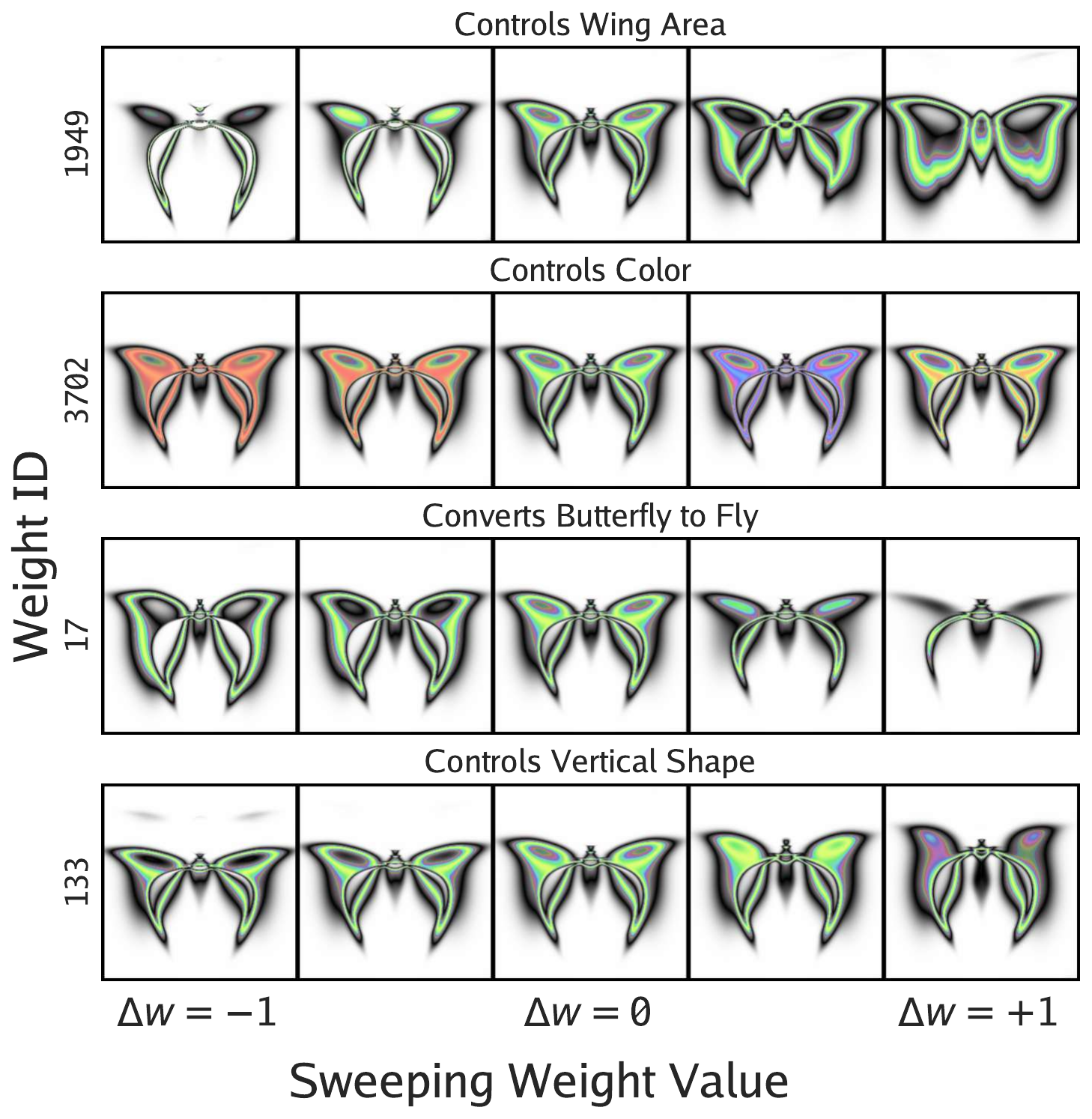}
        \caption{Picbreeder CPPN}
        \label{weight_sweeps_4376_pb}
    \end{subfigure}
    \hspace{0cm}
    \begin{subfigure}{0.49\linewidth}
        \centering
        \includegraphics[width=\linewidth]{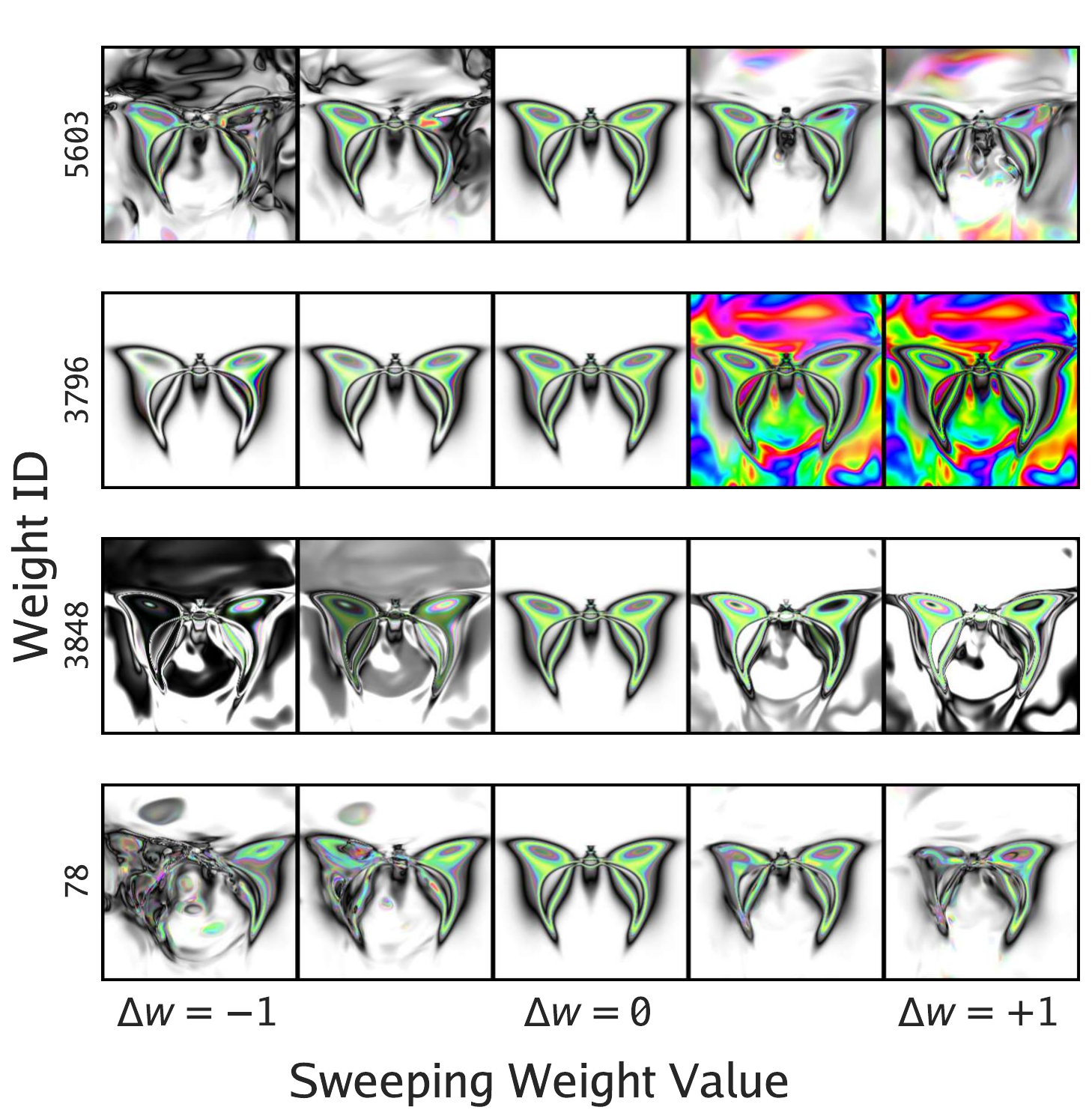}
        \caption{Conventional SGD CPPN}
        \label{weight_sweeps_4376_sgd_pb}
    \end{subfigure}
    \caption{
    \textbf{Select weight sweeps of the Picbreeder and conventional SGD butterfly CPPNs.}
    As with the skull, the sweeping the weights of the Picbreeder and SGD CPPNs reveals a fundamental difference in underlying representation.
    The Picbreeder weight sweeps show human-recognizable changes to the butterfly that also respect its y-axis symmetry.
    The SGD weight sweeps instead yield deleterious changes that break the symmetry.
    }
    \label{fig:weight_sweeps_4376}
\end{figure}

\begin{figure}[t]
    \centering
    \begin{subfigure}{1.0\textwidth}
        \centering
        \includegraphics[width=\textwidth]{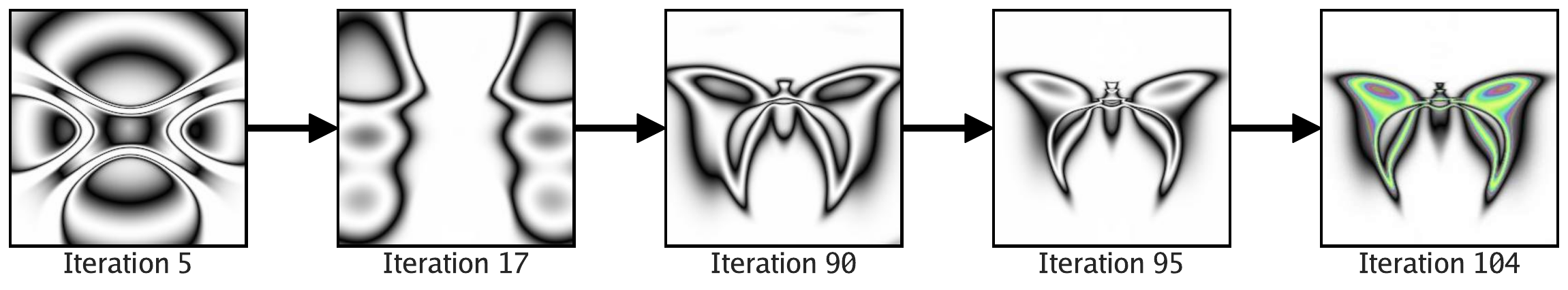}
        \caption{Picbreeder path to the butterfly}
        \label{fig:path_to_4376_pb}
    \end{subfigure}
    \vspace{0.0cm}
    \begin{subfigure}{1.0\textwidth}
        \centering
        \includegraphics[width=\textwidth]{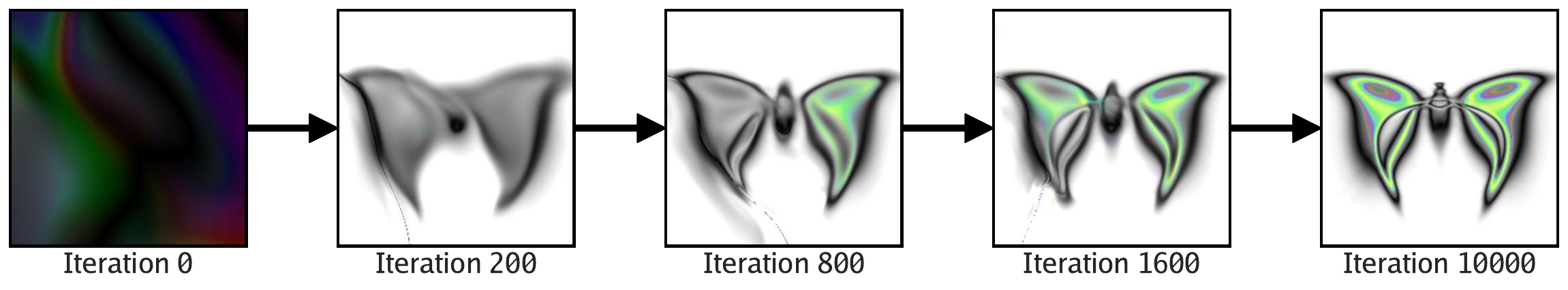}
        \caption{Conventional SGD path to the butterfly}
        \label{fig:path_to_4376_sgd_pb}
    \end{subfigure}
    \caption{
    \textbf{Differing paths taken by Picbreeder and conventional SGD to reach the butterfly image.}
    Just as with the skull, there is a huge difference in the path of discovery between Picbreeder and SGD.
    The Picbreeder path discovers symmetry early and builds on that to represent the structure and eventually the color of the butterfly.
    In contrast, the SGD path simply greedily optimizes for the butterfly and fails to recognize the underlying connection between the left and right wings.
    }
    \label{fig:path_to_4376}
\end{figure}

Figure~\ref{fig:fmaps_4376_pb} shows the Picbreeder CPPN for the butterfly, which exhibits symmetry and regular structure supported by UFR.
Figure~\ref{fig:fmaps_4376_sgd_pb} shows the conventional SGD CPPN trained to mimic the Picbreeder butterfly.
As with the skull, the SGD CPPN perfectly reconstructs the butterfly (the output behavior) but produces an internal representation suggesting FER, resembling a bag of heuristics that lacks the butterfly's symmetry.
This difference in internal representations is further illustrated by the weight sweeps in Figure~\ref{fig:weight_sweeps_4376}.
The Picbreeder CPPN, shown in Figure~\ref{weight_sweeps_4376_pb}, consistently preserves the butterfly's symmetry while making meaningful semantic changes.
In contrast, the conventional SGD CPPN in Figure~\ref{weight_sweeps_4376_sgd_pb} frequently breaks the symmetry and introduces meaningless, deleterious changes.
Similarly to the skull, it is clear from the trajectories through image space taken in search of the butterfly in Picbreeder versus conventional SGD that the path taken to the destination impacts the final representation (Figure \ref{fig:path_to_4376}).

\begin{figure}[t]
    \centering
    \includegraphics[width=1.0\textwidth]{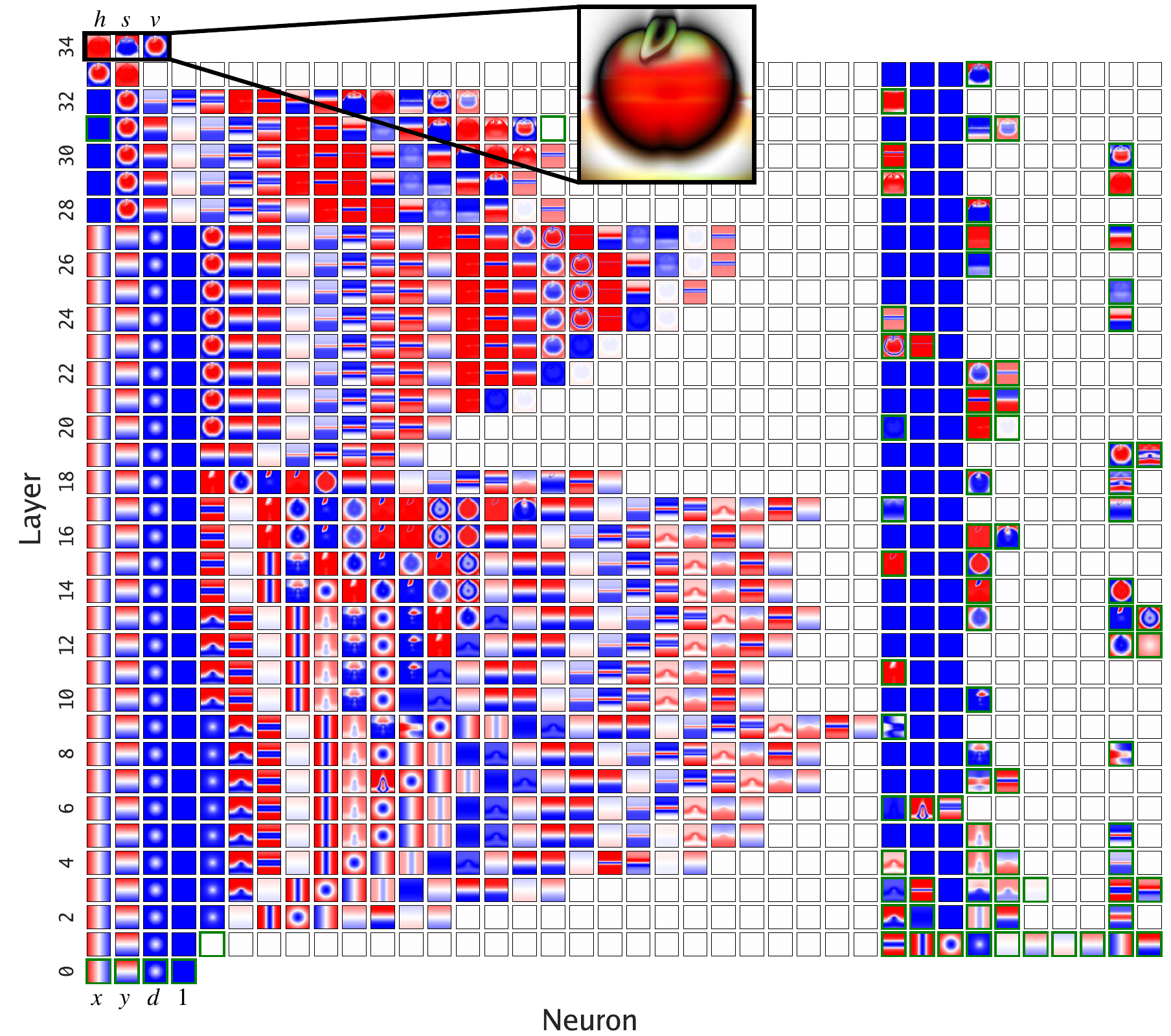}
    \caption{
    \textbf{Internal representation of the Picbreeder apple CPPN.}
    This Picbreeder representation demonstrates a UFR.
    The stem 
    and body
    of the apple are encoded separately early in the network and later combined, forming a modular representation.
    There are only a small number of genuinely novel feature maps (highlighted with green borders), illustrating the efficiency of the apple’s representation.
    }
    \label{fig:fmaps_5736_pb}
\end{figure}

\begin{figure}[t]
    \centering
    \includegraphics[width=1.0\textwidth]{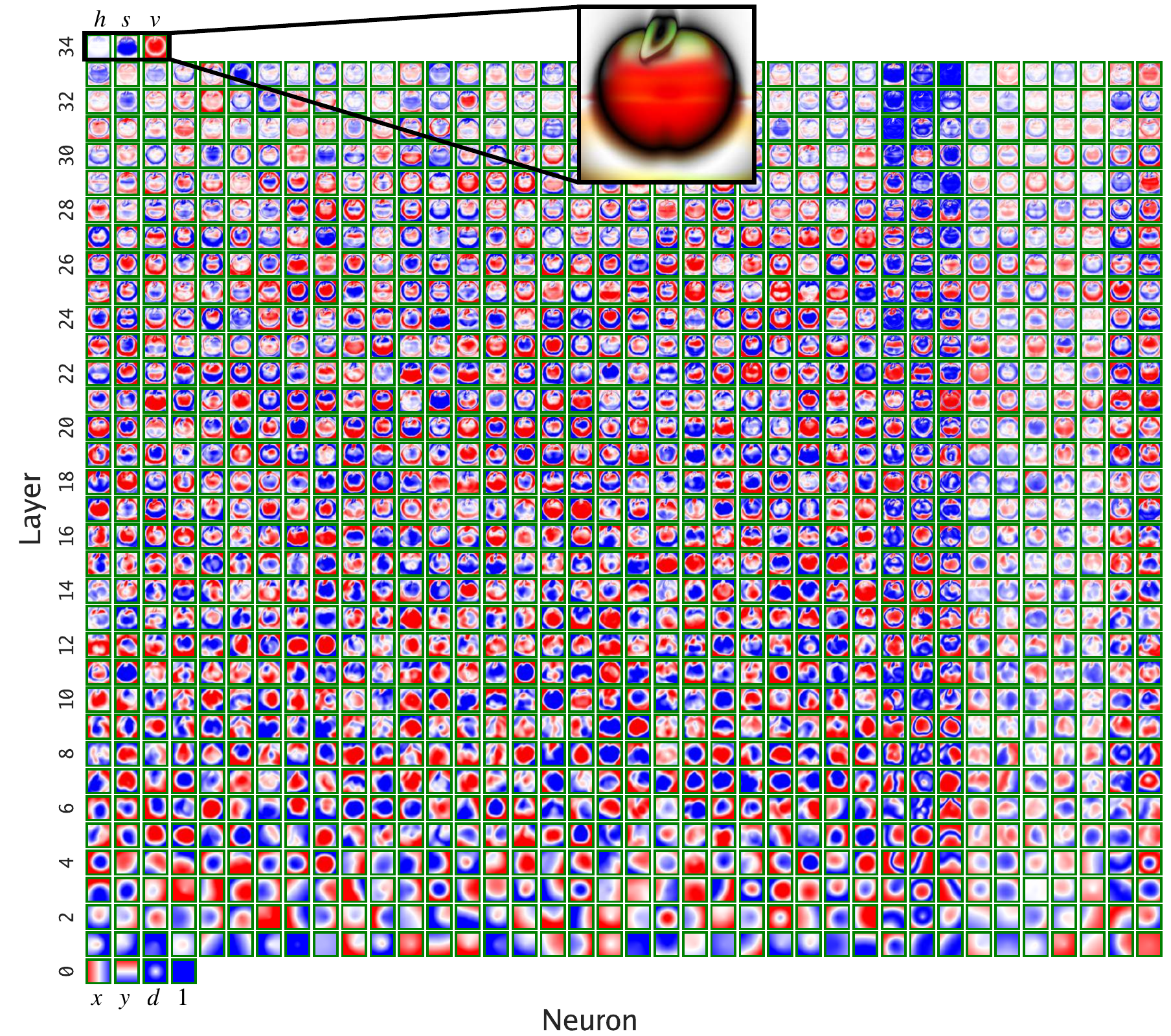}
    \caption{
    \textbf{Internal representation of the conventional SGD apple CPPN.}
    SGD discovers a fragmented representation of the apple, in which nearly every neuron is utilized in a highly entangled manner.
    There is no clear separation between the apple, stem, and background---everything is intertwined in a complex, spaghetti-like structure.
    }
    \label{fig:fmaps_5736_sgd_pb}
\end{figure}

\begin{figure}[t]
    \centering
    \begin{subfigure}{0.49\linewidth}
        \centering
        \includegraphics[width=\linewidth]{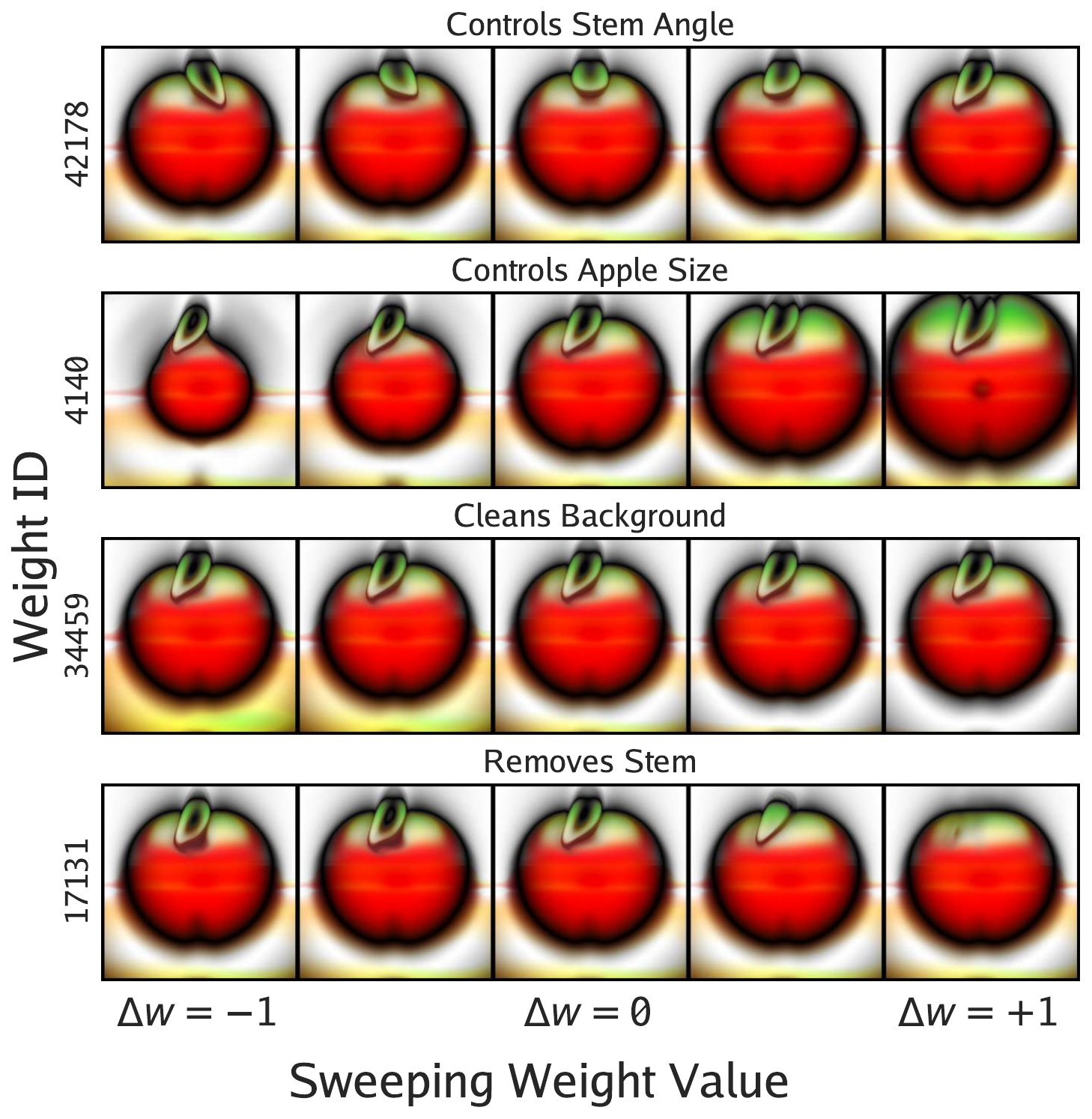}
        \caption{Picbreeder CPPN}
        \label{fig:weight_sweeps_5736_pb}
    \end{subfigure}
    \hspace{0cm}
    \begin{subfigure}{0.49\linewidth}
        \centering
        \includegraphics[width=\linewidth]{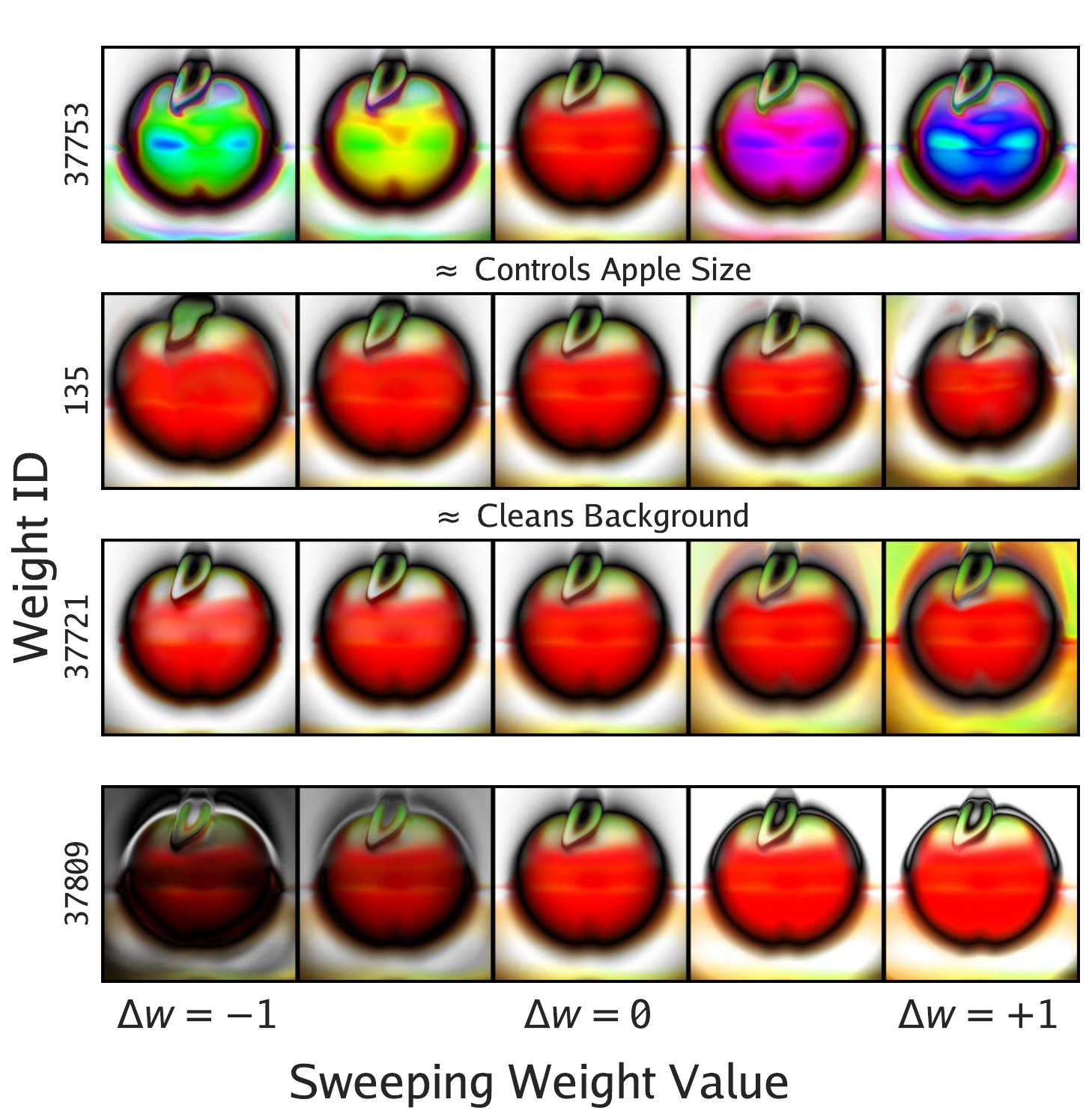}
        \caption{Conventional SGD CPPN}
        \label{fig:weight_sweeps_5736_sgd_pb}
    \end{subfigure}
    \caption{
    \textbf{Select weight sweeps of the Picbreeder and conventional SGD apple CPPNs.}
    The modularity of the Picbreeder apple representation means there are weights that produce changes meaningful to humans (a).
    The SGD representation of this same apple does not produce such nice changes (b).
    Even the weight sweeps in (b) that evoke similar changes, like controlling the apple size, also significantly change \emph{other} parts of the apple (like the stem chaging with the apple size and the stem color changing with the background color), suggesting inherent entanglement.
    }
    \label{fig:weight_sweeps_5736}
\end{figure}

Figure~\ref{fig:fmaps_5736_pb} shows the internal representations of the Picbreeder CPPN for the apple.
Again, the Picbreeder CPPN has a UFR, in this case that properly decomposes the background, apple body, and stem into separate factored units.
The conventional SGD CPPN trained for this apple is shown in Figure~\ref{fig:fmaps_5736_sgd_pb}.
This SGD CPPN does not demonstrate a clean decomposition and rather models the stem and body together, signaling FER.
These observations are further amplified by the weight sweeps in Figure~\ref{fig:weight_sweeps_5736}.
Incredibly, in the Picbreeder CPPN in Figure~\ref{fig:weight_sweeps_5736_pb}, the decomposiion of the background, body, and stem of the apple allow some weights to control  these aspects of the apple independently.
There even exists a single weight solely for ``swinging'' the apple stem around the top of the apple by changing its angle.
The conventional SGD CPPN's weight sweeps in Figure~\ref{fig:weight_sweeps_5736_sgd_pb} shows this CPPN often entangles multiple concepts together in one module, such as the body, stem, and color of the apple.

We provide all weight sweeps for the Picbreeder and SGD versions of the skull, butterfly, and apple at the \href{\codelink}{code link}.

\section{ReLU CPPNs}
\label{sec:relucppn}

Some may argue that the MLP architectures we construct are abnormal due to their diversity of activation functions within a layer.
To demonstrate the FER occurs also with more conventional activations functions, we also train a standard MLP with only ReLU activations.

\begin{figure}[t]
    \centering
    \includegraphics[width=1.0\textwidth]{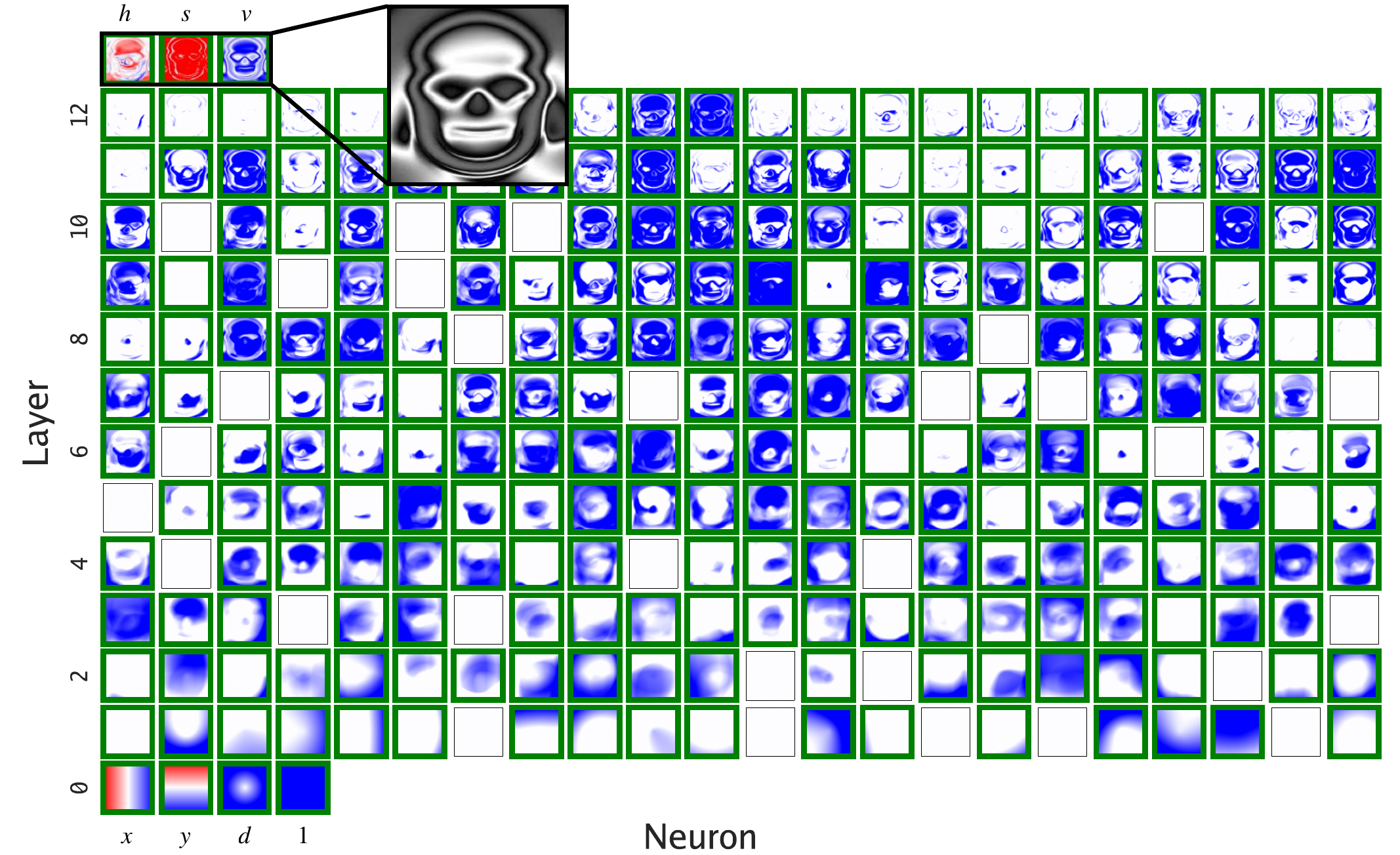}
    \caption{
    \textbf{Internal representation of the conventional SGD skull CPPN (ReLU architecture).}
    Even with a more standard ReLU architecture, SGD produces FER for the skull.
    The feature maps show skull shapes, but with erratic high frequency artifacts added onto it.
    These high frequency patterns end up canceling out at the end, but still degrade the internal representation.
    }
    \label{fig:fmaps_576_sgd_relu}
\end{figure}

\begin{figure}[t]
    \centering
    \includegraphics[width=1.0\textwidth]{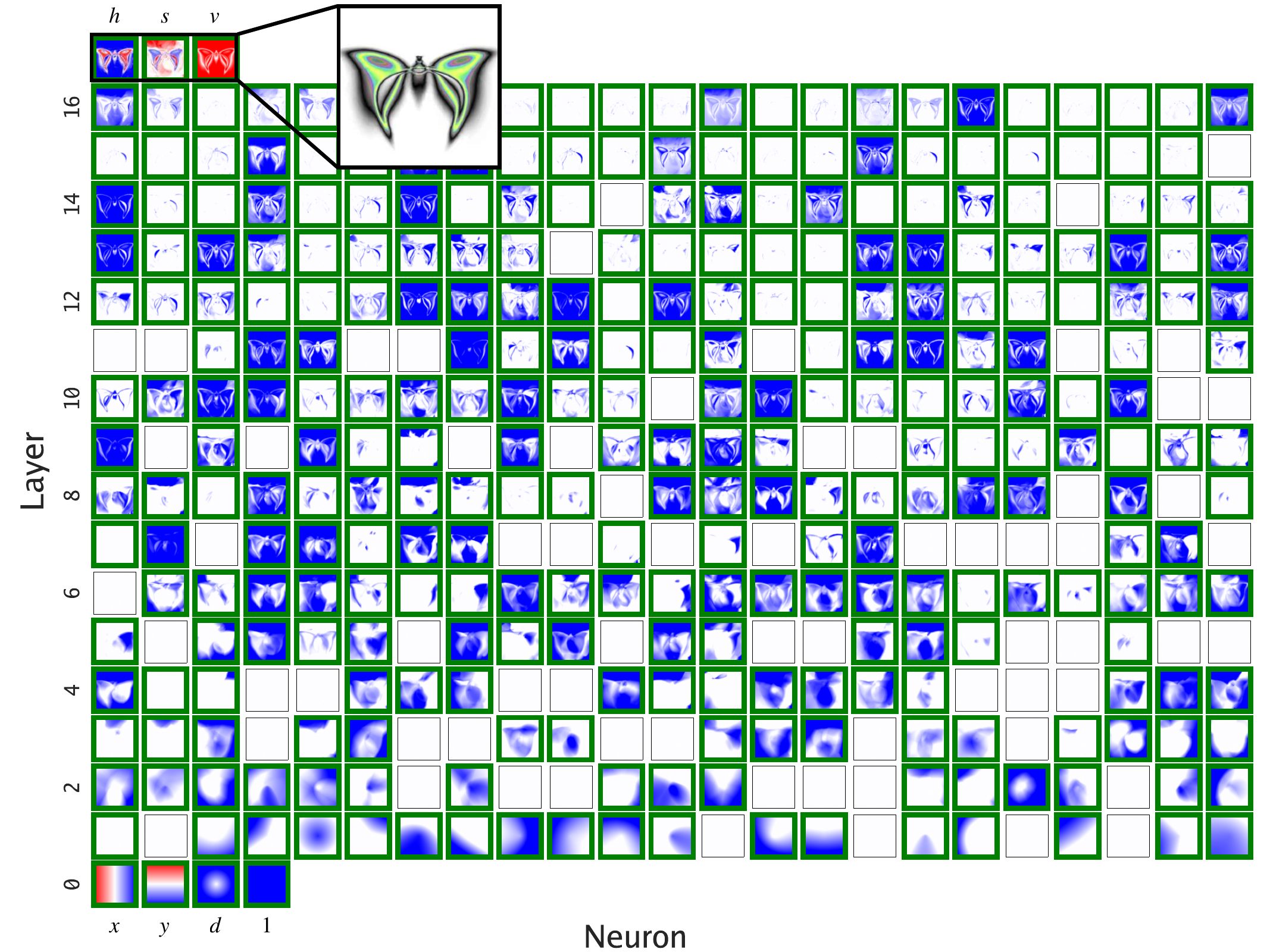}
    \caption{
    \textbf{Internal representation of the conventional SGD butterfly CPPN (ReLU architecture).}
    Like with the ReLU skull, numerous asymmetric and disorganized intermediate pattern hint at FER with ReLUs.
    }
    \label{fig:fmaps_4376_sgd_relu}
\end{figure}

\begin{figure}[t]
    \centering
    \includegraphics[width=1.0\textwidth]{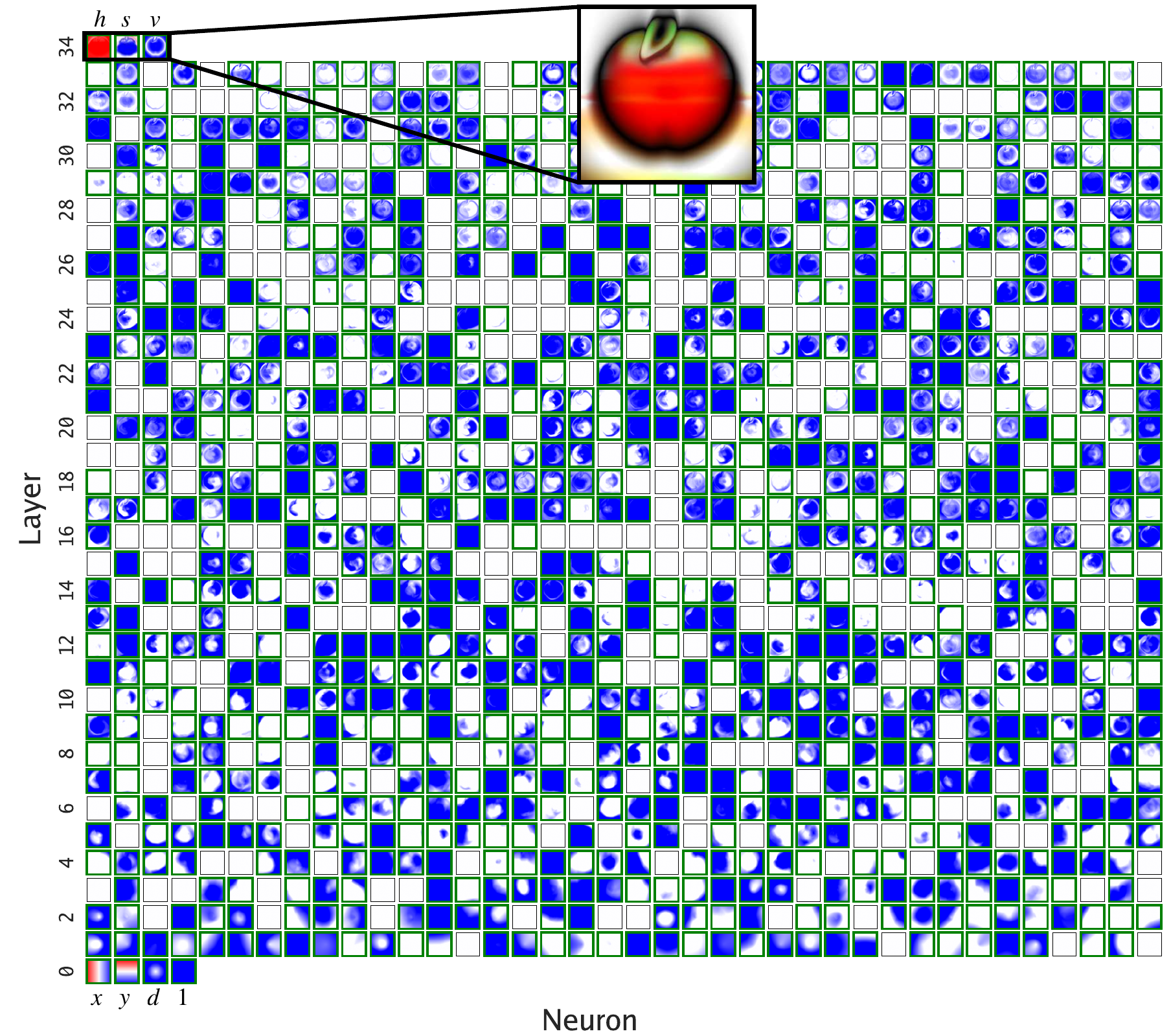}
    \caption{
    \textbf{Internal representation of the conventional SGD apple CPPN (ReLU architecture).}
The lack of organization or structured decomposition here again hints at FER with ReLUs.
    }
    \label{fig:fmaps_5736_sgd_relu}
\end{figure}

The intermediate representation of these MLPs are shown for the skull in Figure~\ref{fig:fmaps_576_sgd_relu}, for the butterfly in Figure~\ref{fig:fmaps_4376_sgd_relu}, and for the apple in Figure~\ref{fig:fmaps_5736_sgd_relu}.
All of these visualizations still exhibit FER.

\section{Change of Basis of the Feature Space}
\label{sec:pca_feature_space}

Because deep learning with SGD uses matrix multiplications to model neural computation, the features at any layer can be represented in one of many possible bases for that vector space.
There is no reason that meaningful directions of variation have to align with a single weight and neuron, because the neurons represent just a single basis of the feature space out of many.

Thus a lingering question is whether it could be possible that conventional SGD does actually find UFR, but just in a different basis of the feature space where the organization is not readily apparent.

\begin{figure}[t]
    \centering
    \includegraphics[width=1.0\textwidth]{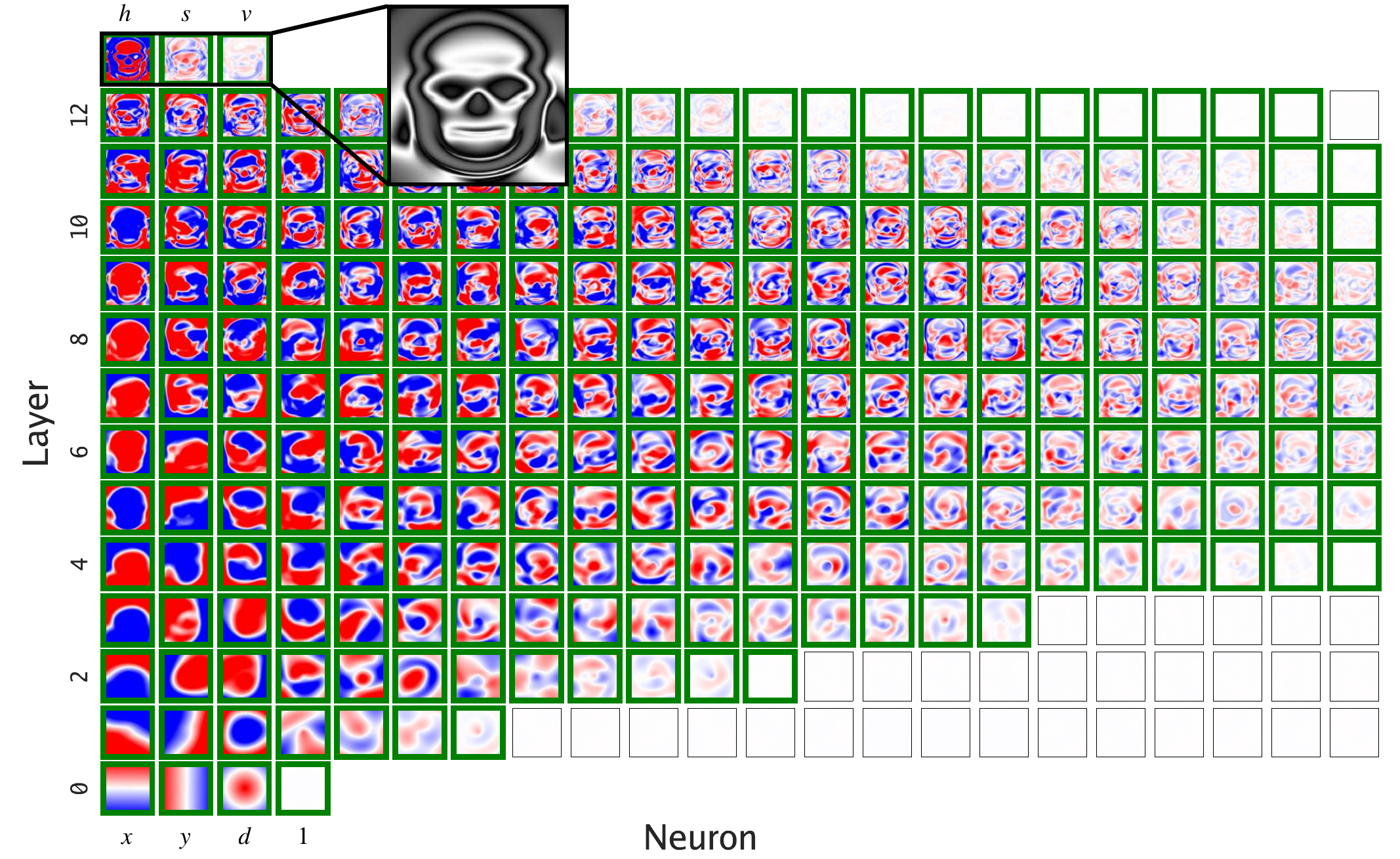}
    \caption{
    \textbf{PCA of the internal representation of the conventional SGD skull CPPN.}
    This figure visualizes the principal component neurons of the representation space, ordered by explained variance.
    This new lens into the feature space does show slightly less fracture, but comes nowhere close to the UFR of the Picbreeder skull CPPN.
    More details are provided in Section~\ref{sec:pca_feature_space}.
    }
    \label{fig:fmaps_576_sgd_pb_pca}
\end{figure}

To address this question, we attempt to ``untransform'' the feature space to a more principled basis where salient features might be easier to observe.
In particular, for each layer, the feature space contains the set of latent data points (computed from each input data point).
In this feature space, we compute the Principal Component Analysis (PCA) transform of the data.
In other words, the feature data is projected onto the principal components to achieve the transformed feature space.
This new feature space can then be visualized as neuron activations like before (where each feature dimension is a neuron) over all inputs to the network, as shown in Figure~\ref{fig:fmaps_576_sgd_pb_pca}.
This visualization shows slightly improved organizational structure compared to the vanilla SGD Skull CPPN from Figure~\ref{fig:fmaps_576_sgd_pb}, but still comes nowhere close to the striking UFR in the Picbreeder Skull CPPN from~\ref{fig:fmaps_576_pb}.

However, note that PCA is merely one possible transformation among many.
An exhaustive study would require addressing all the other possible orthonormal transformations that may reveal UFR.
However, there is a way to see that UFR is unlikely to be discovered no matter the transformation:
weight sweeps can be performed that are agnostic to any transformation of the feature space. 
While previously in Figure~\ref{fig:weight_sweeps_576} we swept the value of single weights and observed how the network’s output behavior changed, now instead of modifying a single weight directly, we sweep the value of a random column of the weight matrix.

\begin{figure}[t]
    \centering
    \begin{subfigure}{0.49\linewidth}
        \centering
        \includegraphics[width=\linewidth]{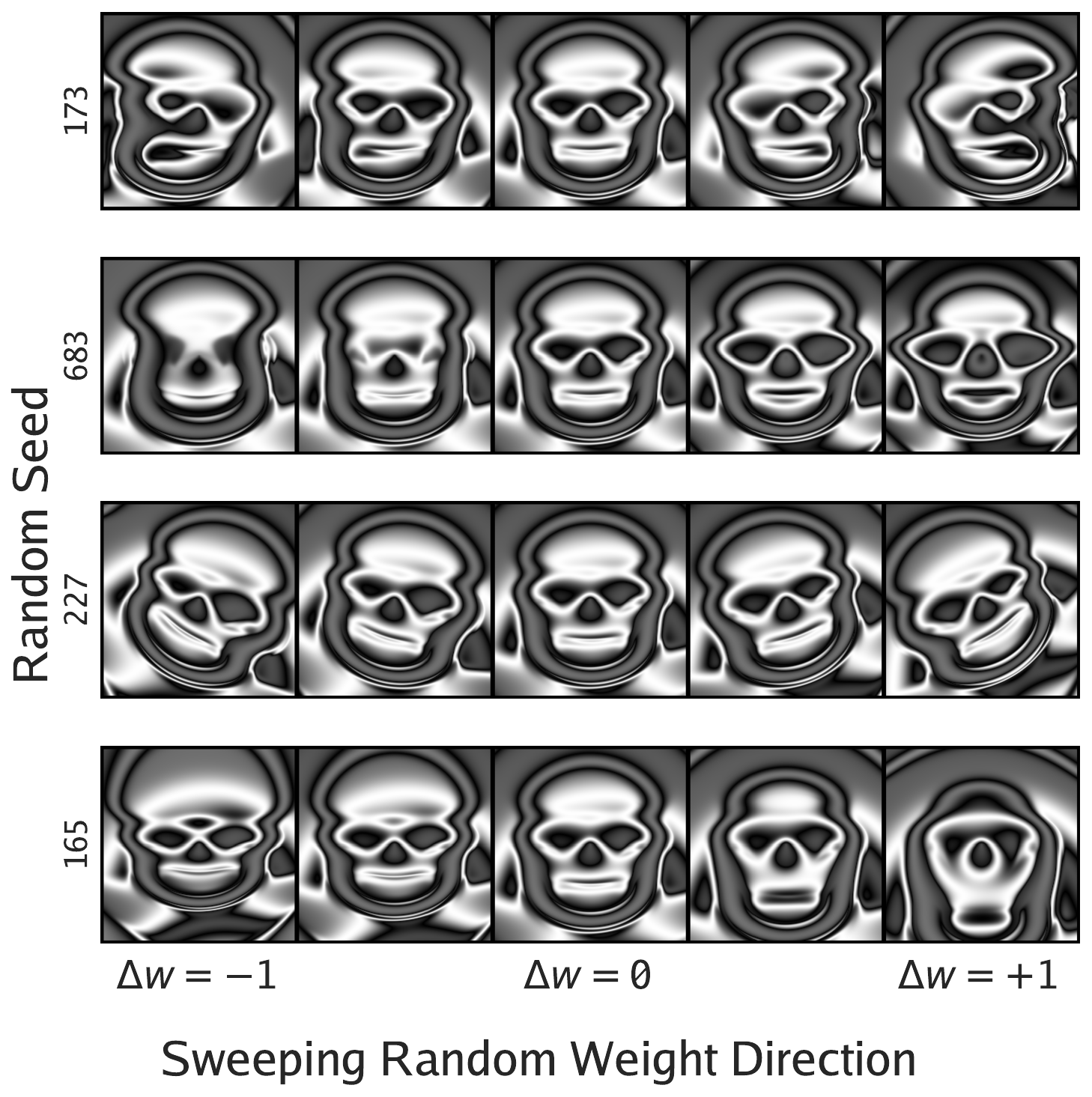}
        \caption{Picbreeder CPPN}
        \label{weight_sweeps_576_pb_random}
    \end{subfigure}
    \hspace{0cm}
    \begin{subfigure}{0.49\linewidth}
        \centering
        \includegraphics[width=\linewidth]{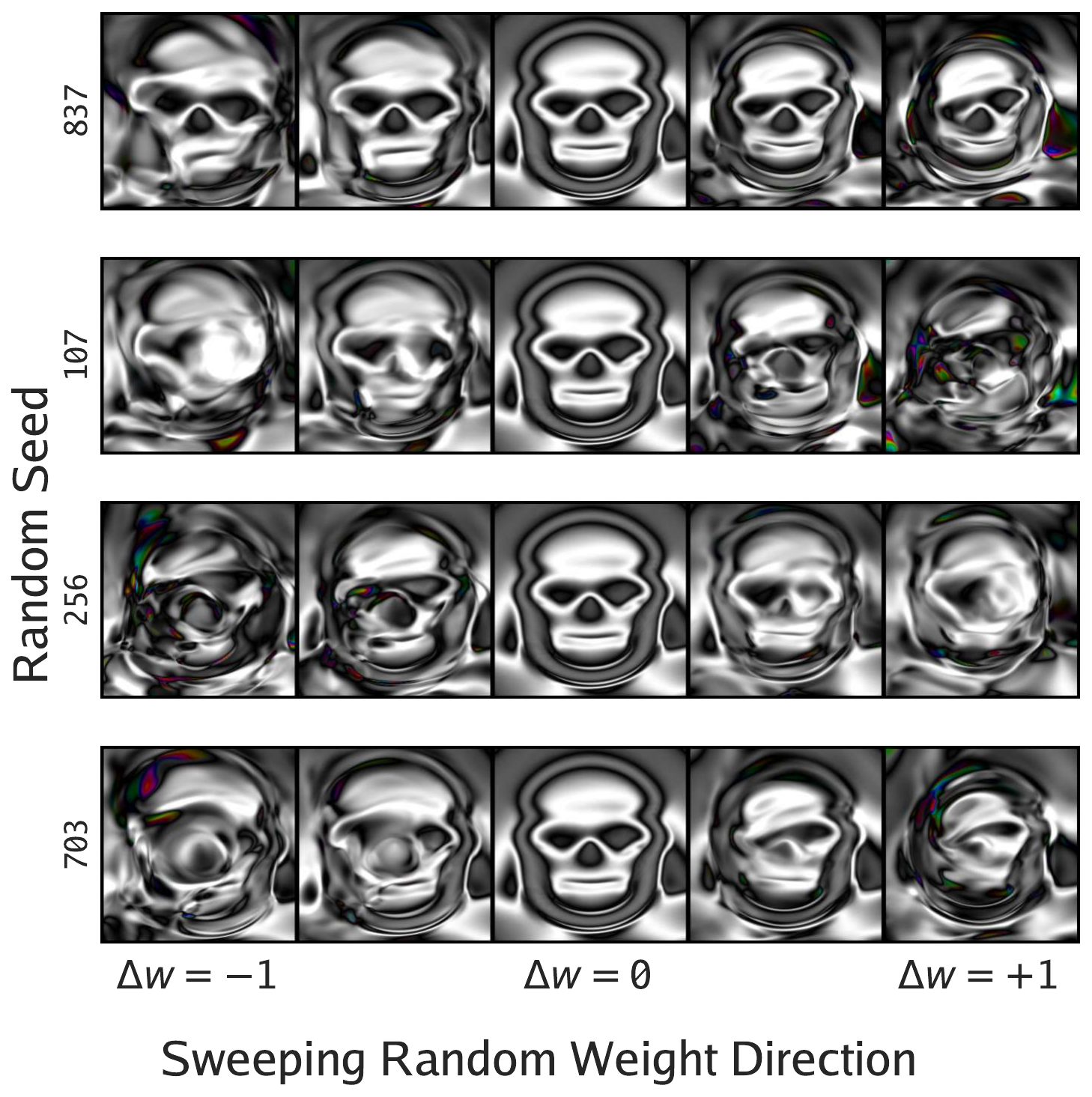}
        \caption{Conventional SGD CPPN}
        \label{weight_sweeps_576_sgd_pb_random}
    \end{subfigure}
    \caption{
    \textbf{Muti-weight sweeps of the Picbreeder and conventional SGD butterfly CPPNs.}
    Rather than sweeping a single weight in the CPPN, we sweep the value of a column of the weight matrix in the direction of a random vector.
    These weight sweeps are agnostic to any orthonormal transformation of the weight matrix.
    Even along these random vector directions, there is a clear difference in the organization between the Picbreeder and SGD internal representations.
    More details are provided in Section~\ref{sec:pca_feature_space}.
    }
    \label{fig:weight_sweeps_576_random}
\end{figure}

Specifically, a random layer and a random column from its weight matrix are selected and then swept in the direction of a random unit vector.
This approach is equivalent to sweeping a single weight of a randomly chosen orthonormal transformation of the weight matrix.
The results of these weight sweeps are shown in Figure~\ref{fig:weight_sweeps_576_random}.
A larger set of sweeps is provided at the \href{\codelink}{code link}.
The Picbreeder CPPN weight sweeps again show meaningful changes to the skull while respecting the y-axis symmetry, while the conventional SGD CPPN weight sweeps reveal meaningless distortions.
These results provide further evidence of the fundamental difference in the nature of the Picbreeder versus conventional SGD representations.

\newpage

\end{document}